\documentclass[a4paper,fleqn]{cas-dc}

\usepackage[authoryear]{natbib}
\def\tsc#1{\csdef{#1}{\textsc{\lowercase{#1}}\xspace}}
\tsc{WGM}
\tsc{QE}
\usepackage{xcolor,colortbl}
\usepackage{graphicx}
\usepackage{multirow}
\usepackage{float} %to use figure inside minipage
\usepackage{version} %to use comment

\newcommand{\cm}{$\bullet$} %\checkmark
\definecolor{lightgray}{rgb}{0.8,0.8,0.8}
\definecolor{darkred}{rgb}{0.8,0,0}
\definecolor{darkgreen}{rgb}{0,0.5,0}
\definecolor{darkorange}{rgb}{0.75,0.5,0}
\definecolor{darkblue}{rgb}{0,0,0.5}
\definecolor{col_review}{rgb}{0,0,0}
\definecolor{r1_col}{rgb}{0,0,1}
\newcommand{\rcm}{{\color{black}$\sim$}} %\cellcolor{red!20}
\graphicspath{{./img/}}
\newcommand{\revision}[1]{\textcolor{black}{#1}}

\newcommand{\GG}{\mathcal{G}}
\newcommand{\SSS}{\mathcal{S}}

\newcommand{\eqsize}{\small}

\setcitestyle{authoryear,open={(},close={)},maxnames=2}

\begin{document}
\let\WriteBookmarks\relax
\def\floatpagepagefraction{1}
\def\textpagefraction{.001}

% Short title
\shorttitle{Superpixel Segmentation: A Long-Lasting Ill-Posed Problem}    

% Short author
\shortauthors{R. Giraud and M. Clément}  

% Main title of the paper
\title [mode = title]{Superpixel Segmentation: A Long-Lasting Ill-Posed Problem}  

% Title footnote mark
% eg: \tnotemark[1]
%\tnotemark[1] 

% Title footnote 1.
% eg: \tnotetext[1]{Title footnote text}
%\tnotetext[1]{} 

% First author
%
% Options: Use if required
% eg: \author[1,3]{Author Name}[type=editor,
%       style=chinese,
%       auid=000,
%       bioid=1,
%       prefix=Sir,
%       orcid=0000-0000-0000-0000,
%       facebook=<facebook id>,
%       twitter=<twitter id>,
%       linkedin=<linkedin id>,
%       gplus=<gplus id>]

\author[1]{Rémi Giraud}[orcid=0009-0004-0043-7699]

% Corresponding author indication
\cormark[1]

% Footnote of the first author
%\fnmark[1]

% Email id of the first author
\ead{remi.giraud@ims-bordeaux.fr}

% URL of the first author
%\ead[url]{}

% Credit authorship
% eg: \credit{Conceptualization of this study, Methodology, Software}
%\credit{}

% Address/affiliation
\affiliation[1]{organization={Univ. Bordeaux, Bordeaux INP, IMS, CNRS UMR 5218},
            addressline={351 cours de la libération}, 
            city={Talence},
%          citysep={}, % Uncomment if no comma needed between city and postcode
            postcode={33400}, 
            state={Nouvelle Aquitaine},
            country={France}}

\author[2]{Michaël Clément}[orcid=0000-0002-0899-3428]

% Footnote of the second author
%\fnmark[2]

% Email id of the second author
\ead{michael.clement@labri.fr}

% URL of the second author
%\ead[url]{}

% Credit authorship
%\credit{}

% Address/affiliation
\affiliation[2]{organization={Univ. Bordeaux, Bordeaux INP, LaBRI, CNRS UMR 5800},
            addressline={351 cours de la libération}, 
            city={Talence},
%          citysep={}, % Uncomment if no comma needed between city and postcode
            postcode={33400}, 
            state={Nouvelle Aquitaine},
            country={France}}

% Corresponding author text
\cortext[1]{Corresponding author}

% Footnote text
%\fntext[1]{}

% For a title note without a number/mark
%\nonumnote{}

% Here goes the abstract
\begin{abstract}
For many years, image over-segmentation into superpixels has been essential to computer vision pipelines, by creating homogeneous and identifiable regions of similar sizes.
Such constrained segmentation problem would require a clear definition and specific evaluation criteria.
However, the validation framework for superpixel methods, typically viewed as standard object segmentation, has rarely been thoroughly studied.
%
%For many years, image over-segmentation into superpixels
%has been a key tool in many computer vision pipelines. 
%\emph{i.e.}, 
%The aim is to produce 
%homogeneous regions of variable shapes
%but rather regular in terms of sizes and clearly identifiable.
%
%\emph{i.e.}, 
%a pseudo-grid 
%The aim is to segment an image into homogeneous regions having variable shapes
%but clearly identifiable and with similar sizes.
%Such constrained segmentation problem would require proper definition and dedicated
%specific evaluation criteria.
%
%Nevertheless, little or no works have deeply questioned 
%the validation framework of superpixel methods, generally considered as a standard object segmentation problem.
% Many different methods have been proposed to .
%
In this work, we first take a step back %on superpixel segmentation, showing
to show that superpixel segmentation is fundamentally an ill-posed problem,
due to the implicit regularity constraint on the shape and size of superpixels.
We also demonstrate through a novel comprehensive
study that the literature suffers from %an evaluation problem, where 
only evaluating certain aspects, 
sometimes incorrectly and with inappropriate metrics.
%
%In this work, we aim at defining an objective evaluation framework for superpixel methods.
%
%trade-off between segmentation accuracy %with respect to image objects
%
%
Concurrently, recent deep learning-based superpixel methods mainly focus on 
the object segmentation task at the expense of regularity.
In this ill-posed context, we show that we can achieve competitive 
results using a recent %computer vision 
architecture like the Segment Anything Model (SAM), 
without dedicated training for the superpixel segmentation task.
This leads to rethinking superpixel segmentation and the necessary properties depending on the targeted downstream task.
 \end{abstract}

% Use if graphical abstract is present
%\begin{graphicalabstract}
%\includegraphics{}
%\end{graphicalabstract}

% Research highlights - Max 85 characters d'après les guidelines
% \begin{highlights}
% \item We detail and discuss the ill-posed nature of the superpixel segmentation problem
% \item We propose the first comprehensive study of the superpixel evaluation framework
% \item We show how to avoid multiple evaluation biases by using a set of recommended metrics
% % \item The lack of proper regularity constraint has lead to very irregular methods
% \item We show how to use high-level segmentation to get accurate and regular superpixels
% %\item We use high-level object segmentation to propose accurate yet regular superpixels
% \end{highlights}

%\nocite{*}

% Keywords
% Each keyword is seperated by \sep
\begin{keywords} 
Superpixels \sep Segmentation \sep Object Segmentation
\end{keywords}

\maketitle

\section{Introduction}

\begin{figure}[t]
\centering
\vspace{0.2cm}
\includegraphics[width=0.4\textwidth,height=0.24\textwidth]{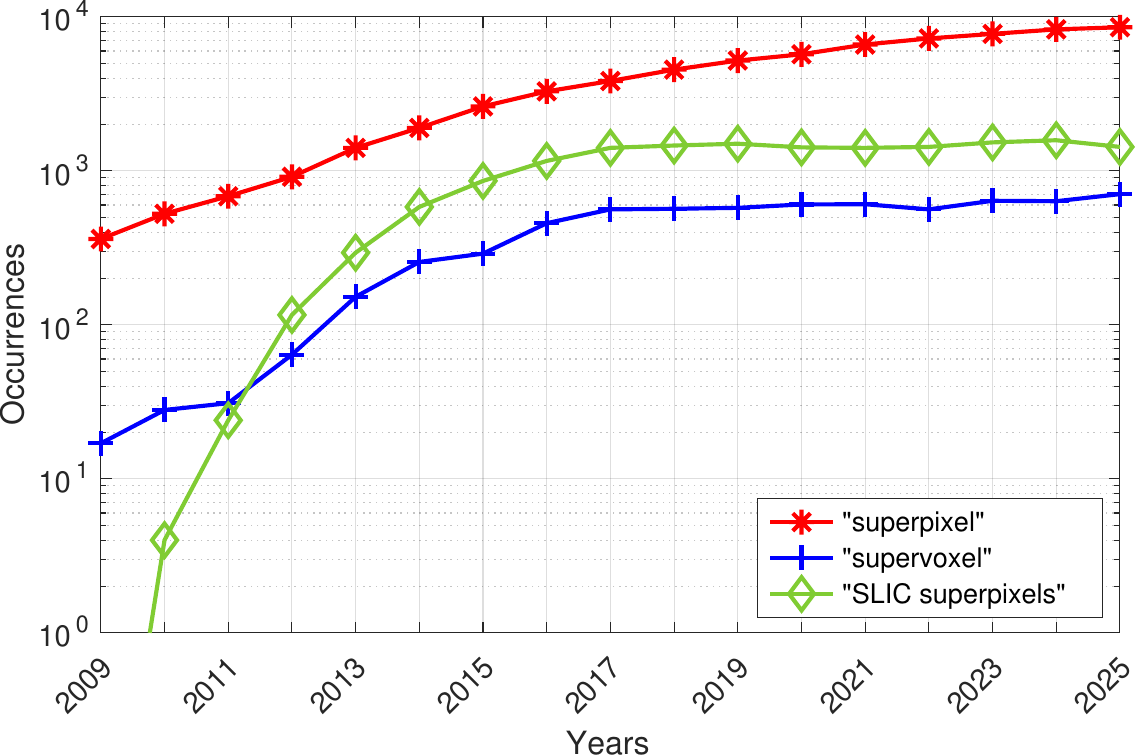} 
\caption{\textbf{Occurrences of superpixel terms in research papers}
(source: Google Scholar). \revision{(updated)}}
\label{fig:occurrences}
\end{figure}

Introduced by \citep{ren2003} and made popular by the SLIC method \citep{achanta2012}, 
superpixel segmentation continues to play a crucial role in the computer vision community,
as illustrated by Figure~\ref{fig:occurrences}, which details the occurrences of related terms in research papers over the years.
\revision{
The use of superpixel segmentation is prevalent across various applications such as 
optical flow \citep{liu2019selflow},
%depth estimation \citep{yang2016efficient},
saliency estimation \citep{wang2015saliency}, %liu2014superpixel,
stereo matching \citep{li2016pmsc},
image segmentation \citep{shen2018submodular}, %dong2015sub
remote sensing image analysis \citep{yin2022slicpolsar,wang2022revisitingslic},
representation learning \citep{ke2023learning,li2024superpixel}.}
\revision{Many methods have been proposed for different image types such as
natural 2D images \citep{achanta2012}, 
videos \citep{chang2013video},
3D medical images \citep{tian2017supervoxel},
and remote sensing~\citep{ma2022taskspecific,ma2022deepsarsp}.}
%or even omnidirectional images \citep{giraud2023generalization}.

Unlike traditional segmentation approaches, 
superpixel methods are initially based on a regularity constraint that enforces regions to have similar sizes, thus following a pseudo-grid structure and resulting in a clearly distinguishable tiling pattern. 
This additional regularity constraint sets superpixel segmentation apart, 
providing a unique approach to partitioning images into meaningful regions.
Such image under representation may be of high interest to reduce the computational burden of
computer vision tasks but also for interactive applications, \emph{e.g.}, \citep{berg2019ilastik}. %long2018efficient

Over the years, numerous methods for superpixel segmentation have been proposed in the literature.
Among these, the SLIC (Simple Linear Iterative Clustering) method \citep{achanta2012} stands out for its popularity, leveraging a local k-means algorithm on simple color and spatial features.
Many extensions of SLIC have been developed, 
incorporating contour constraints \citep{zhang2016,giraud2018_scalp}, 
using other feature spaces \citep{li2015,liu2016manifold}, 
exploring non-iterative variants \citep{achanta2017superpixels},
or computationally more efficient strategies \citep{bobbia2021iterative}.
Other approaches use graph-based algorithms \citep{liu2011,benesova2014,kang2020dynamic}, %,belem2020superpixel}
coarse-to-fine clustering \citep{vandenbergh2012},
watershed \citep{machairas2015}
or propose a hierarchical segmentation, \emph{e.g.}, \citep{wei2018}.

Recently, deep learning methods utilizing convolutional neural networks have emerged \citep{jampani2018superpixel,tu2018learning,yang2020superpixel,zhu2021learning,wang2021ainet,
pan2022fast}, offering the advantage of using higher-level features compared to traditional superpixel methods which rely on pixel- or patch-level features.
With such high-level features, these methods can produce much more accurate superpixel segmentations, also with reduced computational time.
Several studies and reviews have focused on extensive performance
comparisons between these methods %, highlighting the evolving landscape of superpixel segmentation 
\citep{schick2014evaluation,giraud2017_jei,brekhna2017robustness,wang2017superpixel,stutz2018superpixels,barcelos2024comprehensive,wu2024survey}.
We refer the reader to these works for a detailed review of existing superpixels methods.

% \smallskip

\subsection*{An ill-posed problem?}

%\noindent\textbf{An ill-posed problem?}

%
Image segmentation is already considered an ill-posed problem \citep{gonzales1987digital}, 
but this is amplified in the context of over-segmentation into superpixels, 
\revision{where the rules that define superpixels may vary from one work to another.
Particularly, the regularity constraint in superpixel segmentation 
may strongly affect the performance on downstream tasks.
Regularity ensures stable decompositions with easily identifiable components, which is beneficial
for user interaction \citep{berg2019ilastik}, 
%long2018efficient
or for some downstream applications using superpixel neighborhood structures \citep{giraud2017_spm,pan2022fast}.
However, some applications can perform better using irregular (stretched, noisy, of variable sizes) superpixels, \emph{e.g.}, \citep{wilms2021deepfh}.
%{\color{red} add ref ?}, %this constraint 
% superpixel neighborhood-based applications
Generally, enforcing regularity may prevent the accurate segmentation of image objects, 
especially small or thin structures, 
thus these two properties are often viewed as conflicting.}

% that are challenging to capture.
%
Additionally, different methods interpret and respect this regularity constraint to varying degrees, significantly affecting their segmentation performance.
For instance, recent methods choose to generate fewer  superpixels
in low-variance areas, increasing their performance for a given number of requested superpixels \citep{Uziel:ICCV:2019:BASS,kang2020dynamic,xu2022high,zhou2023vine}.
In \citep{xu2022high}, the image is binary segmented to identify low-variance areas in which to produce perfectly regular superpixels.

Therefore, comparisons between methods can be biased, 
and most evaluations of this regularity aspect only rely on qualitative visual assessments. 
As a consequence, recent deep learning-based approaches tend to bypass this constraint, resulting in very irregular regions that are often hard to identify. 
For these reasons, the superpixel segmentation particularly appears as an ill-posed problem.

Moreover, beyond the ill-posed contradictory objective between segmentation accuracy and regularity, and the lack of proper definition of the regularity constraint,
%may be known, 
%\emph{e.g} \citep{giraud2017_jei,stutz2018superpixels}, 
%Beyond the inherent issue induced by 
the validation of superpixel segmentation methods also faces several significant challenges.
Limited comparisons, inadequate or redundant metrics, biased experiments, and  
suboptimal parameters for compared method all contribute to these challenges.
To the best of our knowledge, there is no recent comprehensive study on the validation framework of superpixel methods.

\subsection*{Is it just about object segmentation?}

Most segmentation datasets are usually annotated at a fine level, 
but the segmentation may remain mainly semantic. 
Segmentation performance thus mainly depends on 
%the accurate segmentation of objects and the alignment of superpixels boundaries.
the alignment of superpixels boundaries with those of groundtruth objects.
With the advent of deep learning-based methods,
able to extract object-level features, 
segmentation performance have significantly improved,
but, in turns, methods tend to focus their evaluation on this single aspect.
%evaluation has increasingly focused on object segmentation.
%Nevertheless, recent advances have produced highly 
%effective segmentation methods like SAM, 
%which are nearly semantic-agnostic.
Thus, superpixel segmentation is very close to 
a standard object segmentation problem, but with an implicit, 
rarely and poorly evaluated regularity constraint that greatly impacts
segmentation accuracy.

Low-level feature-based approaches are generally limited but offer better generalizability.
In contrast, deep learning-based methods, while capable of extracting high-level features, may tend to be less generalizable.
Recently, large architectures like \citep{kirillov23sam} have emerged as interesting solutions because they are highly generalizable without the need for retraining.
Such knowledge may be of interest for defining a superpixel segmentation method.

\subsection*{Contributions}

\begin{itemize}
\item We take a step back on superpixel segmentation, showing that it is fundamentally an ill-posed problem (Section \ref{sec:ill_posed}). It is ill-posed because superpixels inherently try to balance capturing colors or textures while maintaining shape regularity, leading to a contradictory objective.
Moreover, this constraint often remains implicit, is rarely or poorly evaluated, which has lead recent deep learning-based methods to generate very irregular superpixel segmentation.

\item We also demonstrate that the superpixel literature suffers from an evaluation problem (Section \ref{sec:metric}).
Recent papers may still use inadequate or redundant metrics, compare to state-of-the-art methods with suboptimal settings, or use biased experiments to establish their superiority.
To address this, we propose the first comprehensive study of the superpixel segmentation validation framework, analyzing the bias in the use of metrics, parameter settings, and validation experiments. 
This study particularly highlights the importance of measuring regularity of superpixel segmentations.

\item Finally, we demonstrate that the problem can be efficiently seen as a standard object segmentation problem (Section \ref{sec:samo}).
To do so, we propose to evaluate an approach that uses a semantic-agnostic segmentation method like SAM (Segment Anything Model \citep{kirillov23sam}) to generate independent object-level proposals, that we merge and then fill with simple regular superpixels based on pixel-level features. % we achieve state-of-the-art performance. 
This way, we only allow high-level object segmentations to be irregular, thus getting the best of both worlds, to generate overall regular and accurate superpixels.
This method, without being trained specifically to generate superpixels, 
achieves state-of-the-art results while maintaining the original regularity constraint of superpixels, contrary to recent deep learning-based methods.
Source code of the method is available here: \url{https://github.com/rgiraud/samo}.

%\item To better assess the capability of superpixel methods, we propose a new evaluation framework and performance representation that considers different levels of regularity for each method.

%\item Finally, we show that a thorough study of recent state-of-the-art methods using our framework allows for establishing a relevant hierarchy and achieving optimal results on proxy tasks.

\end{itemize}

% \section{Related Works}

% \subsection{Superpixel methods}

% SLIC et tous les dérivés

% autres approches watershed etc

% méthodes basées deep

% Question la régularité, ou proposer plus de superpixels dans certaines régions 
% VSSS, SCAC

% \subsection{Superpixel studies}

% présente de manière poussée dans chaque papier de superpixel method

% JEI métrique de régularité et de comparer les méthodes non pas selon le nombre de superpixel (pour lesquels les perfs sont généralezment assez constantes) mais selon les variations de régularité.

% Etudes (stutz, bruit)

\section{Ill-Posed Segmentation Problem\label{sec:ill_posed}}
 
%  \begin{figure}[t]
%   \begin{center}
% \includegraphics[width=0.6\textwidth]{ill_posed.pdf} 
%   \end{center}
%  \end{figure}
%  

\subsection*{Locally constraint segmentation}
\begin{figure}[t!]
\centering
{\footnotesize
 \begin{tabular}{@{\hspace{0mm}}c@{\hspace{1mm}}c@{\hspace{0mm}}}
 \includegraphics[width=0.23\textwidth]{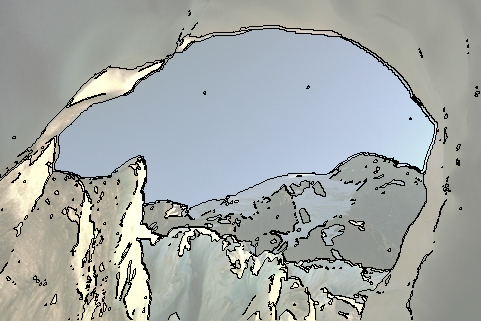}&
\includegraphics[width=0.23\textwidth]{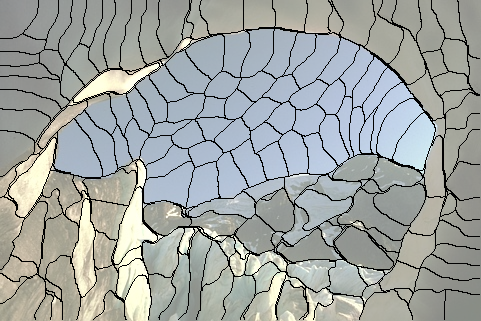}\\
%\multicolumn{1}{@{\hspace{0mm}}c@{\hspace{0mm}}}
{(a) Image segmentation} & 
%\multicolumn{1}{@{\hspace{0mm}}c@{\hspace{0mm}}}
{(b) Superpixel segmentation}\\
{\scriptsize \citep{felzenszwalb2004}} &  {\scriptsize \citep{ren2003}}
\end{tabular}}
  \caption{\textbf{Context of superpixel segmentation.}
  As introduced by \citep{ren2003}, superpixels
  differ from (a) standard image-level segmentation without any spatial constraint except from having connected regions
  by (b) clustering homogeneous pixels into regions 
  having approximately the same size and being clearly identifiable. 
}
\label{fig:context}
\end{figure}

An over-segmentation into superpixels, as introduced by \citep{ren2003}, 
differs from standard image-level segmentation methods by introducing a regularity constraint.
In this way, superpixels approximately have the same size, almost following a pseudo-grid tiling (see Figure~\ref{fig:context}(a)-(b)).
Superpixels have variable shapes but should remain clearly identifiable, 
especially for interactive applications, and able to adapt to any image content.

% although it may impact the quantitative evaluation.
%

  %that may help to capture the image objects.
The regularity paradigm of superpixels impacts their ability to capture image objects and the quantitative evaluation of segmentation methods.
For example, a very strong regularity that produces almost square shapes would poorly capture image objects.
On the contrary, a very irregular method that can fit to the image content 
may be more accurate but would also generate superpixels of highly variable sizes, shapes and with 
noisy borders that less respect the implicit pseudo-grid layout.
This lack of proper constraint definition may be an issue to compare superpixel methods.
%For most methods, superpixels are constrained to not gather pixels that are too far away.

\begin{figure}[t!]
\centering
{\footnotesize
 \begin{tabular}{@{\hspace{0mm}}c@{\hspace{1mm}}c@{\hspace{0mm}}}
\includegraphics[width=0.23\textwidth]{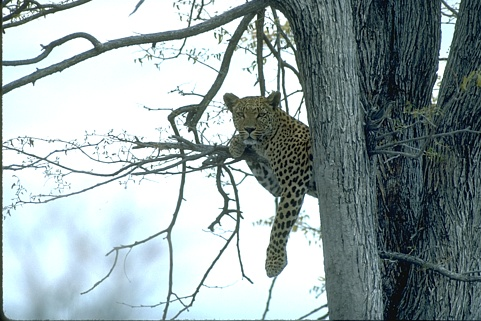}&
\includegraphics[width=0.23\textwidth]{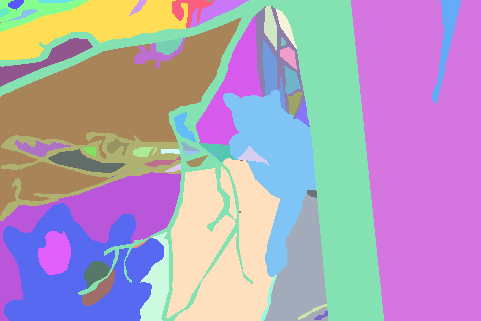}\\
%\multicolumn{1}{@{\hspace{0mm}}c@{\hspace{0mm}}}
{(a) Image } & 
%\multicolumn{1}{@{\hspace{0mm}}c@{\hspace{0mm}}}
{(b) Groundtruth labels}\\
%\multicolumn{2}{@{\hspace{0mm}}c@{\hspace{3mm}}}{(c)  and groundtruth labels containing} \\
%\multicolumn{1}{@{\hspace{0mm}}c@{\hspace{0mm}}}
%{(Felzenswalb et al., 2004) \citep{felzenszwalb2004}} & 
%\multicolumn{1}{@{\hspace{0mm}}c@{\hspace{0mm}}}
%{(Ren et al., 2003) \citep{ren2003}} &
%\multicolumn{2}{@{\hspace{0mm}}c@{\hspace{3mm}}}{thin structures from the BSD \citep{martin2001}}
\end{tabular}}
\caption{\textbf{Thin structures in image segmentation.}
Many natural images present very thin semantic structures.
Hence, the regularity constraint of superpixels may highly impact the ability of the segmentation 
to respect the image objects.
  %in the Berkeley Segmentation Dataset. 
%   Superpixels that are supposed to respet a certain regularity may struggle to respect such object borders.
}
\label{fig:context_thin}
\end{figure}

\subsection*{The issue of thin structures}

\begin{figure}[t]
\centering
  {\footnotesize
  \begin{tabular}{@{\hspace{0mm}}c@{\hspace{3mm}}c@{\hspace{0mm}}}
\includegraphics[width=0.22\textwidth]{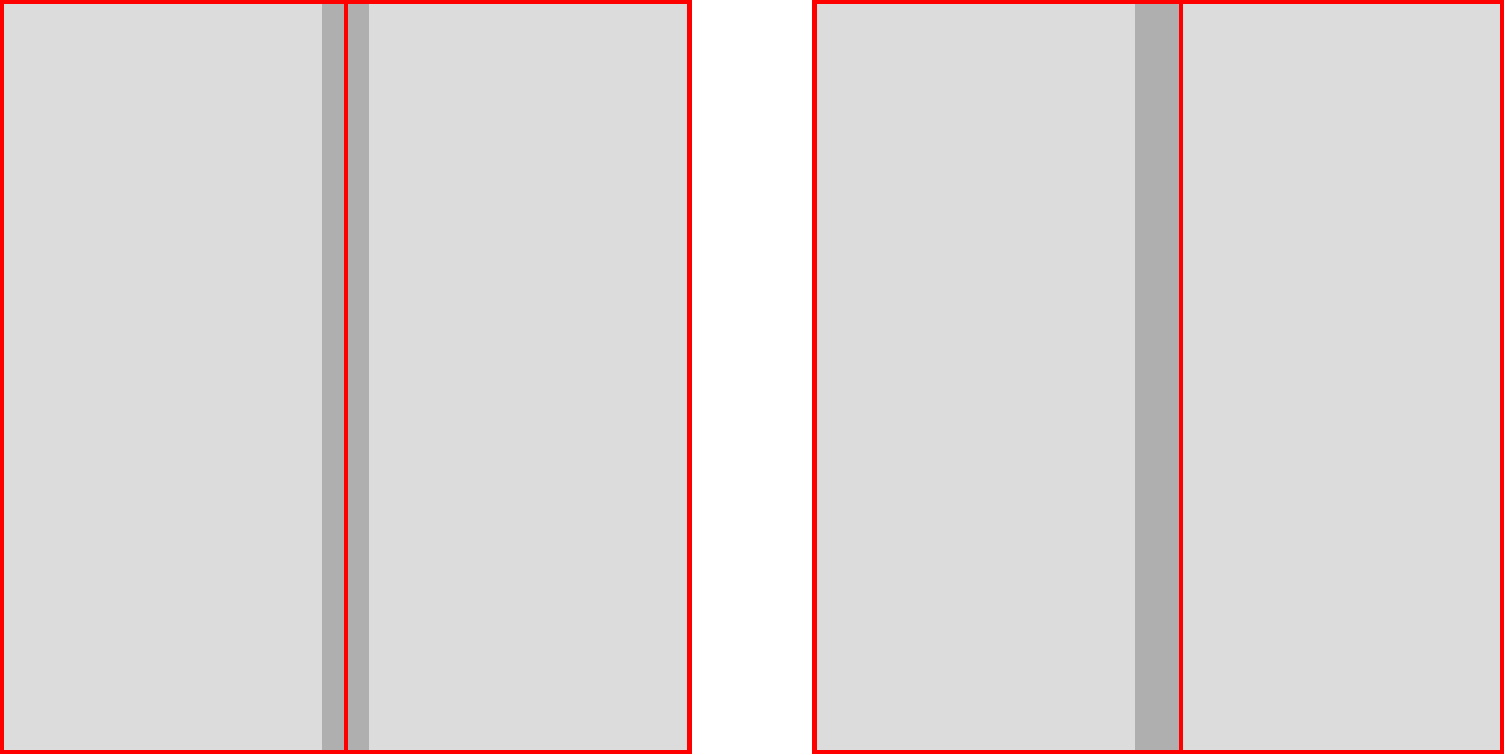} &
\includegraphics[width=0.22\textwidth]{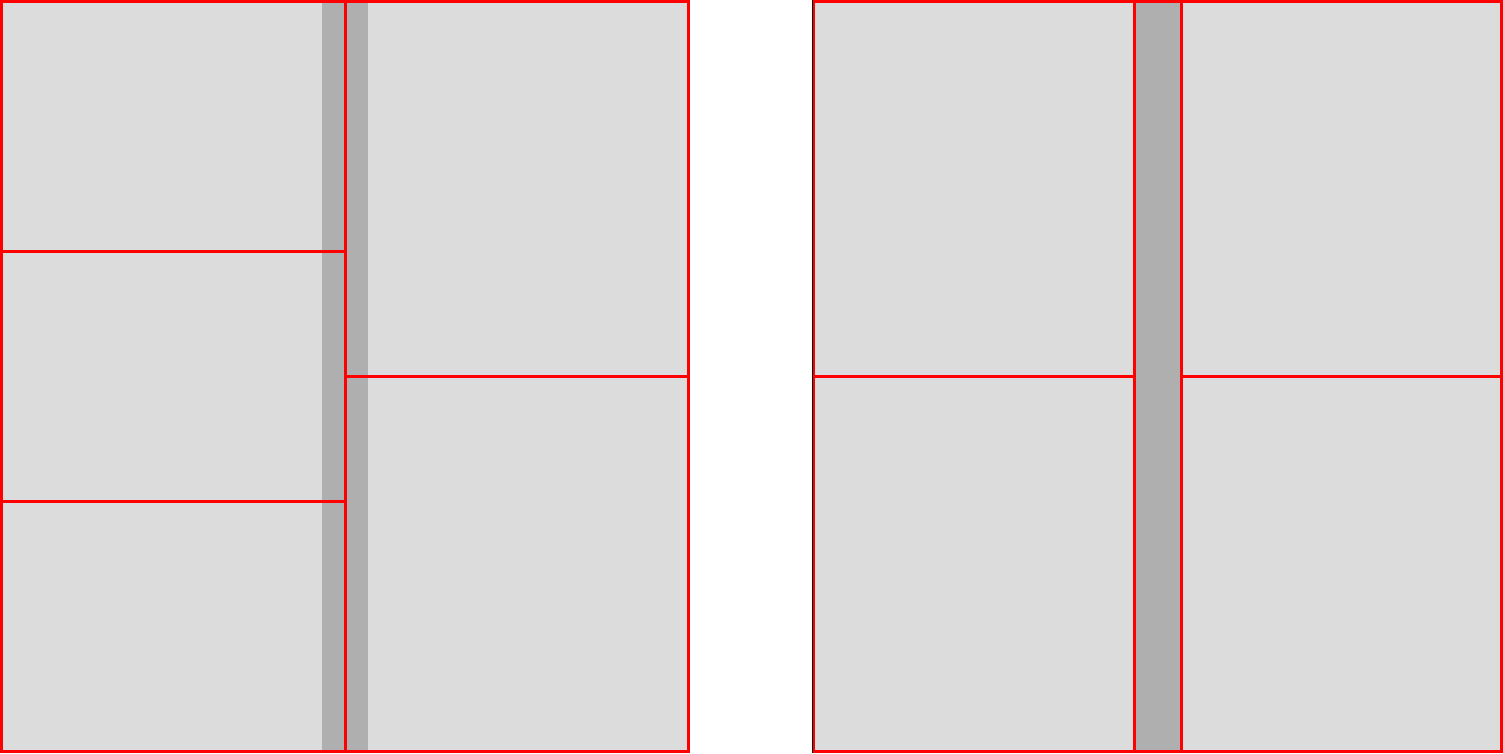} \\
(a) $N_\text{SP}=2$ superpixels & (b) $N_\text{SP}=5$ superpixels \\ %[-1ex]
  \end{tabular}}
  \caption{\textbf{Examples of clustering behaviors on thin structures.}
  (a) Is it better to split the thin dark grey area equally into the two regions or to arbitrary associate it to one region?
  (b) With a sufficient number of superpixels, is it relevant to capture this area at the expense of higher shape irregularity? }
  \label{fig:ill_posed}
 \end{figure}

In Figure~\ref{fig:context_thin}, we represent an image from the reference Berkeley Segmentation Dataset (BSD)~\citep{martin2001}, associated with manual groundtruth segmentations containing thin structures.
This type of very thin objects seems to contradict the definition of superpixels.
In Figure~\ref{fig:ill_posed}, we illustrate different superpixel clustering behaviors on a synthetic image containing a thin structure.
In the two cases, all segmentations seem relevant, although they may lead to very different quantitative performance.
The presence of thin structures exposes the ill-posed nature of the superpixel segmentation.
The limited number of superpixels and their implicit regularity constraint may indeed prevent to accurately capture such regions.
%
%   In Sec. \ref{sec:metrics}, we show that 

\begin{figure*}[t!]
\centering
{\scriptsize
 \begin{tabular}{@{\hspace{0mm}}c@{\hspace{3mm}}c@{\hspace{3mm}}c@{\hspace{3mm}}c@{\hspace{0mm}}}
 \centering
 & ASA = 0.90 | BR = 0.0 | GR = 1.0
 & ASA = 0.90 | BR = 1.0 | GR = 0.50
 & ASA = 0.93 | BR = 1.0 | GR = 0.74
 \\
\includegraphics[width=0.23\textwidth]{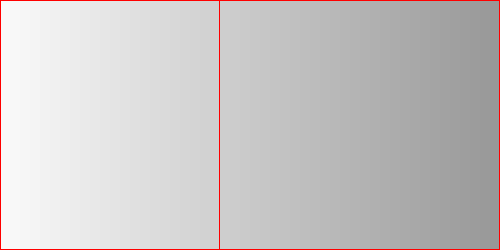}&
\includegraphics[width=0.23\textwidth]{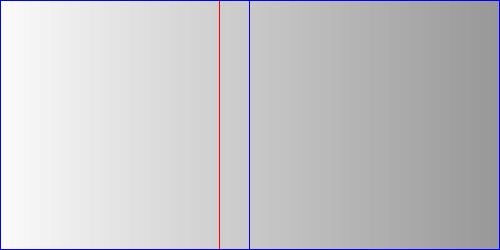}&
\includegraphics[width=0.23\textwidth]{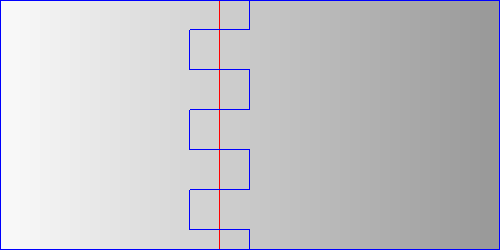}&
\includegraphics[width=0.23\textwidth]{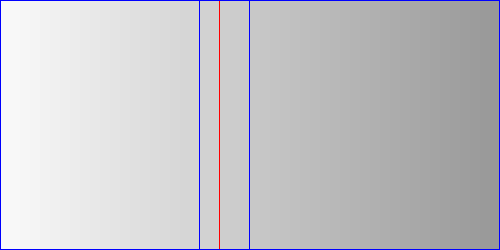}\\ 
(a) & (b) & (c) & (d) \\
\end{tabular}}
\caption{\textbf{Sensitivity of metrics to segmentation accuracy}.
(a) An image with a blurred object border and its ground truth segmentation contours shown in red.
(b), (c), and (d) present three segmentation examples with contours in blue, 
along with the reported object segmentation accuracy (ASA), contour detection (BR) and regularity (GR) metrics (ranging from 0 to 1, with 1 being the maximum). For BR, only the central separation is considered, and a precision of 2 pixels is generally used.
(b) This segmentation can be qualitatively considered as quite accurate and very regular but a slight misalignment in segmentation can result in a complete failure to detect contours (BR = 0.0).
(c) Noisy superpixels can help capture the contours (BR = 1.0), while segmentation accuracy may remain unchanged but regularity logically drops compared to (b). 
(d) A thin superpixel surrounding the uncertain region can improve both metrics: the contours are considered captured, and the segmentation overlap error is reduced.
However, superpixels now have different shapes and sizes and the regularity drops compared to (b). % are far to have the same shape and size
These examples illustrates the interest for superpixel methods to bypass the regularity constraint to improve segmentation accuracy metrics.}
\label{fig:asa_br}
\end{figure*}

\subsection*{Segmentation accuracy VS regularity}

There are several evaluation issues with the standard metrics, 
which will be discussed in details in Section \ref{sec:metric}.
Notably, threshold effects in some metrics and lack of regularity evaluation have influenced the visual results of certain methods, particularly deep learning-based ones.
The mainly used metrics evaluate segmentation accuracy:
ASA (Achievable Segmentation Accuracy) \citep{liu2011}
and BR (Boundary Recall) \citep{martin2004}, detailed in Sections~\ref{subsubsec:asa} and~\ref{subsubsec:br}, respectively.
ASA measures the overlap area between superpixels and ground truth objects, while BR evaluates the capture of object contours based on a given threshold.

In Figure~\ref{fig:asa_br}, we illustrate the results of these metrics on a synthetic segmentation example. 
We also report the rarely assessed measure of regularity using GR \citep{giraud2017_jei}, which evaluates how closely superpixels resemble smooth squares.
It can be observed that seeking to optimize solely for segmentation accuracy metrics may lead to produce irregular segmentation boundaries (Figure~\ref{fig:asa_br}(c)) and to the emergence of thin superpixels along object edges (Figure~\ref{fig:asa_br}(d)).
This behavior is found in recent deep learning-based methods
\emph{e.g.}, 
\citep{zhu2021learning,wang2021ainet}.
%These methods 
%do not strongly enforce the regularity constraint 
%so their results can severely change according to the used parameter.
%   
%However, recent deep learning-based methods that obtain 
as illustrated in Figure~\ref{fig:diff_regu}(a).
We can see that these methods then to bypass
the regularity constraint and produce very irregular segmentations 
with noisy and thin superpixels around the object borders.
In their official implementations, 
disconnected or one pixel large regions may even be generated.
As shown in Figure \ref{fig:asa_br}, this is the consequence of seeking to maximize 
ASA and BR-like metrics, sometimes directly used as training losses.

\subsection*{Fair comparison between superpixel methods}

\begin{figure*}[t!]
\centering
{\footnotesize
 \begin{tabular}{@{\hspace{0mm}}c@{\hspace{1mm}}c@{\hspace{2mm}}c@{\hspace{1mm}}c@{\hspace{1mm}}@{\hspace{0mm}}}
 \centering
{\scriptsize GR = 0.21} &
{\scriptsize GR = 0.29} &
{\scriptsize GR = 0.27} &
{\scriptsize GR = 0.67} \\
\includegraphics[width=0.235\textwidth]{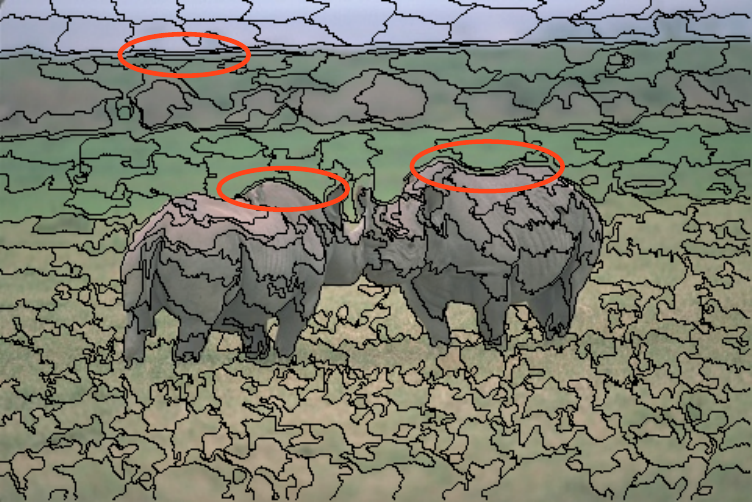}&
\includegraphics[width=0.235\textwidth]{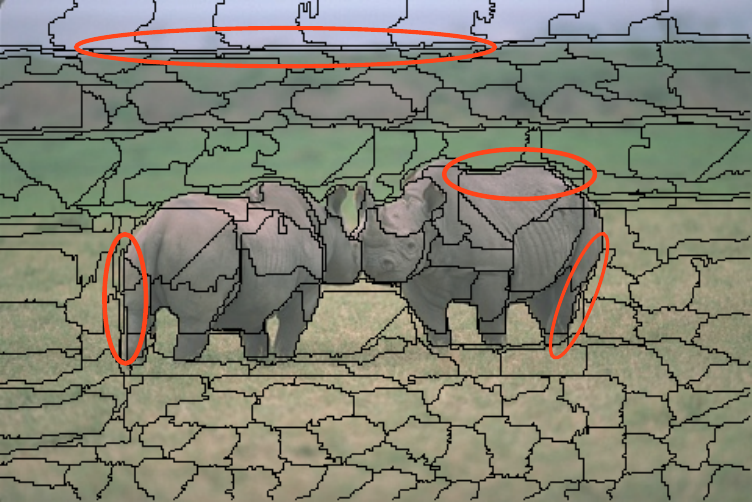}&
\includegraphics[width=0.235\textwidth]{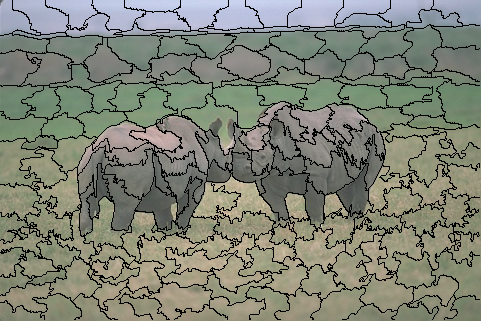}&
\includegraphics[width=0.235\textwidth]{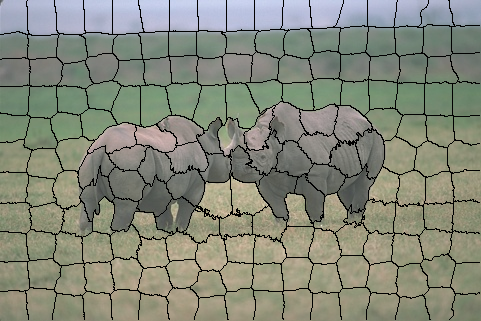}\\
\multicolumn{2}{@{\hspace{0mm}}c@{\hspace{0mm}}}{(a) Deep learning-based superpixel segmentation using} &
\multicolumn{2}{@{\hspace{0mm}}c@{\hspace{3mm}}}{(b) Superpixel segmentations with different regularity} \\ 
LNS-Net \citep{zhu2021learning} & AINet \citep{wang2021ainet} & %[-1ex]
\multicolumn{2}{@{\hspace{0mm}}c@{\hspace{0mm}}}{settings using SLIC \citep{achanta2012}}
\end{tabular}}
\caption{\textbf{Variability in the regularity constraint.}
(a) Recent deep learning-based methods tend to bypass the regularity constraint and initial pseudo square grid paradigm to produce very irregular segmentations, with noisy and thin superpixels at object borders in order to capture uncertainty regions (see red ellipses).
%
%This may lead to significant difference in the ability to accurately segment the image objects.
%
(b) Most traditional methods, including %the most popular 
SLIC \citep{achanta2012}, 
tend to more follow the layout of a regular grid and generate easily identifiable superpixels. Moreover, they can produce superpixels of different regularity levels, %according to the regularity setting, 
leading to potential bias in their evaluation of performance when used with suboptimal settings.
% as for many state-of-the-art methods.
% in the Berkeley Segmentation Dataset. 
%   Superpixels that are supposed to respet a certain regularity may struggle to respect such object borders.
} \vspace{-0.1cm}
\label{fig:diff_regu}
\end{figure*}

Contrarily to deep learning-based methods, most traditional methods, 
\emph{e.g.}, \citep{achanta2012,chen2017,achanta2017superpixels,wu2020fuzzy},
%including %the most popular SLIC \citep{achanta2012}, 
are capable of releasing or strengthening  their regularity constraint (Figure~\ref{fig:diff_regu}(b)) and tend to more follow the pseudo-grid layout.

Therefore, since regularity may have a strong impact on performance,
 having different regularity constraints for each method lead to significant bias in the quantitative evaluation of methods.
Although the regularity could be evaluated to put into perspective the performance, such results are rarely reported (see Section \ref{sec:metric}).
More broadly, the ill-posed nature of superpixel segmentation is highlighted by the often biased evaluation of its performance.
%also on aspects non-related to the regularity.

\smallskip 
 
In the following, we carry a thorough study of the mainly used superpixel metrics
in the literature, and the potential bias in the use of some metrics.
We illustrate why and how some aspects should be relevantly evaluated to ensure a minimal and fair comparison of methods in the literature. 
%and we encourage the research community to adopt these recommendations.
%\emph{i.e.}, using at least this set of metrics.

% \begin{figure}[h!]
% \centering
% {\footnotesize
%  \begin{tabular}{@{\hspace{0mm}}c@{\hspace{1mm}}c@{\hspace{3mm}}c@{\hspace{1mm}}c@{\hspace{1mm}}@{\hspace{0mm}}}
% \includegraphics[width=0.24\textwidth]{seg_ex_img1.png}&
% \includegraphics[width=0.24\textwidth]{seg_ex_gt1.png}&
% \includegraphics[width=0.24\textwidth]{seg_ex_img2.png}&
% \includegraphics[width=0.24\textwidth]{seg_ex_gt2.png}\\ 
% Image &  Groundtruth labels &  Image &  Groundtruth labels \\
% \end{tabular}}
%   \caption{\textbf{Example of thin structures} in the Berkeley Segmentation Dataset. 
%   Superpixels that are supposed to respet a certain regularity may struggle to respect such object borders.
% }
%   \label{fig:br}
%   \end{figure}
%   
%   felzenswalb - ren   puis   slic c faible slic c fort
%   fine d'un coté, normal de l'autre
%   

% \begin{itemize}
% % \item Impossible trade-off between accuracy and regularity
% %  for instance Thin structures - utiliser (hierarchique) image seg sur beuacoup d'images naturelles hors dataset on a des structures de ce genre
% %  \item What to do on textures
% %  \item Need for object level info
% %\item  Texture. Plusieurs échelles sur des rayures. multi-scale or hiérarchical approaches
% %\item  color/texture feature vs object boundaries. Need for object level information -> deep-larning based methods.
% \item  Compromis impossible entre régularité et fines structures. ?
% tentative classif
 
% \end{itemize}

\section{Relevant Superpixel Validation Framework\label{sec:metric}}

In this section, we propose the first comprehensive study of the superpixel segmentation validation framework, analyzing the bias in the use of metrics, parameter settings and experiments. 
We show that the literature may suffer from insufficient or flawed evaluation, making it difficult to assess the performance. % of the methods.

In Table~\ref{tab:metrics}, we propose a review of the mainly used metrics and performed experiments in %significant papers of 
the superpixel literature.
We classify the metrics into 4 categories that are mandatory to evaluate to ensure a fair comparison of methods on their ability:
(i) to segment the image objects and 
(ii) to capture the object contours with respect to a groundtruth segmentation;
(iii) to generate regularly shaped superpixels;
(iv) to gather pixels with similar colors.
We also report when processing time and robustness to noise are evaluated.
Finally, performance are always evaluated for several number of superpixels ($N_\text{SP}$) and rarely according to different regularity levels.
%
%Finally, we also report when superpixel methods are validated on proxy tasks.

\begin{table*}[t]
\caption{Review of used metrics and experiments in the literature of superpixel segmentation methods (including supplementary).
\revision{Italics indicate studies or papers introducing metrics.}
\revision{Orange and green colors respectively indicate higher and lower levels of bias in metrics.
Green metrics should be prioritized over orange ones.
In the case of EV (orange/green), it is less biased than ICV but more than SIRS.}
Orange dots indicate a biased report of performance according to the number of superpixels.
Gray cells indicate the introduction of the metric.
\revision{For the robustness to noise, B, GN, IN respectively indicate tests
on Blur, Gaussian Noise and Impulse Noise.}
See text for details.} %\vspace{-0.4cm}
{\scriptsize
\renewcommand{\arraystretch}{1}
\begin{center}
    \setlength{\tabcolsep}{3.5pt}
\begin{tabular}{|l@{\hspace{0.7mm}}|@{\hspace{0.2mm}}c@{\hspace{0.2mm}}|c|c|c|c|c|c|c|c|c|@{\hspace{0mm}}c@{\hspace{0mm}}|c|c|c|c|c|}
\hline
\textbf{} & \textbf{} &
\multicolumn{3}{c|}{Object seg.} & \multicolumn{3}{c|}{Contour detection} &\multicolumn{2}{c|}{Regularity}  &\multicolumn{3}{c|}{Color homogeneity} &
%\cellcolor{orange!25}
Pro.& %\cellcolor{orange!25}
Rob. & \multirow{2}{*}{{/N$_\text{SP}$}} % \cellcolor{green!25}}
& %\cellcolor{green!25}
\multirow{2}{*}{/Regu.}\\[0ex] \cline{3-12}
\textbf{Method/\textit{Study}} & \textbf{Venue}   &
\cellcolor{orange!25}UE & \cellcolor{green!25}CUE\hspace{0.0cm} & \cellcolor{green!25}\hspace{-0.235cm}$\approx$ ASA & \cellcolor{orange!25}BR & \cellcolor{green!25}/P\text{ } & \cellcolor{green!25}/CD & \cellcolor{orange!25}CO & \cellcolor{green!25}GR & \cellcolor{orange!25}ICV &
\hspace{-0mm}\colorbox{orange!25}{\hspace{1.5mm}E}\hspace{-1mm}\colorbox{green!25}{\hspace{-1mm}V \text{ } \text{ } }\hspace{-0.25cm}\text{ }&
\cellcolor{green!25}\revision{SIRS}  &
%\cellcolor{orange!25}
time &
%\cellcolor{orange!25}
noise& %{\cellcolor{green!25}}
&  %\cellcolor{green!25}
\\  \hline
\noalign{\global\arrayrulewidth=0.1mm}
  \arrayrulecolor{lightgray}
ERS {\tiny \citep{liu2011}} & CVPR'11 &\cm & & \cellcolor{gray!25}\cm & \cm & &  & & & & & &  & &\rcm &\\  \hline
SLIC {\tiny \citep{achanta2012}} & TPAMI'12   & \cm  & & & \cm  & & & & &  &&&\cm&&\cm & \\
 \hline
 SEEDS {\tiny \citep{vandenbergh2012}} & ECCV'12   &\cm & \cellcolor{gray!25}\cm &\cm &\cm  & \cellcolor{gray!25}\cm && &  & & &&\cm&&\cm&\\  \hline
 \textit{Schick et al.} {\tiny \citep{schick2014evaluation}} & PRL'14 & &  &  & & & &  \cellcolor{gray!25}\cm& & & & &\cm & &\cm  & \cm \\  \hline
 MSS {\tiny \citep{benesova2014}} & MVML'14 & & \cm & & \cm & &  & & &\cellcolor{gray!25}\cm && &\cm & &\cm &\\  \hline
LSC {\tiny \citep{li2015}} & CVPR'15 &   \cm & & \cm & \cm &  & & &  & & &&\cm&&\rcm& \\  \hline
WP {\tiny \citep{machairas2015}} & TIP'15 & & & & & &\cellcolor{gray!25}\cm  & & & & && \cm &&\cm &\cm\\ \hline
%SCALP {\tiny \citep{giraud2016_scalp}} & ICPR'16    & & &  & \cm  & & \cm & \cm & & \cm & \cm & & \cm &  \\\hline
MSLIC {\tiny \citep{liu2016manifold}} & CVPR'16 & \cm & & \cm & \cm & & & & & & & &\cm& & \rcm &  \\ \hline %cited 156
SLICB {\tiny \citep{zhang2016}} &  TCSVT'17 &  \cm &  & \cm & \cm & & \cm & \cm & & & && \cm & & \rcm  & \cm  \\ \hline
\textit{Giraud et al.} {\tiny \citep{giraud2017_jei}} & JEI'17 &  & \cm &\cm & &\cm &\cm &\cm & \cellcolor{gray!25}\cm  & \cm& \cm & && & \cm &\cm   \\ \hline
\textit{Brekhna et al.} {\tiny \citep{brekhna2017robustness}} & JEI'17 & \cm & &\cm  & \cm & & &\cm & & &  &&\cm& \revision{\scriptsize B,GN,IN} &\rcm &  \\ \hline
\textit{Wang et al.} {\tiny \citep{wang2017superpixel}} & SP:IC'17 & \cm & & \cm & &  &\cm & & & &  &&\cm & &\rcm &  \\ \hline
SNIC {\tiny \citep{achanta2017superpixels}}& CVPR'17 &   &\cm &  & &  \cm && &   & &&&&&\cm& \\  \hline
\textit{Stutz et al.} {\tiny \citep{stutz2018superpixels}} & CVIU'18 & \cm &\cm &\cm &\cm & & &\cm &  &\cm & \cm && \cm & \revision{\scriptsize B,IN }& \cm &\\ \hline
%\rowcolor{gray} MBS {\tiny \citep{hu2018minimum}} & IVC'18 & & & & &  & & &  & &&&&&  \\ \hline
SH {\tiny \citep{wei2018}} & TIP'18   & &  \cm & \cm & \cm &  & & &  & & & &\cm & &\rcm & \\ \hline
GMMSP  {\tiny \citep{Ban18}} & JRTIP'18  & \cm &  & \cm& \cm& &  & &   & && &\cm &&\cm&\cm\\ \hline
SCALP {\tiny \citep{giraud2018_scalp}} & CVIU'18     &   & &  \cm & & \cm & \cm &  & \cm & &&&\cm  & \revision{\scriptsize GN} &  \cm &\\\hline
CAS {\tiny \citep{xiao2018content}} &TIP'18  & \cm   & & \cm  &\cm  &   & & & & & && \cm  & & \rcm&\\ \hline
SSN {\tiny \citep{jampani2018superpixel}}&  ECCV'18   & & & \cm &\cm &   \cm& & & & &&&\cm&&\rcm&  \\ \hline
SEAL {\tiny \citep{tu2018learning}} & CVPR'18 &  & & \cm & \cm & &&&&& &&&&\rcm&\\ \hline
%TASP {\tiny \citep{giraud2019texture}} & ICIP'19 &  & & & & & & &  &  & & & &  \\
ISF {\tiny \citep{vargas2019iterative}} & TIP'19 & &\cm &&\cm & & & & & &&&\cm & & \cm&\\ \hline
qd-CSS {\tiny \citep{ye2019fast}} &ICCV'19  & \cm   & & & \cm & & &\cm &  &&&& \cm &  & \rcm &\\ \hline
BASS {\tiny \citep{Uziel:ICCV:2019:BASS}} & ICCV'19 &  ? & ? & & \cm & & &  & & & \cm & && & \cm &  \\ \hline
DRW {\tiny \citep{kang2020dynamic}} &TIP'20 &  &\cm & &\cm &  \cm& & &  & & && \cm & & \cm &  \\ \hline
WSGL {\tiny \citep{yuan2020watershed}} &TIP'20  &  &  \cm & \cm & \cm & & & \cm &  & & \cm &&  \cm & &\cm & \cm   \\ \hline %comp with CO
FSLIC {\tiny \citep{wu2020fuzzy}} &TCSVT'20 &  & \cm & & && \cm  & &\cm  & & &&\cm & \revision{\scriptsize GN,IN} & \rcm&   \\ \hline
\revision{RSS} {\tiny \citep{chai2020rooted}} &IJCV'20 &  & \cm &  &\cm &  & & \cm &  &&\cm  &&\cm  &&\cm&  \\ \hline
SFCN {\tiny \citep{yang2020superpixel}} &CVPR'20 &  & & \cm & &  \cm& & \cm &  & && &\cm  &&\cm&  \\ \hline
%DISF {\tiny \citep{belem2020superpixel}} &  SPL'20&  & \cm & &\cm & &  & & & &  && &\cm &  \\ \hline
%\rowcolor{gray}ECCPD {\tiny \citep{ma2020convex}} &TIP'20  &   & & & & & & &  & & & & & &  \\ % suit pas les real NSP
\revision{ODISF} {\tiny \citep{belem2021towards}} & SIBGRAPI'21 & &\cm &&\cm & & & & & &&&\cm & & \cm&\\ \hline
\revision{IBIS} {\tiny \citep{bobbia2021iterative}} & Access'21  &  ? & ? & \cm & \cm & & & \cm &  & &&&\cm &  &\rcm &  \\ \hline
\revision{SIN} {\tiny \citep{yuan2021sin}} &PRICAI'21 &  &  &\cm  & &\cm  & & \cm &  & & &&\cm  &&\cm&  \\ \hline
LNS-Net {\tiny \citep{zhu2021learning}} & CVPR'21 &  & & \cm &\cm & & & &   & &&&\cm & & \rcm&  \\ \hline
AINet {\tiny \citep{wang2021ainet}} &ICCV'21 & & &\cm & & \cm & & &  & &&& \cm  & & \cm &  \\ \hline
%\rowcolor{gray} SCAC {\tiny \citep{yuan2021superpixels}} & TIP'21 &  & & & & && & &  & & & & &   \\
FLS {\tiny \citep{pan2022fast}} & TIP'22 & &&\cm & & &\cm & \cm &\cm & &&&\cm & & \cm&\cm\\ \hline %cite 3
HERS {\tiny \citep{peng2022hers}} & WACV'22 &  & & \cm & \cm & & & &   & &\cm && \cm & &\rcm  &  \\ \hline
HQSGRD {\tiny \citep{xu2022high}} & TCSVT'22 & \cm  & &\cm &\cm & & &  &  & &\cm & &\cm& & \rcm& \\ \hline
\revision{SICLE} {\tiny \citep{belem2022fast}} & DGMM'22 &   & \cm & &\cm & & &  &  &  & &&& & \cm& \\ \hline
\revision{CRTrees} {\tiny \citep{yan2022hierarchical}} & TIP'22 &   & \cm & &\cm & & & &&&\cm  & &\cm & & \cm& \\ \hline
\revision{\textit{SIRS}} {\tiny \citep{barcelos2022improving}} & TIP'22 &   & & & & & & &&  & \cm & \cellcolor{gray!25}\revision{\cm}& & &\cm & \\ \hline
%: A comprehensive review and new taxonomy on superpixel segmentation.
%\rowcolor{gray} VSSS {\tiny \citep{zhou2023vine}} & TIP'23 &  & & & & & & &  & & & & & &  \\ \hline
FSA {\tiny \citep{ng2023fuzzy}} & PR'23 & & \cm & \cm & & & \cm & \cm & & & &&\cm & & ?  &\\ \hline
%ESNet {\tiny \citep{xu2023esnet}} & TCSVT'23 & & &&&&&&&&\\ \hline
VSSS {\tiny \citep{zhou2023vine}} & TIP'23 & & \cm &\cm& \cm &&&\cm&&&\cm&&\cm & &\cm&\\ \hline
\textit{Barcelos et al.} {\tiny \citep{barcelos2024comprehensive}} & ACM'24 &  &\cm & &\cm & & & \cm & &&\cm   & \revision{\cm}&\cm & \revision{\scriptsize B,IN}  & \cm &  \\ \hline
\textit{Wu et al.} {\tiny \citep{wu2024survey}} & TechRxiv'24 & \cm & & & & & \cm & & \cm & & && \cm & \revision{\scriptsize GN,IN} &\cm&  \\  \hline
\revision{BINet} {\tiny \citep{zhao2024rethinking}} & ASC'24 & & & \cm &  &\cm &&\cm&& &&& & &\cm&\\ \hline
\revision{CDS} {\tiny \citep{xu2024learning}} & AAAI'24 & &\cm & \cm &  &\cm &&&& &&&\cm & &\cm&\\ \hline
\revision{HHTS} {\tiny \citep{chang2024hierarchical}} & CVPR'24 &?&? & \cm & \cm &\cm &&\cm && \cm&\cm&&\cm & &\cm&\\ \hline
\revision{RSCE} {\tiny \citep{yang2025regular}} & SP'25 & & & \cm & \cm & &&\cm&&\cm &&&\cm &&\rcm&\\ \hline
\revision{SIT-HSS} {\tiny \citep{xie2025hierarchical}} & SDM'25 & &\cm & \cm & \cm & &&&&&\cm &&\cm & &\rcm&\\ \hline
\noalign{\global\arrayrulewidth=0.1mm}
  \arrayrulecolor{black}
  \hline
%\\[1ex] % \hline
%\textbf{Study} &  & & & & & & & &  &  & & & && \\ \hline
%Chen et al. {\tiny \citep{chen2020review}} & ICISCAE'20 &  & & & & & & &  & & &  & &  \\ %12
 \end{tabular}
\end{center}
%\vspace{-0.5cm}
  }
\label{tab:metrics}
\end{table*}

\subsection{Superpixel metrics\label{subsec:metrics}}

\subsubsection{\label{subsubsec:asa}Object segmentation} 

The main property used to evaluate the accuracy of superpixel segmentation is 
the respect of image objects usually obtained from a groundtruth segmentation.
Several metrics can be used to answer the question: are the generated superpixels well contained into the image objects?
The Undersegmentation Error (UE) proposed in \citep{levinshtein2009} 
compares the groundtruth segments to all overlapping superpixels but appears to be very sensitive to small overlaps \citep{stutz2018superpixels}.
In \citep{achanta2012}, a tolerance threshold of 5\% of the superpixel shape is introduced but the metric remains unstable \citep{vandenbergh2012,stutz2018superpixels}.
Nevertheless, some works choose this formulation over more recent ones.

In \citep{vandenbergh2012,neubert2012}, Corrected UE (CUE) metrics have been proposed to 
measure the leakage of superpixels outside their largest overlap with a groundtruth object.
\revision{This metric closely matches the Achievable Segmentation Accuracy (ASA) \citep{liu2011} as they has been proven to be strongly proportional \citep{stutz2018superpixels,giraud2017_jei}, to the point that the error between the two metrics (ASA and 1-CUE) is negligible.
Therefore, reporting the two metrics jointly is quite redundant, so other aspects should be prioritized.
}
ASA is defined for a superpixel segmentation $\SSS=\{S_k\}$ of an image $I$ containing $|I|$ pixels, and  a set of groundtruth objects $\GG=\{G_j\}$ as: \vspace{-0.15cm}

{\eqsize
\begin{equation}
%ASA(\SSS,\GG) = \frac{\sum_{k}\max_i|S_k\cap G_j|}{\sum_{i}|G_j|}.  \label{asa}
%\text{ASA}(\SSS,\GG) = \frac{1}{\sum_{S_k}|S_k|}\sum\limits_{S_k}\underset{G_j}{\max}|S_k\cap G_j|.  \label{asa}
\text{ASA}(\SSS,\GG) = \frac{1}{|I|}\sum_{S_k}\underset{G_j}{\max}|S_k\cap G_j|.  \label{asa}
\end{equation}
}

% In \citep{liu2011}, the Achievable Segmentation Accuracy
% Mais dans \citep{giraud2017_jei}, la métrique a été démontrée comme extrêmement corrélée à la métrique ASA \citep{} sous la relation UE$\approx$(1-ASA)/2.
%
Table~\ref{tab:metrics} shows that many methods still report both CUE and ASA metrics,
although they almost contain the exact same information.
To evaluate the respect of image objects, using CUE or ASA seems sufficient.

\begin{figure*}[t!]
\centering
{\footnotesize
 \begin{tabular}{@{\hspace{0mm}}c@{\hspace{1mm}}c@{\hspace{1mm}}c@{\hspace{1mm}}c@{\hspace{1mm}}@{\hspace{0mm}}}
&&ASA=0.974 \text{ } {\color{darkred}BR=0.865} &
 ASA=0.974 \text{ } {\color{darkgreen}BR=0.927}  \\ 
&&{\color{darkgreen}P=0.145 \text{ } CD=0.186}   &
{\color{darkred}P=0.117 \text{ } CD=0.255} \\ 
\includegraphics[width=0.24\textwidth]{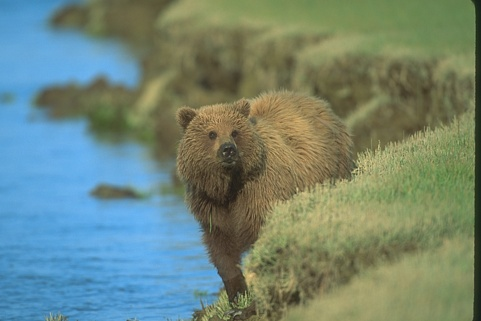}&
\includegraphics[width=0.24\textwidth]{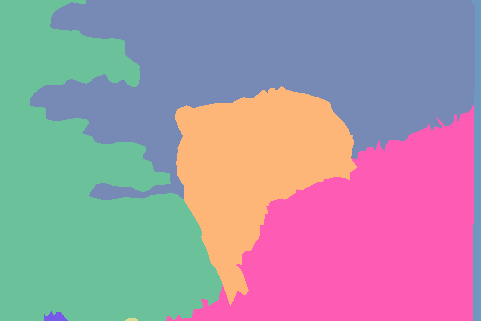}&
\includegraphics[width=0.24\textwidth]{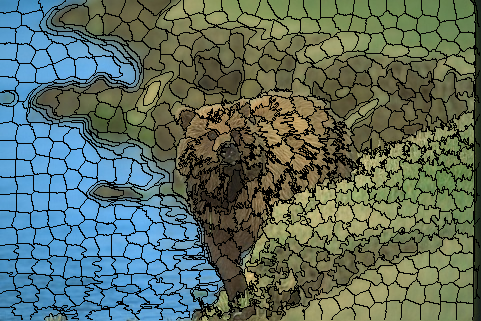}&
\includegraphics[width=0.24\textwidth]{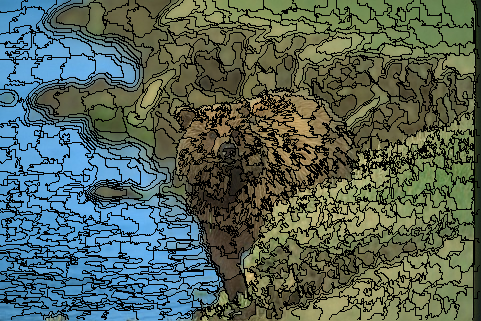}\\
(a) Image & (b) Groundtruth labels & (c) Regular seg \citep{li2015} & (d) Irregular seg \citep{li2015}\\ %[-1ex]
\end{tabular}}
\caption{
\textbf{Illustration of contour detection metrics.} 
The segmentation examples (c) and (d) have the same segmentation performance (ASA), 
but noisy contours (d) help to obtain higher Boundary Recall (BR).
Precision (P) or Contour Density (CD) should be used to express irregularity or 
noise in the segmentation.
} 
\label{fig:br}
\end{figure*}

\begin{figure*}[t!]
\centering
{\footnotesize
 \begin{tabular}{@{\hspace{0mm}}c@{\hspace{1.5mm}}c@{\hspace{1.5mm}}c@{\hspace{1.5mm}}c@{\hspace{1.5mm}}c@{\hspace{0mm}}}
% {\color{black}CO=0.864} \hspace{0.02mm} {\color{black}GR=1.000} &
% {\color{black}CO=0.939} \hspace{0.02mm} {\color{black}GR=0.863}  &
%  {\color{black}CO=0.386} \hspace{0.02mm} {\color{black}GR=0.494} &
%   {\color{black}CO=0.405} \hspace{0.02mm} {\color{black}GR=0.350} &
%   {\color{black}CO=0.782} \hspace{0.02mm} {\color{black}GR=0.553} \\ 
{\color{black}CO=0.86} \hspace{0.02mm} {\color{black}GR=1.00} &
{\color{black}CO=0.94} \hspace{0.02mm} {\color{black}GR=0.86}  &
 {\color{black}CO=0.39} \hspace{0.02mm} {\color{black}GR=0.49} &
  {\color{black}CO=0.41} \hspace{0.02mm} {\color{black}GR=0.35} &
  {\color{black}CO=0.87}   {\color{black}GR=0.48} \\ 
\includegraphics[width=0.2\textwidth,height=0.1425\textwidth]{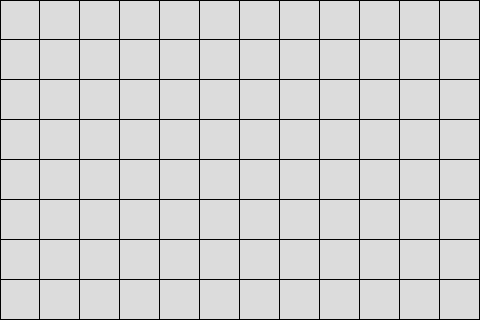}&
\includegraphics[width=0.2\textwidth,height=0.1425\textwidth]{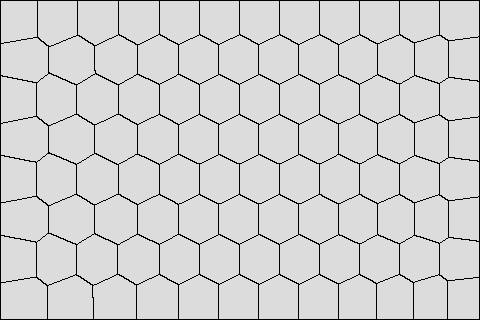}&
\includegraphics[width=0.2\textwidth,height=0.1425\textwidth]{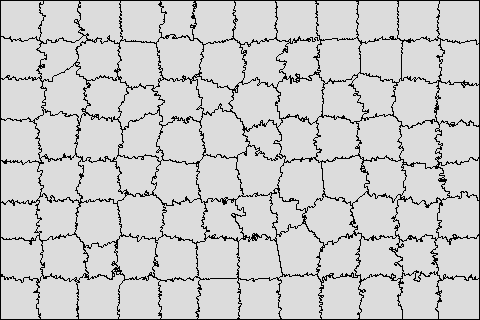}&
\includegraphics[width=0.2\textwidth,height=0.1425\textwidth]{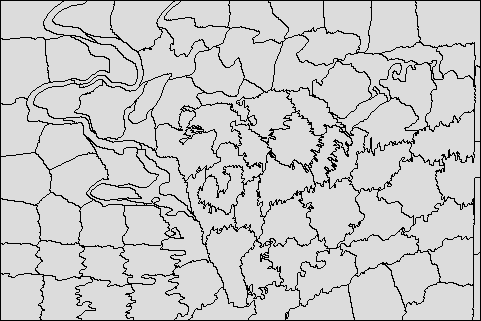}&
\includegraphics[width=0.1425\textwidth,height=0.1425\textwidth]{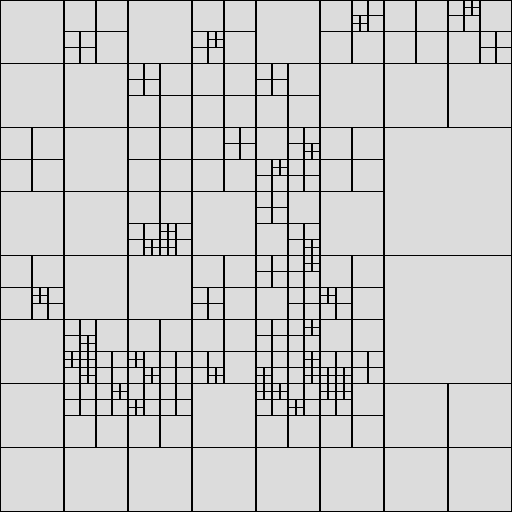}\\
(a) Square shapes & (b) Hexagonal shapes & (c)  Noisy squares & (d) Natural seg. & (e) Quadtree \\ %[-1ex]
\end{tabular} }
\caption{
\textbf{Illustration of the regularity metrics.} 
(a) CO does not provides the highest measure for square shapes contrary to GR.
(b) CO metric is higher with more circular shape like hexagons although the borders are not consistent with the core of the segmentation.
CO is largely impacted by noise, so that a slightly noisy square decomposition (c) may have lower CO than a natural irregular segmentation (d).
(e) GR only is able to penalize the shape inconsistency through the decomposition.
} \vspace{-0.2cm}
\label{fig:gr}
\end{figure*}

\subsubsection{\label{subsubsec:br}Contour detection}

The respect of groundtruth objects may also be computed 
by focusing on their contours.
Results may differ from object segmentation, 
for instance due to thin objects having small areas but large contours.
In this context, the most used metric by far is the Boundary Recall (BR) 
evaluating the detection of the groundtruth contours $\mathcal{B(\GG)}$ 
by superpixel boundaries $\mathcal{B}(\SSS)$: %\vspace{-0.15cm}

{\eqsize
\begin{equation}
%BR(\SSS,\GG) = \frac{\sum_{p\in\mathcal{B}(\GG)}\delta[\min_{q\in\mathcal{B}(\SSS)}\|p-q\|< \epsilon]}{|\mathcal{B}(\GG)|}  ,   \label{br}
\text{BR}(\SSS,\GG) = \frac{1}{|\mathcal{B}(\GG)|}\sum_{p\in\mathcal{B}(\GG)}\delta[\min_{q\in\mathcal{B}(\SSS)}\|p-q\|< \epsilon]  ,   \label{br}
\end{equation}
}
\noindent with 
$\delta[a]=1$ when $a$ is true and $0$ otherwise,
and $\epsilon$ is a distance threshold generally set to $2$ pixels \citep{liu2011,vandenbergh2012}. 
%
%Each ground truth pixel is considered as detected, if a superpixel boundary is at less than an $\epsilon$ distance. 
%

In most papers, only BR performance are reported although many works \emph{e.g.}, \citep{machairas2015,giraud2017_jei,stutz2018superpixels}, have shown that 
noisy superpixel methods are very likely to obtain higher BR (see Figure~\ref{fig:br}), 
although it disregards the superpixel definition that should have clearly identifiable regions.
Hence, BR 
%does not reflect the ability to efficiently capture the object contours and
should be measured along Precision (P), 
{\color{black} the percentage of true detections among superpixel boundaries,}
as for standard precision-recall measures \citep{vandenbergh2012}, 
or along
Contour Density (CD), the number of superpixel contour pixels
\citep{machairas2015}.
%can also be used to balance BR. 

\subsubsection{Regularity\label{subsubsec:gr}}

The over-segmentation into superpixels is supposed 
to have a certain regularity, with clearly identifiable regions,
and respecting a pseudo grid layout.
Nevertheless, this aspect is generally only evaluated qualitatively 
or with a non robust metric.

The most used metric is the compactness (CO) \citep{schick2014evaluation}, also called circularity is defined as:
{\eqsize
\begin{equation}
%\text{C}(S) = \frac{4\pi |S|}{|\mathcal{B}(S)|^2} , 
\text{CO}(S_i) = \frac{4\pi |S_i|}{|P(S_i)|^2} ,  \label{circu}
\end{equation} %\vspace{-0.1cm}
}
where $P(S)$ is the superpixel perimeter.
However, this metric has been proven 
non-robust to noise and scale, and favors circular and ellipsoidal shapes over square ones \citep{machairas2015,giraud2017_jei}, 
although square-like shapes should be considered the most regular overall.

\medskip

In \citep{giraud2017_jei}, the Global Regularity (GR) metric is proposed to 
address the non robustness of the compactness,
and evaluates both shape regularity and consistency. 
%Although this metric has been regularly used to evaluate the regularity (see Table I of the main paper),
%it has been proven very limited \cite{giraud2017_jei}.
%
%The GR metric \citep{giraud2017_jei} has been introduced to address the
%non robustness of the compactness.
%
GR is composed of two terms: a Shape Regularity Criteria (SRC) 
that independently evaluates each superpixel shape and 
a Smooth Matching Factor (SMF) that evaluates the consistency of the shapes.

First, instead of comparing the superpixels to a circle as for the CO metric, 
SRC robustly evaluates convexity, contour smoothness, and 2D balanced repartition of each superpixel shape.
For a supeprixel shape $S_i$, a discrete convex hull $H_{S_i}$ is computed and used such as: %\vspace{-0.2cm}
%
% as follows:

 {\eqsize
 \begin{equation}
%% % \text{SRC}(\SSS) = \frac{1}{\sum_i |S_i|}\sum_i |S_i|.\frac{\text{CC}(H_{S_i})}{\text{CC}(S_i)}\text{V}_{\text{xy}}(S_i) ,  \label{circu}
 \text{SRC}(S_i) = \frac{\text{CC}(H_{S_i})}{\text{CC}(S_i)}\text{V}_{\text{xy}}(S_i) ,  \label{src}
 \end{equation}}%
 \noindent  where 
 CC measures the ratio between the perimeter and the area of a shape,
 $\text{V}_\text{xy}(S_i) = {\min(\sigma_x,\sigma_y)/\max(\sigma_x,\sigma_y)}$,
 evaluates the balanced repartition of the shape
 with $\sigma_x$ and $\sigma_y$ 
 the square root of standard deviations of pixel positions $x$ and $y$ in $S_i$.
% % with the 
% % convex hull of the superpixel shape, and $P$ denoting the shape perimeter.
% %
% %Contrary to the compactness, SRC remains consistent with the superpixel size, moderately impacted by noise at the boundary and also gives the highest regularity score for square (not only circles).
% This way SRC robustly evaluates convexity, contour smoothness, and the 2D balanced repartition of pixels.

To also evaluate the superpixel segmentation has a whole, 
a Smooth Matching Factor (SMF) term
%inspired from \citep{machairas2015} 
is introduced in GR.
First, each superpixel $S_i$ is registered on 
its barycenter to obtain a registered shape $S_i^*$.
Then, these registered superpixel shapes are averaged to compute $S^*$, 
the average superpixel shape.
%created from the superposition of all superpixels.
 %
 Finally, each registered superpixel $S_i^*$ is compared to the average shape $S^*$ such that: \vspace{-0.2cm}
% %\vspace{-0.35cm}
% %

 {\eqsize
\begin{equation}
% % \text{SMF}(\SSS) = 1- \frac{1}{\sum_i |S_i|}\sum_i |S_i|.\left\|\frac{S^*}{|S^*|} - \frac{S_i^*}{|S_k^*|}\right\|_1/{2}. \label{smf}
 \text{SMF}(S_i) =  1 - \frac{1}{2}\left\|\frac{S^*}{|S^*|} - \frac{S_i^*}{|S_i^*|}\right\|_1 %\bigg/{2} 
 . \label{smf}
 \end{equation}}%
 %\noindent that compares each superpixel 
 %after registration on its barycenter,
 %to $S^*$ the average superpixel shape, created from
 %the superposition of all registered superpixel shapes. %on their barycenters and averaging their shapes.

Finally, GR combines these two terms to be defined as: \vspace{-0.2cm} %\vspace{-0.1cm}

{\eqsize
\begin{equation}
\text{GR}(\SSS) = \frac{1}{\sum\limits_{S_i\in \mathcal{S}} |S_i|}
\sum\limits_{S_i\in \mathcal{S}} |S_i| \, \text{SRC}(S_i) \, \text{SMF}(S_i) . \label{gr} 
\end{equation}
}%

The relevance of GR over CO is illustrated with several examples in Figure~\ref{fig:gr}.
Hence, we advocate using GR to evaluate the superpixel regularity.
%of superpixel segmentations.

\subsubsection{Color homogeneity}

\begin{figure*}[t!]
\centering
{\footnotesize
 \begin{tabular}{@{\hspace{0mm}}c@{\hspace{1mm}}c@{\hspace{2mm}}c@{\hspace{1mm}}c@{\hspace{1mm}}@{\hspace{0mm}}}
 &&ASA=1.00 \text{ } BR=1.00 &ASA=1.00 \text{ } BR=1.00\\
 &&{\color{darkgreen}EV=0.841}  \text{ }  {\color{darkgreen}SIRS=0.906} & {\color{darkred}EV=0.550}  \text{ }  {\color{darkred}SIRS=0.757}\\
% &&{\color{darkgreen}SIRS=0.906} & {\color{darkred}SIRS=0.757} \\
\includegraphics[width=0.24\textwidth]{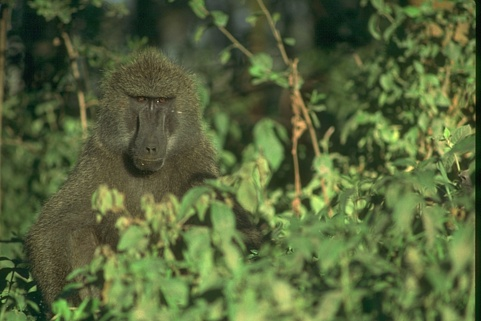}&
\includegraphics[width=0.24\textwidth]{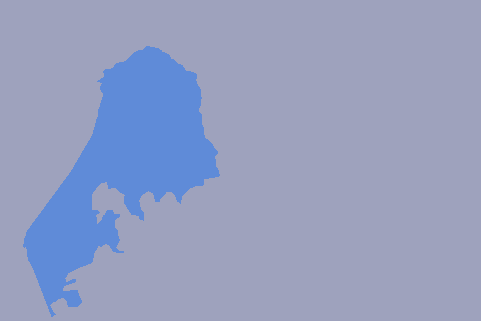}&
\includegraphics[width=0.24\textwidth]{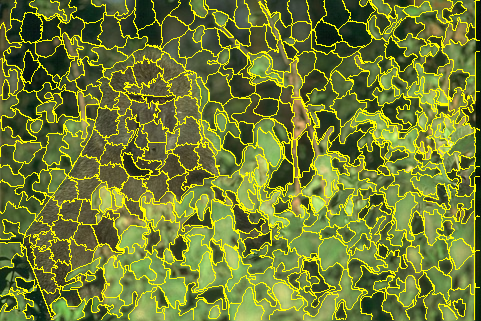}&
\includegraphics[width=0.24\textwidth]{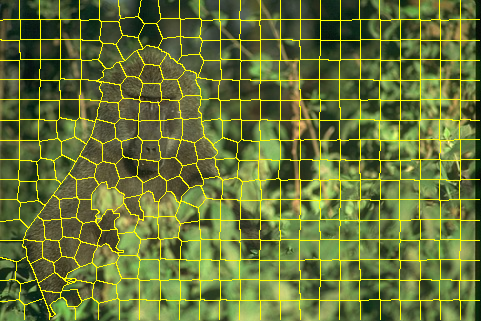}\\
(a) Image & (b) Groundtruth labels & (c) Low-level feature & (d) Non low-level feature\\
&& sensitive seg. & sensitive seg. \\[-1ex]
\end{tabular}}
\caption{\textbf{Illustration of low-level sensitivity on the clustering.} 
On this example, segmentation (ASA) and contour detection (BR)
performance are extremely correlated to high or object-level feature information since the image (a) only contains one complex object in its groundtruth (b).
Both segmentations (c) and (d) have the same maximum ASA and BR scores, although (d) seems to only focus on high-level information and ignores low-level features such as color or texture. 
\revision{The EV and SIRS metrics express the ability of superpixels to cluster homogeneous colors regardless of the presence of objects.}
} \vspace{-0.25cm}
\label{fig:ev}
\end{figure*}

As for regularity, color homogeneity can be evaluated regardless of any groundtruth information.
Such aspect is initially at the core of the superpixel definition that tries to provide a local grouping of pixels with similar low-level features such as color.
In \citep{benesova2014}, the intra-cluster variation (ICV) metric has been proposed 
%to evaluate such property, 
but later studies report its non-robustness to scale and to varying image dynamics.
These studies %seem to have successfully converged 
has led to the use of the Explained Variation (EV) \citep{moore2008} that has a similar formulation considering superpixel color variance $\mu(S_k)$, but is independent to scale and is normalized by the image color variance, allowing a fair comparison of EV scores between images: \vspace{-0.2cm}

{\eqsize
\begin{align}
\text{EV}(\SSS)   
%&= \frac{\sum_{S_j}{|S_j|\left(\mu(S_j) - \mu(I)\right)^2}} 
%             {|I|\sigma^2(I)}  \nonumber \\
&= \frac{\sum_{S_k}{|S_k|\left(\mu(S_k) - \mu(I)\right)^2}} 
          {\sum_{p\in I}{\left(I(p)-\mu(I)\right)^2}} . \label{ev}
\end{align}
}

EV gets close to 1 as the number of superpixels increases to the number of pixels, or contain pixels with the same color, 
 and equals 0 when there is only
one superpixel.
\revision{However, EV remains limited since it considers the mean color of a superpixel \citep{moore2008}, thus causing uniform areas to have a disproportionate impact on the metric.}

\revision{To overcome the mean color drawback, the Similarity between Image and Reconstruction from
Superpixels (SIRS) \citep{barcelos2022improving} 
has recently been proposed to model the color homogeneity problem as an image reconstruction problem.
The metric uses an RGB Bucket Descriptor (RBD) representing the colors of each superpixel $S_i$ as a small set of its
most relevant colors, such that 
$\text{RBD}(S_i) = \{c_1,\dots,c_\alpha\}$, with $\alpha$ set to $4$ by default \citep{barcelos2022improving}.
SIRS proposes to evaluate the best image reconstruction possible from
the most relevant colors by first building for each pixel $p\in S_i$,
%$R(p) =  \underset{c_i\in V}{\text{argmin}}\{\|I(p)-c_i\|_1\}$.
$R(p) = \text{argmin}_{c_j\in \text{RBD}(S_i)}\{\|I(p)-c_j\|_1\}$.
Then, it computes the Mean
Exponential Error (MEE) between $R$ and the initial image $I$ as:
\begin{equation}
    \text{MEE}(\mathcal{S}) = \frac{1}{|I|}\sum_{S_i\in\mathcal{S}} \sum_{p\in S_i} \|R(p) - I(p)\|_1^{2-\psi},
\end{equation}
in which $\psi = \text{max} \|c_l - c_j\|_1$ and $c_l, c_j \in \text{RBD}(S_i)$. 
With this formulation, 
superpixels that capture uniform areas have a limited impact while 
the ones that require a palette of highly discrepant colors would lead to higher errors, since they correspond to complex patterns. %Conversely, if the relevant colors are similar and, thus,are representing a more uniform texture, such impact must be light. 
Finally, SIRS is defined as:
\begin{equation}
    \text{SIRS} = \text{exp}^{-\frac{\text{MEE}(\mathcal{S})}{\sigma^2}},
\end{equation}
where $\sigma$ controls the importance given to small error variations and is set to 0.01  \citep{barcelos2022improving}.
SIRS also ranges from 0 to 1, with 1 being the best value.}
%higher the value, better is
%the color homogeneity of the superpixels in S, represented within [0, 1].

%Let $G^{S_i} \in \mathbf{S}(S_i, 7)$ represent the set of 7 disjoint groups related to each RGB
%cube vertices, whose colors are $c_l \in [0, 1]^3$, forming a
%color set $V=\{c_l\}$
%in which $1 \leq l \leq 7$. 
%Then, each $G_l^{S_i} \in G^{S_i}$ is populated
%by assigning every $p \in S_i$ to its most similar group using a mapping function $M(p)$:
%\begin{equation}
%M(p) =  \underset{c_i\in V}{\text{argmin}}\{\|I(p)-c_i\|_1\}
%\end{equation}

The color homogeneity property is the rarest one evaluated, %in the litterature, 
although we show in Figure~\ref{fig:ev} that it may express a significant behavior of a superpixel segmentation method.
Since groundtruth information is generally object-wise
and recent deep learning-based methods extract high-level features, 
their task of superpixel segmentation gets closer 
to a standard object segmentation problem.
Therefore, it remains necessary to evaluate color homogeneity that
expresses the ability of methods to gather homogeneous pixels even in object-free areas. 
%Otherwise, 
%
% Note that the EV metric should always be considered in comparison with a regularity metric 

\subsection{Processing time}

\revision{The processing time is an interesting criterion and should %systematically
be reported when presenting a superpixel method.
Most works provide runtime comparisons by executing methods on the same machine for several numbers of superpixels, although such evaluations are only meaningful if conducted under strictly comparable conditions.
% In the past, some approaches emphasized theoretical computational complexity, which, although informative, often remains difficult to interpret in practice and does not necessarily reflect real-world performance. 
Theoretical computational complexity provides useful insight into how a method scales with image size or the number of superpixels.
% We argue that reporting actual wall-clock runtime is more transparent and practically relevant, 
Yet, reporting actual execution time is equally important, as it better reflects real-world performance,
provided that it is rigorously measured on the same hardware, under identical experimental settings, and includes all components of the pipeline, such as pre-processing (\emph{e.g.}, filtering) and post-processing (\emph{e.g.}, connectivity enforcement).}

\revision{At the same time, runtime comparisons must be put into perspective with respect to the implementation details.
Methods implemented in different programming languages or with different levels of optimization may exhibit significant discrepancies that do not reflect their intrinsic algorithmic efficiency \citep{wu2024survey}. 
Furthermore, some algorithms are iterative and allow a trade-off between speed and accuracy: a method that appears slower at convergence may reach competitive performance with fewer iterations and reduced runtime.
Therefore, implementation choices and stopping criteria should be clearly stated. 
Interestingly, with the growing adoption of deep learning-based approaches, this particular bias tends to diminish, as most methods rely on similar software frameworks and hardware acceleration (\emph{e.g.}, GPU-based inference), making runtime comparisons more homogeneous, although still dependent on model size and inference strategy.}
% The processing time is often reported, with methods run on the same computer for several number of superpixels.
% %
% However, methods may be implemented in different languages \citep{wu2024survey} that
% do not represent their inherent complexity.
% %
% Another aspect is the iterative behavior of certain methods. 
% %
% Can a slower but more precise method reach the same processing time with fewer iterations and still be more accurate?
% %
% Hence, processing time should always be put into perspective and 
% the nature of the implementation mentioned.
%
%Finally, note that the comparison tends to be more relevant with recent
%deep learning-based methods implemented in Pytorch.

\subsection{Robustness to noise\label{subsec:noise}}

\begin{figure*}[t!]
\centering
{\footnotesize
 \begin{tabular}{@{\hspace{0mm}}c@{\hspace{1mm}}c@{\hspace{1mm}}c@{\hspace{1mm}}c@{\hspace{1mm}}@{\hspace{0mm}}}
\includegraphics[width=0.24\textwidth]{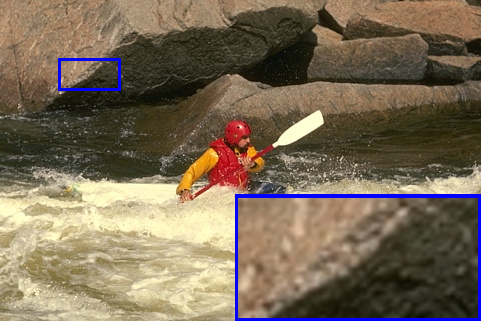}&
\includegraphics[width=0.24\textwidth]{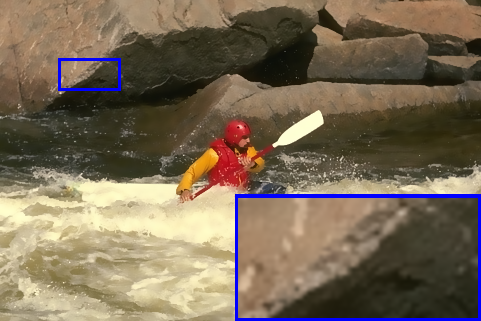}&
\includegraphics[width=0.24\textwidth]{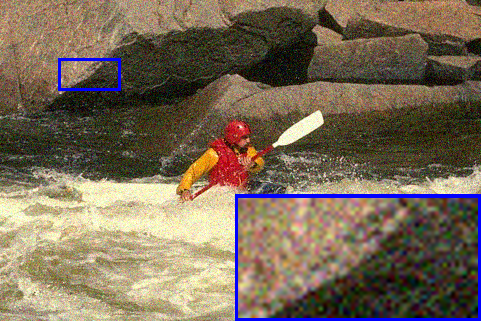}&
\includegraphics[width=0.24\textwidth]{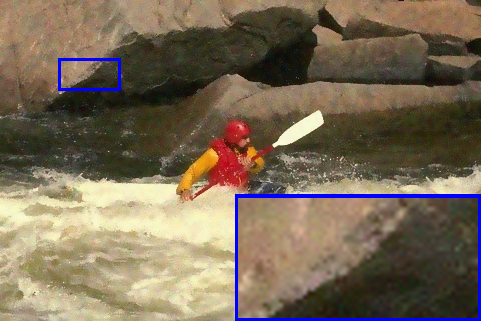}\\
(a) Image & (b) Filtered image & (c) Noisy image & (d) Filtered noisy image\\
\end{tabular}}
\caption{\textbf{Impact of the filtering pre-processing.} 
A simple 7x7 bilateral filtering may slightly smooths image textures while preserving contours (a)-(b)
or be sufficient to limit the  impact of noise (c)-(d) that may greatly affect the performance of superpixel algorithms (here additive Gaussian of variance 20).
%may greatly improve the performance of superpixel methods.
  } 
  \label{fig:bilateral}
  \end{figure*}
  
\begin{figure*}[ht!]
\centering
{\footnotesize
 \begin{tabular}{@{\hspace{0mm}}c@{\hspace{2mm}}c@{\hspace{2mm}}c@{\hspace{2mm}}c@{\hspace{0mm}}}
 \multicolumn{2}{c}{\scriptsize Traditional methods} &  \multicolumn{2}{c}{\scriptsize DL-based methods}\\[1ex]
\includegraphics[width=0.24\textwidth,height=0.16\textwidth]{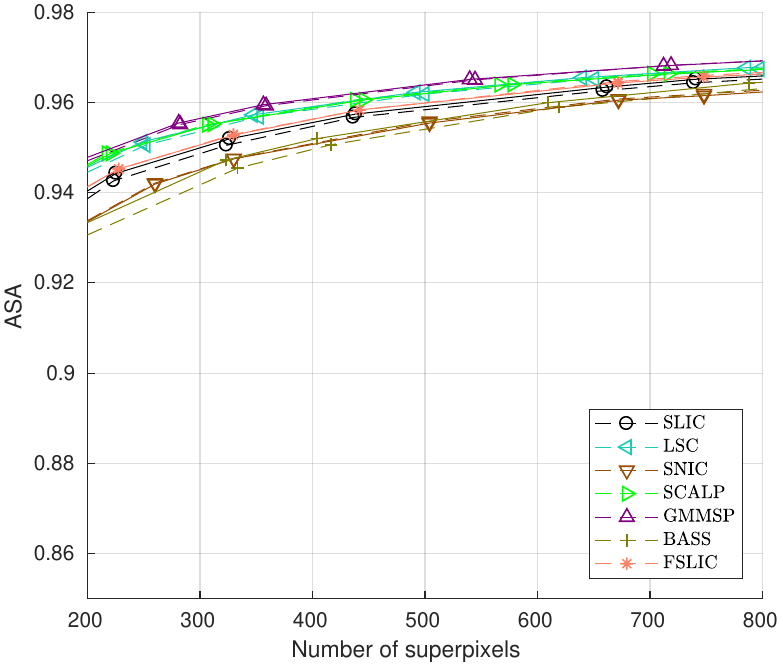}&
\includegraphics[width=0.24\textwidth,height=0.16\textwidth]{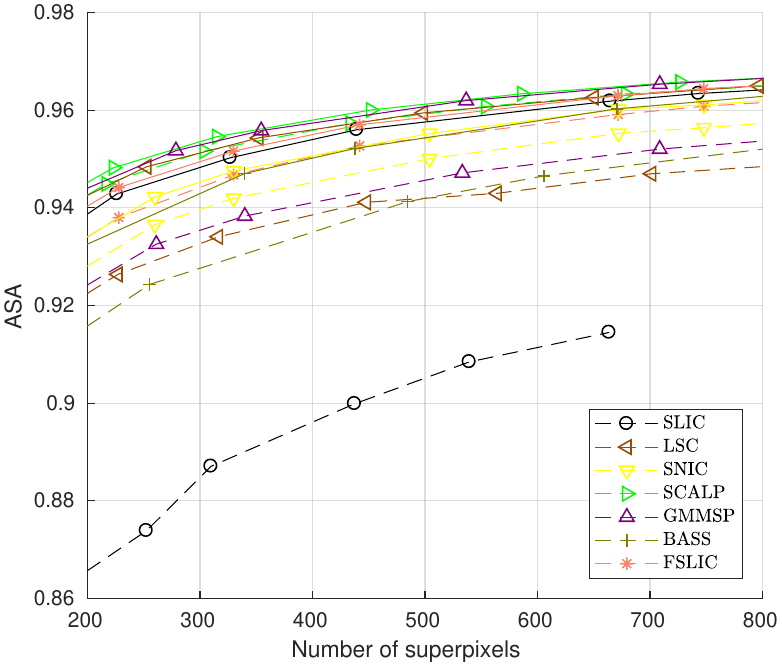}&
\includegraphics[width=0.24\textwidth,height=0.16\textwidth]{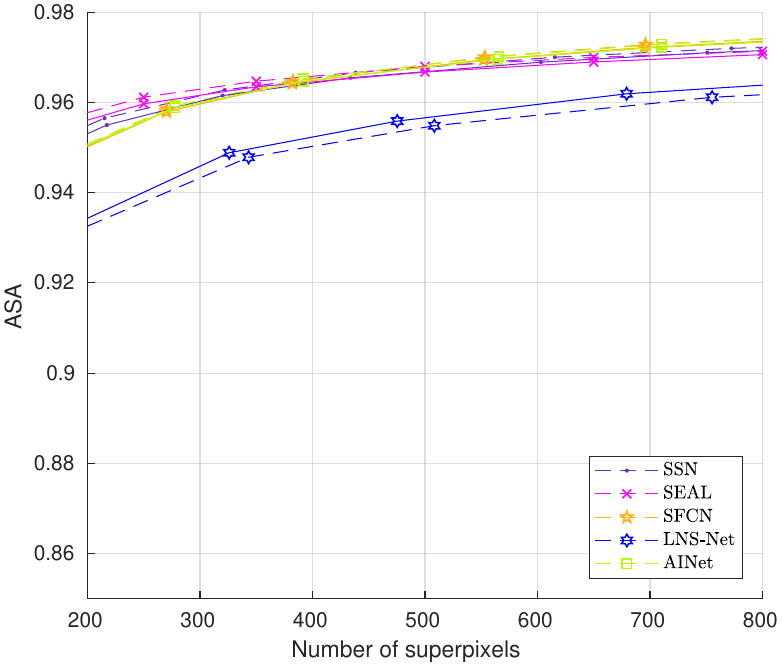}&
\includegraphics[width=0.24\textwidth,height=0.16\textwidth]{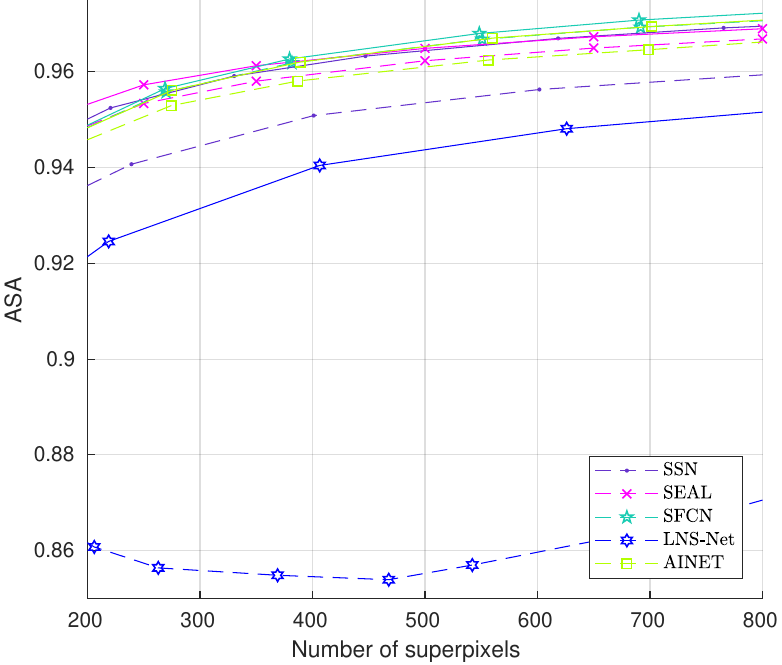}\\
(a)  Noise-free BSD images  & (b) Noisy BSD images  & (c)  Noise-free BSD images  & (d) Noisy BSD images  \\
\end{tabular}}
\caption{\textbf{Evaluation of the robustness to noise.} 
On BSD images,
the pre-filtering has a slightly positive impact on segmentation accuracy (ASA) for most traditional methods (a) and a slightly overall negative impact on deep learning (DL)-based methods (c). 
%but significantly improves the regularity (see supp. mat.).
However, on noisy images (b)-(d), the pre-filtering largely improves the accuracy and prevent dramatic losses of performance, \emph{e.g.}, for SLIC or LNS-Net. 
Results w/o and w/ pre-filtering are respectively reported with dotted (-\hspace{0.15mm}-) and plain lines (--).
  } 
  \label{fig:noise_perf}
  \end{figure*}

% \begin{table}[t]
%     \centering
%     {\footnotesize
%     \begin{tabular}{c|c|c|c|c|c|c|c|c|c|c|c|c|c}
%     & SLIC & LSC & SNIC & SCALP & GMMSP & BASS & FSLIC & SSN  & SEAL & SFCN & LNS-Net & AINet \\ \hline
%     Initial images \\
%     Initial images + BF \\ \hline
%     Noisy images \\
%     Noisy images + BF \\ \hline
%     \end{tabular}
%     }
%     \caption{Caption}
%     \label{tab:my_label}
% \end{table}

In several works \citep{giraud2018_scalp,wu2020fuzzy}
and studies \citep{brekhna2017robustness,stutz2018superpixels,barcelos2024comprehensive,wu2024survey}, the robustness to noise is evaluated as a property.
\revision{Several noises are encountered, including additive Gaussian ones (GN in Table \ref{tab:metrics}) with a variance below 20, and impulse noise (IN) such as salt-and-pepper.
Some works also evaluate the robustness to Blur (B).}
Methods such as SLIC, based on a simple trade-off between color and spatial features 
%and using a post-processing step to enforce connectivity, 
may be highly impacted by the presence of noise.
%
%{\color{red} Conclusion études}

\revision{Among the types of noise that can be used, we recommend prioritizing impulse noise since it may largely affect performance, while Gaussian noise may be almost canceled by a pre-filtering step}.
For instance in \citep{giraud2018_scalp}, the color image used as input of the algorithm is first pre-processed with a simple bilateral filter, computed almost instantaneously.
Using this pre-processing step allows the method to withstand the impact of noise and also improves the segmentation accuracy on noise-free images.
A 7x7 bilateral filtering is illustrated in Figure~\ref{fig:bilateral}.

Such filtering can be seen as a pre-processing step that is independent of the method, and consequently applicable to all state-of-the-art methods regardless of the presence of noise.
%
% a prior Gaussian smoothing can also be encountered in some public implementations of superpixel methods.
%
In Figure~\ref{fig:noise_perf}, we compare the performance of some traditional and deep learning (DL)-based methods w/ and w/o such bilateral pre-filtering on the noise-free BSD images and the images afflicted by a strong Gaussian noise of variance 20.
On noise-free images, the pre-filtering only very slightly improves or degrades segmentation performance. %while significantly improving regularity (see supp. mat.).
On noisy images, it largely improves the performance for all methods and prevent dramatic losses of accuracy.
For instance, GMMSP performs better than SNIC after pre-filtering on noisy images.
%
%Other quantitative and qualitative results are reported in the supp. mat.

% So we see that  even changing the performance hierarchy between methods.
% %
%It may remain interesting to study the inherent robustness of superpixel algorithms to noise. 

\revision{Hence, the robustness to Gaussian noise should be put into perspective
since a simple pre-filtering such as \citep{giraud2018_scalp}, 
%enables to
largely improves accuracy for all methods on noisy images, without degrading average performance on noise-free images.}
%improve the performance 
%for all methods, but significantly 
%in both 
%noise-free and significantly noisy
%cases.
%
\revision{Therefore, we recommend prioritizing the study of robustness to impulse noise
at several levels, as it is done in Section \ref{subsubsec:perf}.}
%Therefore, it may be limited to compare the robustness of methods without applying it. 
%such as \citep{giraud2018_scalp}
%
Finally, the evaluation of the robustness to stronger noises is questionable since 
having to deal with such images would require dedicated pre-denoising.

\begin{figure}[t!]
\centering
{\footnotesize
 \begin{tabular}{@{\hspace{0mm}}c@{\hspace{1mm}}c@{\hspace{0mm}}}
\includegraphics[width=0.24\textwidth]{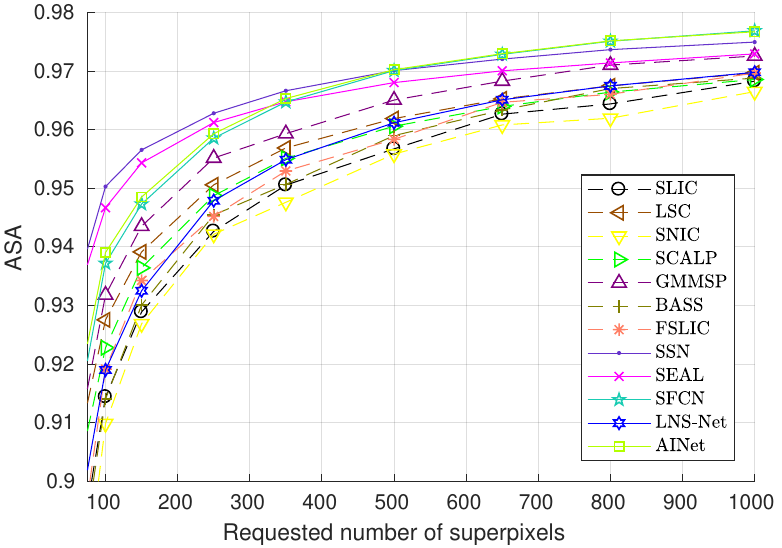}&
\includegraphics[width=0.24\textwidth]{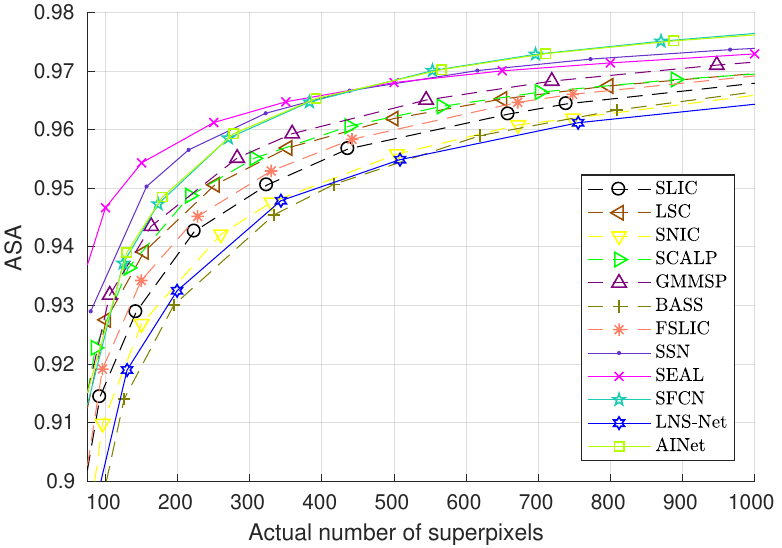}\\
%(a) ASA vs generated number of SP & (b) ASA vs requested number of SP\\[-1ex]
\end{tabular}}
\caption{\textbf{Performance over the number of superpixels.} 
ASA is reported for traditional (-\hspace{0.15mm}-) and DL-based (--) methods according to the requested number of superpixels (left), and the actual average number of generated superpixels (right). 
See how the hierarchy is changed between the two representations.}  \vspace{-0.1cm}
  \label{fig:number}
  \end{figure}

\subsection{Number of superpixels}

Results on previous metrics are generally computed for several scales, 
\emph{i.e.}, several numbers of superpixels.
\revision{Nevertheless, almost half of the works in Table~\ref{tab:metrics} 
(represented with orange dots)
may incorrectly report the performance by considering 
the required number of superpixels instead of the actual generated one, as pointed out in \citep{stutz2018superpixels}.}
Indeed, many superpixel methods do not exactly generate the required number of superpixels, due to a square initialization grid, or a post-processing step that removes too small superpixels.
This difference between required and generated number 
is generally around 10\% for concerned methods,
which can be enough to change the hierarchy of performance.
Moreover, some methods such as LNS-Net \citep{zhu2021learning} can generate up two twice
more than requested superpixels,
%compared to the requested number, 
leading to
unfair comparison if reporting results based on the required number.

In Figure~\ref{fig:number}, we report object segmentation results (ASA) for some state-of-the-art methods using the two representations.
We see that the hierarchy between methods differ, for instance between SEAL and SSN or SCALP and LSC.
Since the scale has a significant impact on performance, 
the average generated number should be measured for each method 
and used in the representation.
Finally,  such aspect can also be measured with the Variance of Superpixel Number (VSN) \citep{wu2020fuzzy} measuring the average square error between required and generated number.

\subsection{Impact of the regularity setting}

As illustrated in Section \ref{sec:ill_posed} with Figure~\ref{fig:diff_regu},
the same method can produce superpixels with a variable regularity level.
%that has a direct impact on the ability to fit to the image objects.
%
In several papers, the chosen parameters are not mentioned, 
although setting high regularity constraints %for compared methods 
is very likely to degrade the object segmentation and contour detection performance.
So naturally, it is mandatory to set the parameters of methods, according to the authors recommendations, that generally provide a trade-off between regularity and segmentation accuracy.
%
%Therefore, comparing methods with their recommended parameters is mandatory.

\smallskip

Moreover, only comparing methods on object and contour aspects 
 may be biased since it does not measure their respect of the implicit grid nor their potential with different settings. 
In this context, the GR metric offers a dedicated measure of a superpixel segmentation regularity, yet few methods have compared 
in terms of regularity (see Table~\ref{tab:metrics}).

\smallskip

To go further, %as in \citep{giraud2017_jei}, 
the evaluation can be done for several regularity settings.
Very few works have reported this experiment (/Regu. column in Table~\ref{tab:metrics}),
%to compare their potential and extreme behaviors.
%
and it is generally only done for the presented method.
A simple way to do so %compare state-of-the-art methods
is to carefully set a superpixel scale
and to sample the range of regularity settings for each method.
This evaluation is reported in Section \ref{subsubsec:perf}
and expresses the potential and the extreme behaviors of methods.
{\color{black}
For instance, 
most deep learning-based
methods do not allow to set a variable regularity level without
a new training process and a dedicated model.
}
%many deep learning-based methods cannot get close to a square grid.
%
Finally, it answers the question: can methods achieve higher performance with lower or higher regularity?

\section{Superpixel Segmentation from Object Segmentation\label{sec:samo}}
%\section{Superpixels from Sparse Object Segmentation}

\begin{figure*}[t!]
\centering
{\footnotesize
 \begin{tabular}{@{\hspace{0mm}}c@{\hspace{2mm}}c@{\hspace{1mm}}c@{\hspace{1mm}}c@{\hspace{1mm}}@{\hspace{0mm}}}
\includegraphics[width=0.23\textwidth]{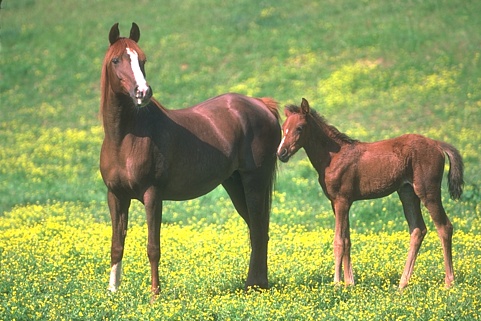}&
 \includegraphics[width=0.23\textwidth]{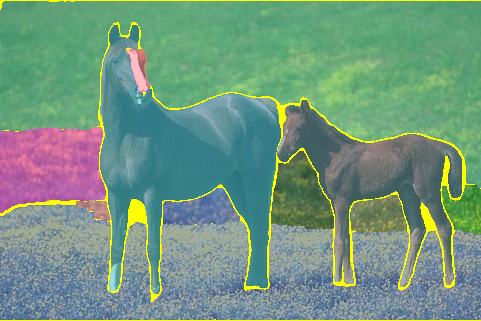}&
\includegraphics[width=0.23\textwidth]{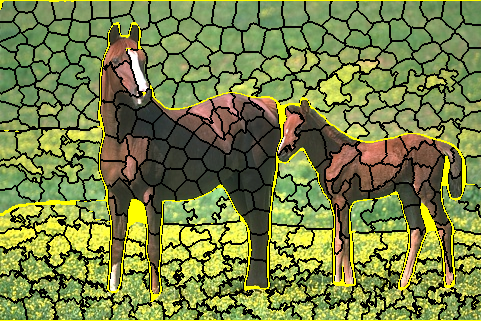}&
\includegraphics[width=0.23\textwidth]{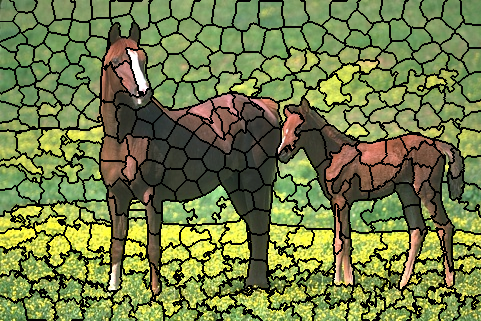}\\
\includegraphics[width=0.23\textwidth]{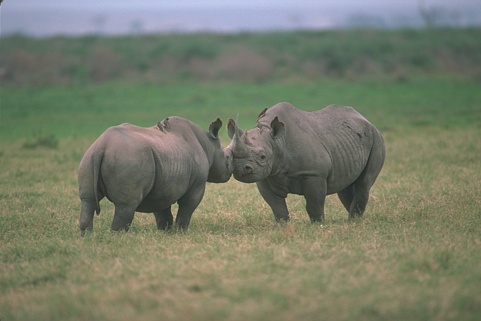}&
\includegraphics[width=0.23\textwidth]{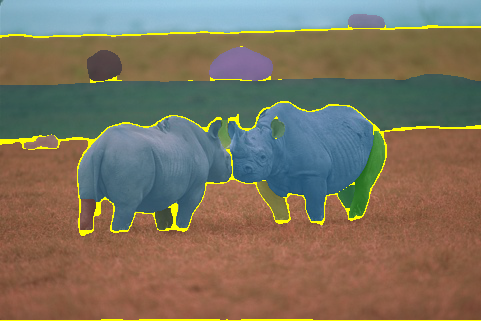}&
 \includegraphics[width=0.23\textwidth]{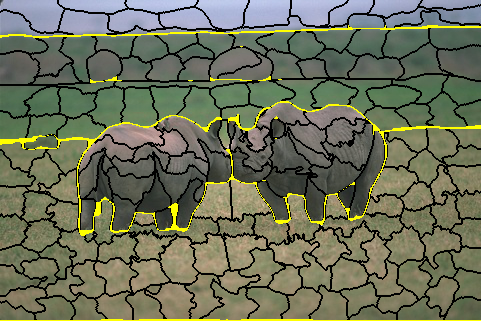}&
 \includegraphics[width=0.23\textwidth]{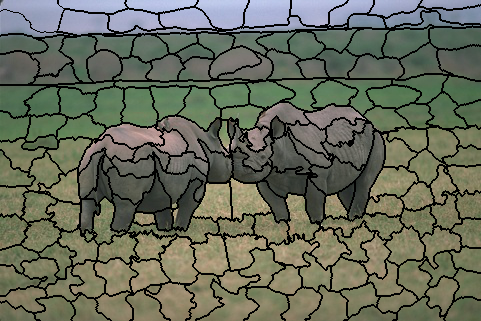}\\
(a) Image & (b) SAM objects & (c) Contained superpixels using & (d) Final superpixels\\
& \citep{kirillov23sam}  & \citep{irving2016mask}  \\%[-1ex]
\end{tabular}}
\caption{\textbf{Proposed superpixel segmentation from object segmentation.} 
(b) SAM \citep{kirillov23sam} is used to generate object proposals (with unlabeled pixels in yellow),
(c) filled with superpixels using mask-SLIC \citep{irving2016mask}. (d) Remaining unlabeled pixels are finally associated by nearest neighbor to produce a final segmentation that accurately respects image objects, even containing thin structures, but also have regular superpixels within.}% 
  \label{fig:samo}
  \end{figure*}

\subsection{Object segmentation proposal\label{subsec:samo_details}}

In this section, we show that with the advent of 
extremely robust semantic-agnostic deep learning models,
superpixel segmentation can
be efficiently tackled as a standard object segmentation problem. 
We propose a simple method to obtain a superpixel segmentation constrained by a prior object segmentation.
This method is composed of three main steps illustrated in Figure \ref{fig:samo}.
First, a large pre-trained segmentation model is used to obtain a prior object segmentation.
Then, as this prior can contain overlapping and/or superfluous objects, we apply a processing step to transform it into a proper segmentation (\emph{i.e.}, the set of regions form an image partition).
Finally, a superpixel segmentation algorithm is applied 
individually on each region %of this image segmentation 
to obtain object-level superpixels.

\revision{The idea behind this method is to benefit from the capacity of large pre-trained segmentation models to capture large objects and from the ability of superpixels to achieve finer segmentation. 
Both steps are complementary as large models may miss or struggle to segment small objects, while superpixels generally fail to capture the geometry of large objects.
The different steps of this method 
are detailed hereafter.}

%\smallskip
%\noindent\textbf{Pre-trained segmentation model}
%
\subsubsection{Pre-trained segmentation model}

First, we use the SAM model~\citep{kirillov23sam} with default Vit-H pre-trained weights to obtain a prior object segmentation.
As input prompt for the model, we use a uniform grid of foreground points ($32 \times 32$ in the following).
%or a single foreground point per region of the ground truth segmentation (sampled at the center of ground truth masks, as in~\citep{kirillov23sam}), depending on the application.
In both cases, even after non-maximum suppression, the set of objects segmented by SAM does not result in a proper partition of the image, as (i) the objects often overlap, and (ii) the union of objects does not include all image pixels, leaving unlabeled either large regions (which can be considered as background) or thin boundaries between objects.

%\medskip
%\noindent\textbf{Segmentation processing}
%
\subsubsection{Segmentation processing}
We propose a processing step to transform the prior object segmentation into a proper image segmentation.
First, \revision{we only keep objects larger than a minimum area threshold 
set to $\sqrt{|I|}/4$}.
Then, we remove the overlap between these objects, by greedily removing them from each other in increasing area size (\emph{i.e.}, we remove the first object from all larger objects, and so on).
We also split unlabeled pixels into connected components and only keep those larger than the minimum area threshold.
We also apply morphological opening to differentiate between large background regions and thin boundaries between objects.
All remaining large regions (objects from SAM without overlap and large background connected components) are kept as objects for computing object-level superpixels, while remaining unlabeled pixels will be handled separately.
%At the end of this process, 
The union of these regions and unlabeled pixels form a proper image partition (Figure \ref{fig:samo}(b)).

%\smallskip
%\noindent\textbf{Object-level superpixels}

\begin{figure*}[t]
\centering
{\footnotesize
 \begin{tabular}{@{\hspace{0mm}}c@{\hspace{1mm}}c@{\hspace{1mm}}c@{\hspace{1mm}}c@{\hspace{1mm}}c@{\hspace{0mm}}}
\includegraphics[width=0.24\textwidth]{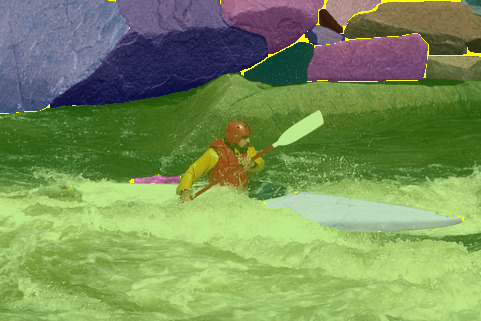}&
\includegraphics[width=0.24\textwidth]{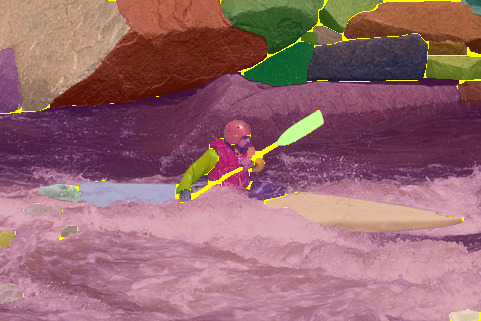}&
\includegraphics[width=0.24\textwidth]{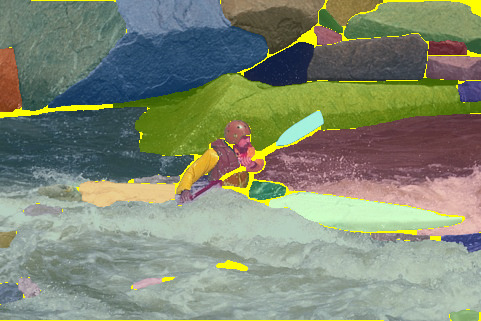}&
\includegraphics[width=0.24\textwidth]{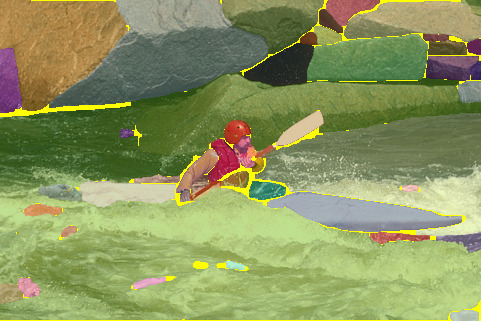}\\
 8x8 (14 objects) &  16x16 (31 objects) &  32x32 (36 objects) & 64x64 (49 objects)\\[2ex]
 \includegraphics[width=0.24\textwidth]{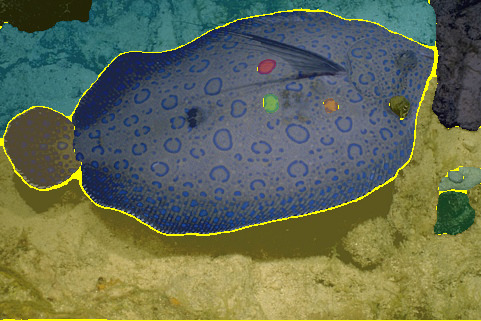}&
\includegraphics[width=0.24\textwidth]{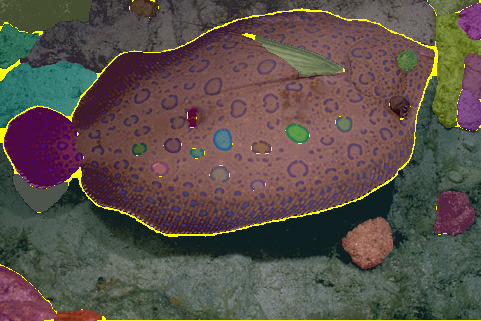}&
\includegraphics[width=0.24\textwidth]{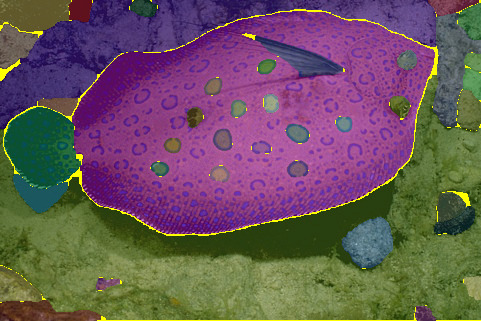}&
\includegraphics[width=0.24\textwidth]{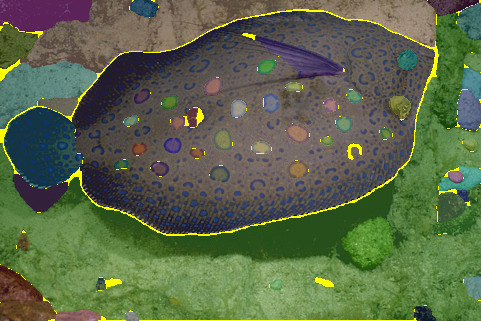}\\
 %Image & 
 8x8 (13 objects) &  16x16 (32 objects) &  32x32 (41 objects) & 64x64 (60 objects)\\
\end{tabular}}
\caption{\textbf{Comparison of SAM object detection} for different input grid levels (with unlabeled pixels in yellow). Highlighted objects are the result of the segmentation processing described in Section \ref{subsec:samo_details} of the main article (\textit{i.e.~} overlaps removal, area filtering and connected components analysis).
} 
\label{fig:ablation_sam}
\end{figure*}

\begin{figure*}[t]
\centering
{\footnotesize
 \begin{tabular}{@{\hspace{0mm}}c@{\hspace{2mm}}c@{\hspace{2mm}}c@{\hspace{2mm}}c@{\hspace{0mm}}}
\includegraphics[width=0.24\textwidth]{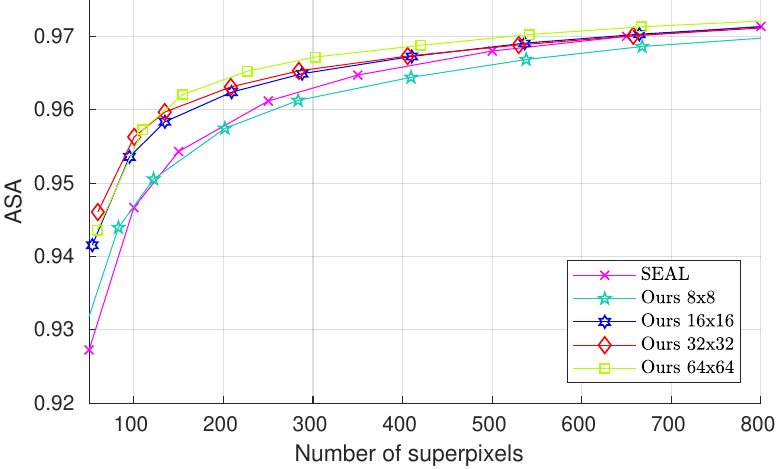}&
\includegraphics[width=0.24\textwidth]{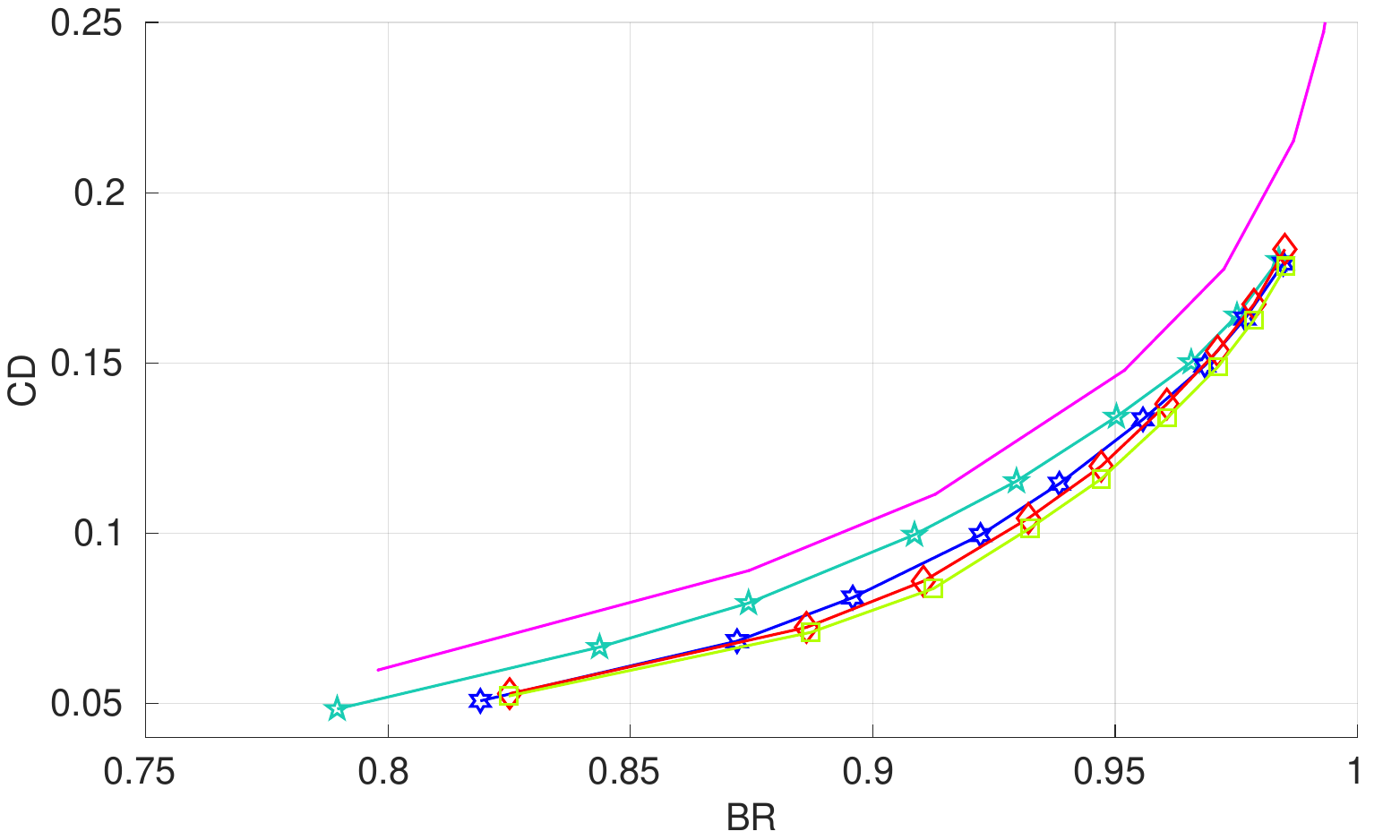}&
\includegraphics[width=0.24\textwidth]{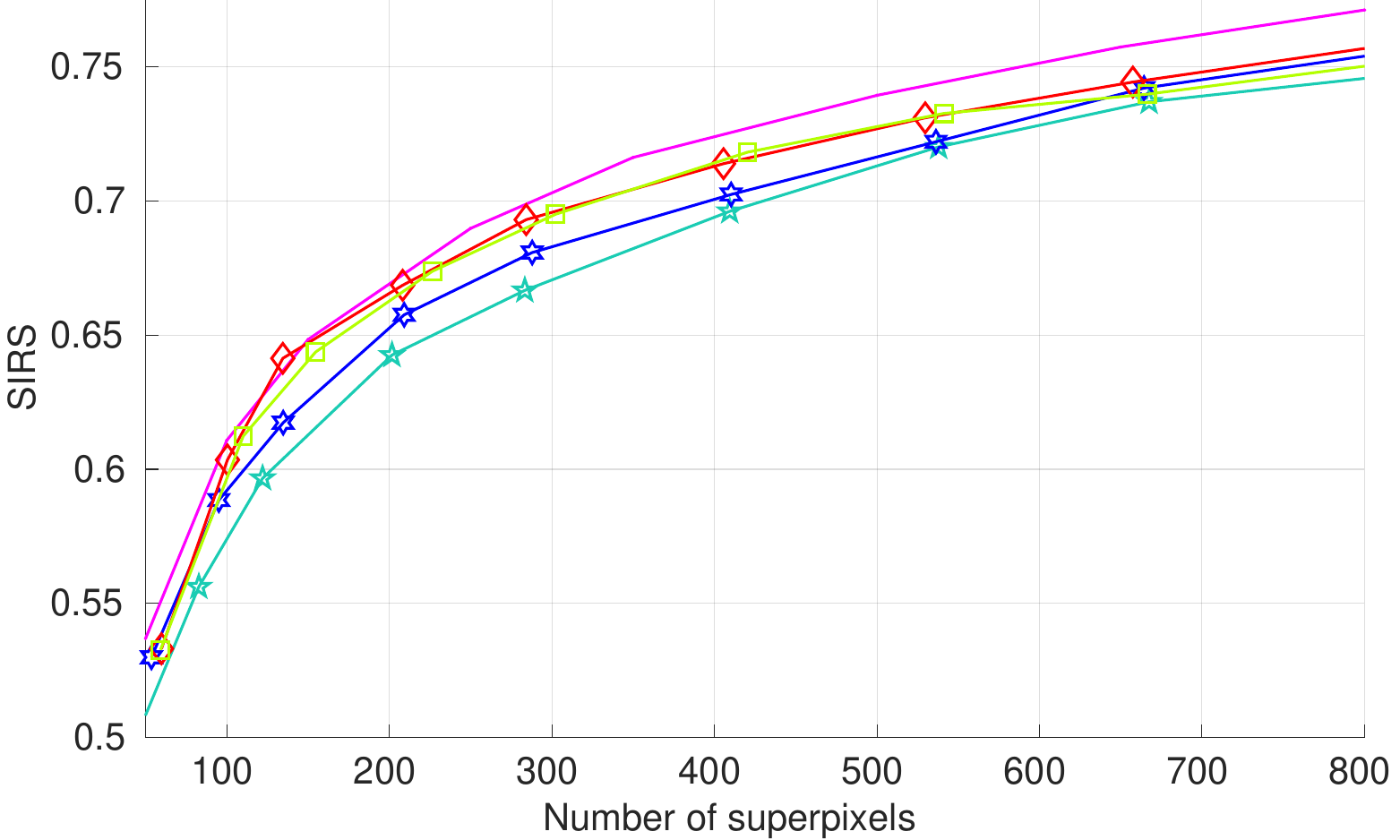}&
\includegraphics[width=0.24\textwidth]{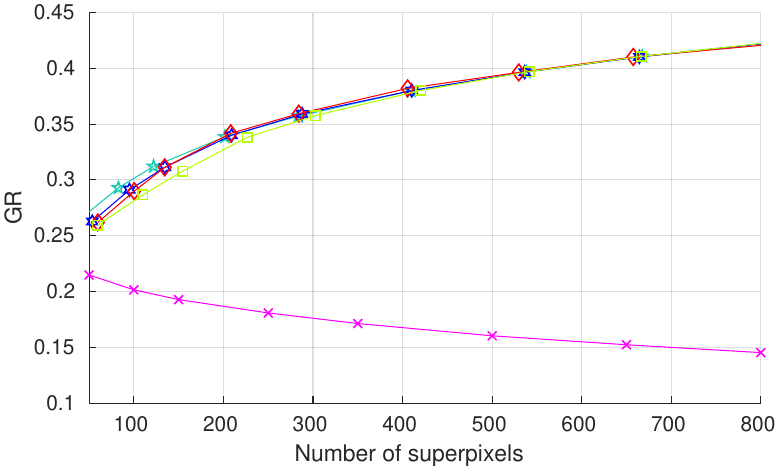}\\
\end{tabular}}
\caption{\textbf{Ablation study of SAM segmentation}. We compare the influence of the number of input points for the initial SAM segmentation on BSD images. 
  } 
  \label{fig:ablation_sam_results}
  \end{figure*}

\subsubsection{Object-level superpixels}
Given this partition, we apply the maskSLIC algorithm \citep{irving2016mask}
 to obtain superpixels for each region of the initial segmentation processing (Figure \ref{fig:samo}(c)).
\revision{The maskSLIC method applies the SLIC algorithm on a specific region of the image
starting from random seeds.
We use pre-filtered RGB color features as in Section \ref{subsec:noise},
a trade-off parameter of 5 between color and spatial features and 10 $k$-means iterations.
For an image of size $|I|$ segmented into $K$ superpixels, on a region object of size $r_i$,
we apply maskSLIC with a number of superpixels proportional to its area, \emph{i.e.}, $\frac{|r_i|}{|I|}K$, in order to keep an approximate total of superpixels corresponding to the requested number.
Unlabeled pixels are then assigned to their most appropriate superpixel by applying nearest neighbor search. For each surrounding superpixel, the barycenter and average color are considered to measure the distance to an unlabeled pixel.}
A final enforce connectivity step is applied to ensure that no additional regions are created.
In the end, we obtain a proper superpixel segmentation of the image where superpixel boundaries match the ones for the prior object segmentation obtained with SAM (Figure \ref{fig:samo}(d)).

%Nombre d'objets moyen par dataset :\\
%BSD :       GT: min=2, max=173, avg.=22 \quad SAM: min=5, max=138, avg.=48 \\
%NYUV2 :     GT: min=5, max=79,  avg.=25 \quad SAM: min=, max=, avg.= \\
%SBD :       GT: min=2, max=135, avg.=20 \quad SAM: min=, max=, avg.= \\
%SUNRGBD :   GT: min=4, max=82,  avg.=23 \quad SAM: min=, max=, avg.= \\

\subsection{Ablation study}
We propose an ablation study of our proposed approach using SAM~\citep{kirillov23sam} and maskSLIC~\citep{irving2016mask}.
In particular, we report in Figure~\ref{fig:ablation_sam} examples of the influence of the number of input points for the SAM segmentation, which has an impact on the number of objects and their resolution.
Note that in the comparison to state-of-the-art methods, we use a 32x32 grid of input points.

In Figure~\ref{fig:ablation_sam_results}, the influence of the number of input grid points for the initial SAM segmentation is reported for the recommended metrics. % for different numbers of input points. 
On the test images of the BSD, 8x8, 16x16, 32x32 and 64x64 grid points respectively 
generate on average 26, 46, 49 and 71 objects per image.
The performance on all metrics increase, except for regularity that logically slightly suffers from more object-level segmentation.
These results highlight that superpixel segmentation can be seen as an object segmentation problem, and efficiently tackled by accurate models such as SAM.

\subsection{Comparison to state-of-the-art methods}

\subsubsection{Dataset}

Results are reported on the mainly used datasets in superpixel segmentation, the BSD \citep{martin2001},
containing 200 test images of size 321x481 with fine segmentations up to 173 groundtruth objects or regions per image, 
NYUV2 \citep{silberman2012indoor} with 399 images of size 448x608 and SBD \citep{gould2009decomposing} containing 477 images mainly of size 240x320.
Note that all compared DL-based methods were trained on the BSD.
%are reported in the supp. mat.
\revision{We also report results on MoNuSeg \citep{kumar2019multi}, 
a medical dataset consisting of 14  1000x1000 images of cell nuclei.}

\subsubsection{Relevant validation framework}

\begin{figure*}[t!]
\centering
{\scriptsize
 \begin{tabular}{@{\hspace{0mm}}c@{\hspace{2mm}}c@{\hspace{2mm}}c@{\hspace{2mm}}c@{\hspace{2mm}}c@{\hspace{0mm}}}
 \rotatebox{90}{\hspace{1cm}{BSD}}&
\includegraphics[width=0.23\textwidth,height=0.15\textwidth]{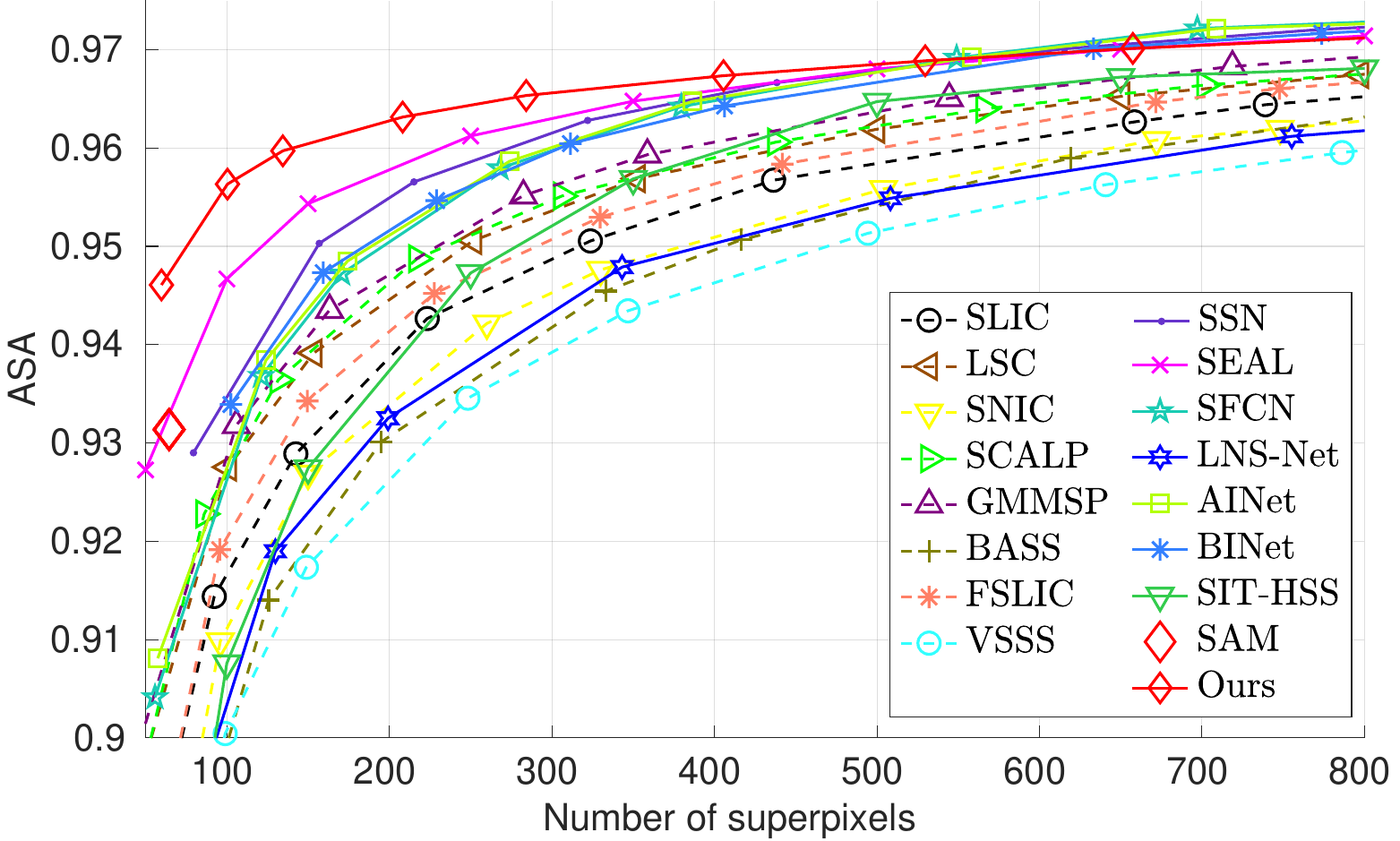}&
\includegraphics[width=0.23\textwidth,height=0.15\textwidth]{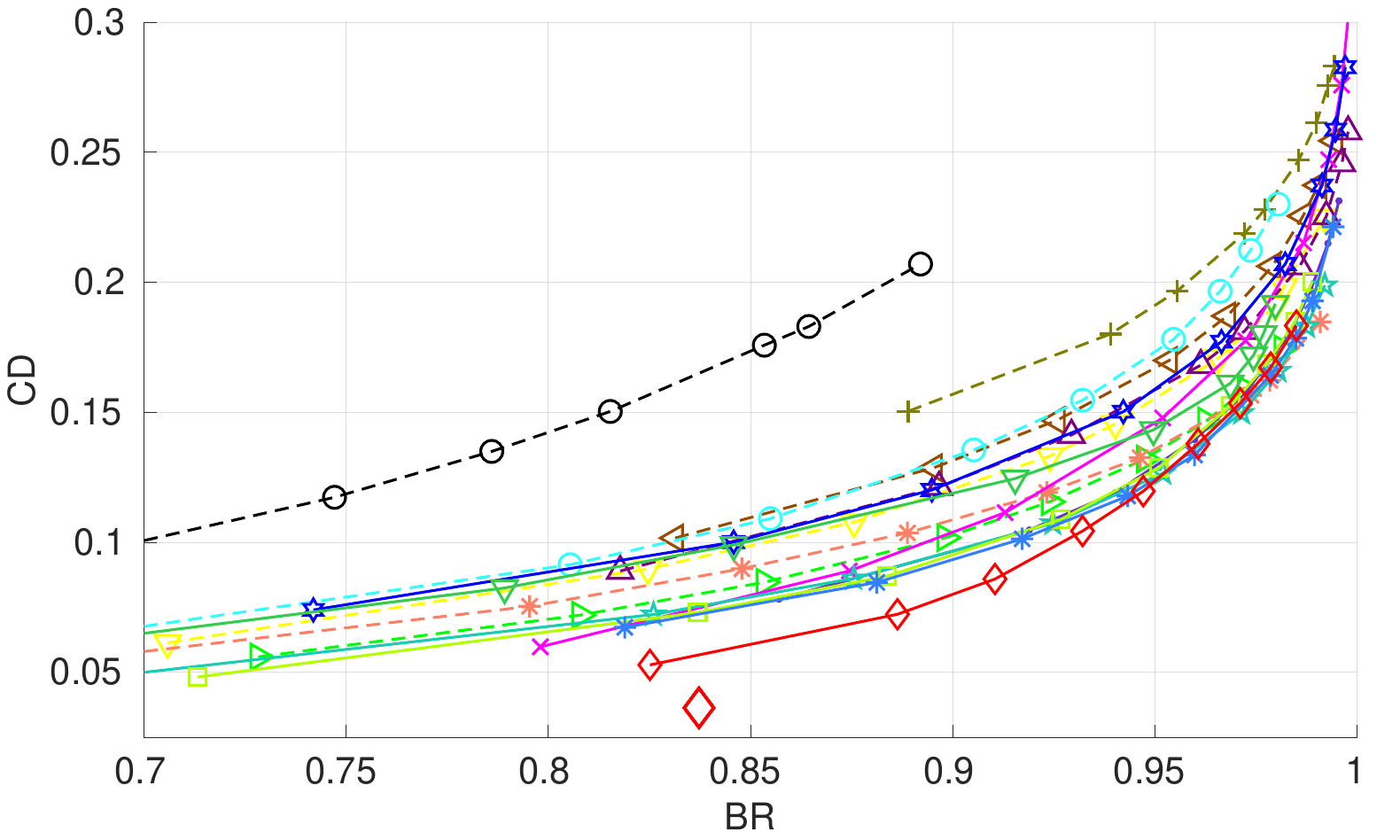}&
%(a) Object segmentation (ASA) & (b)  Contour detection (BR vs CD)   \\[2ex]
\includegraphics[width=0.23\textwidth,height=0.15\textwidth]{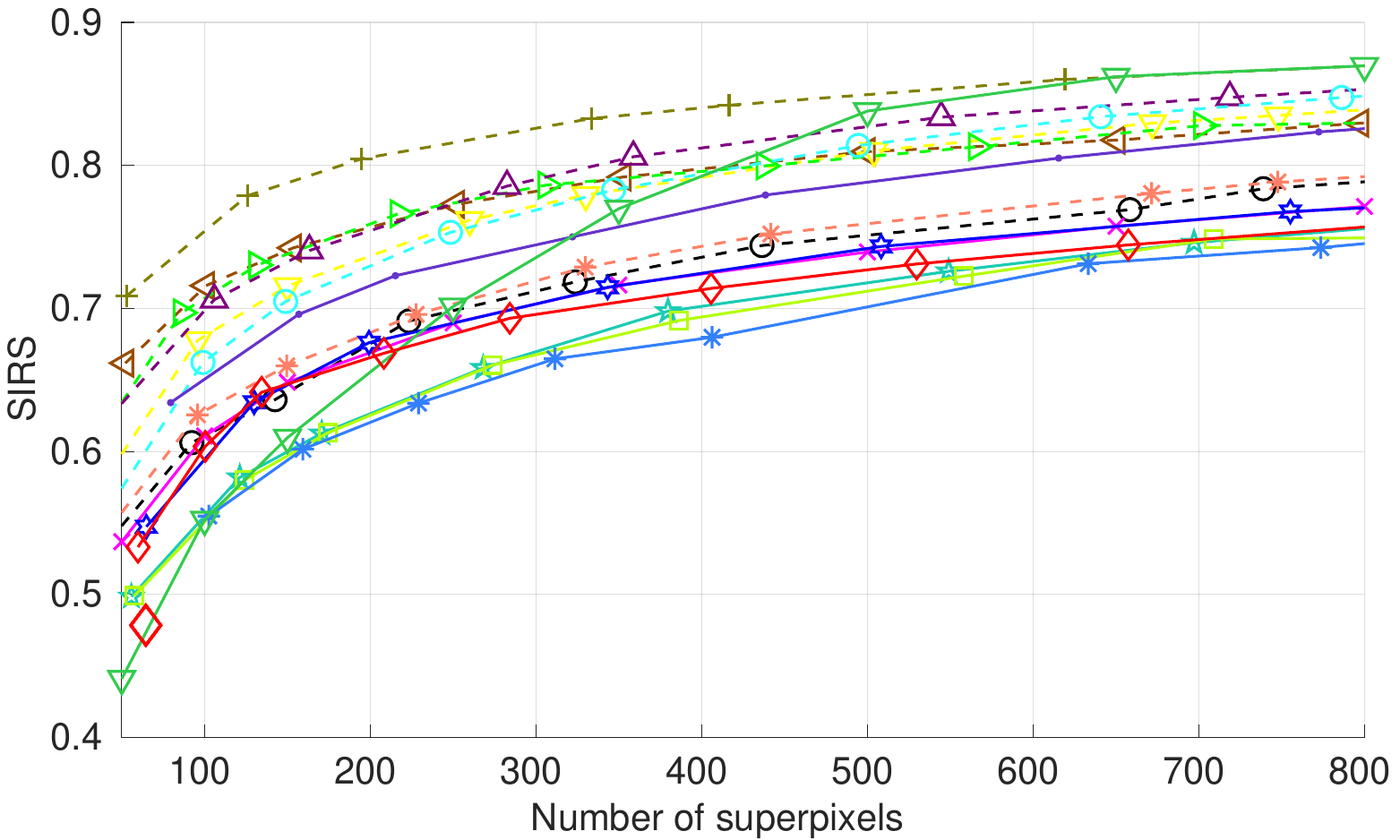}&
\includegraphics[width=0.23\textwidth,height=0.15\textwidth]{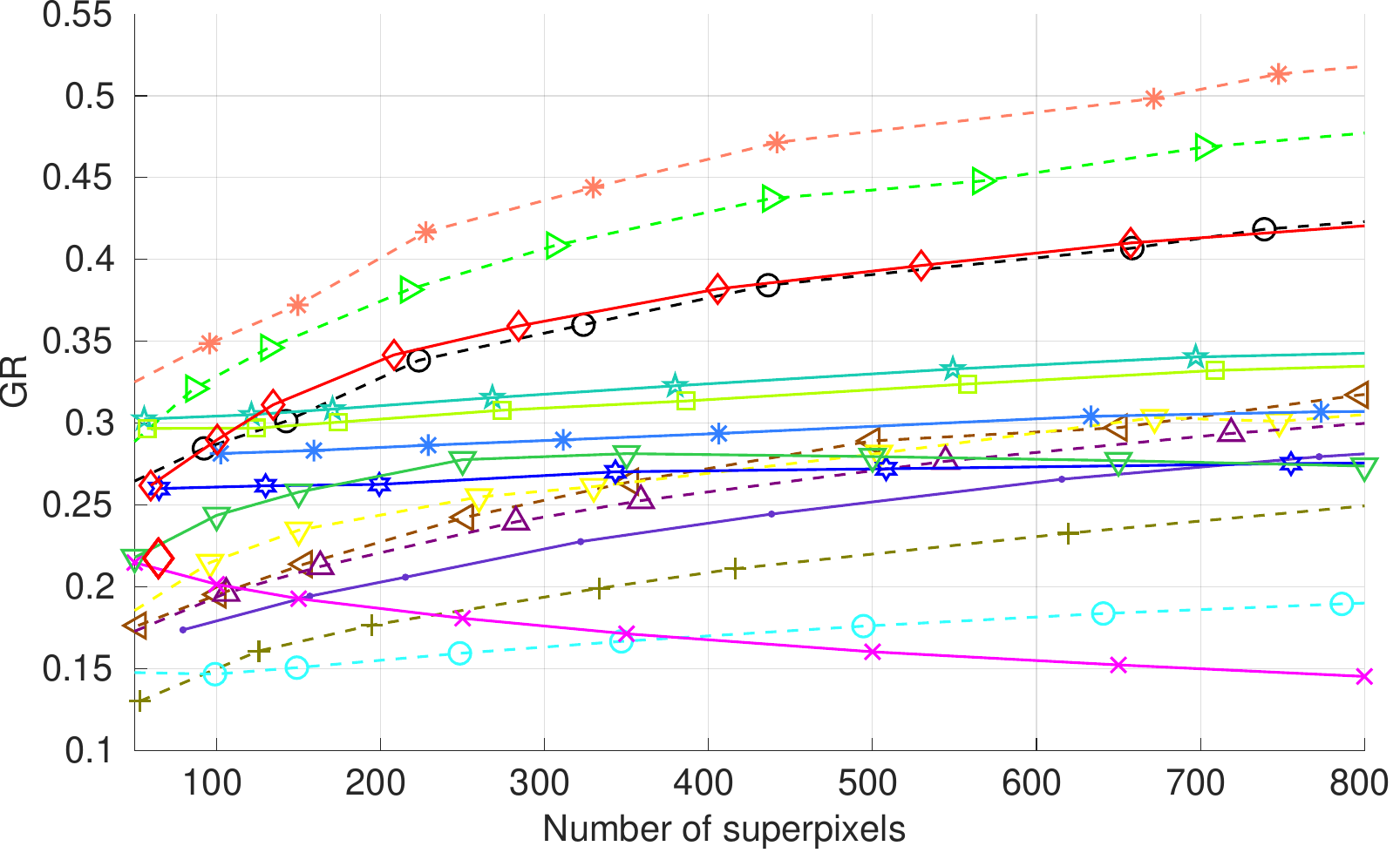}\\ %[-1ex]
 \rotatebox{90}{\hspace{1cm}{NYUV2}}&
\includegraphics[width=0.23\textwidth,height=0.15\textwidth]{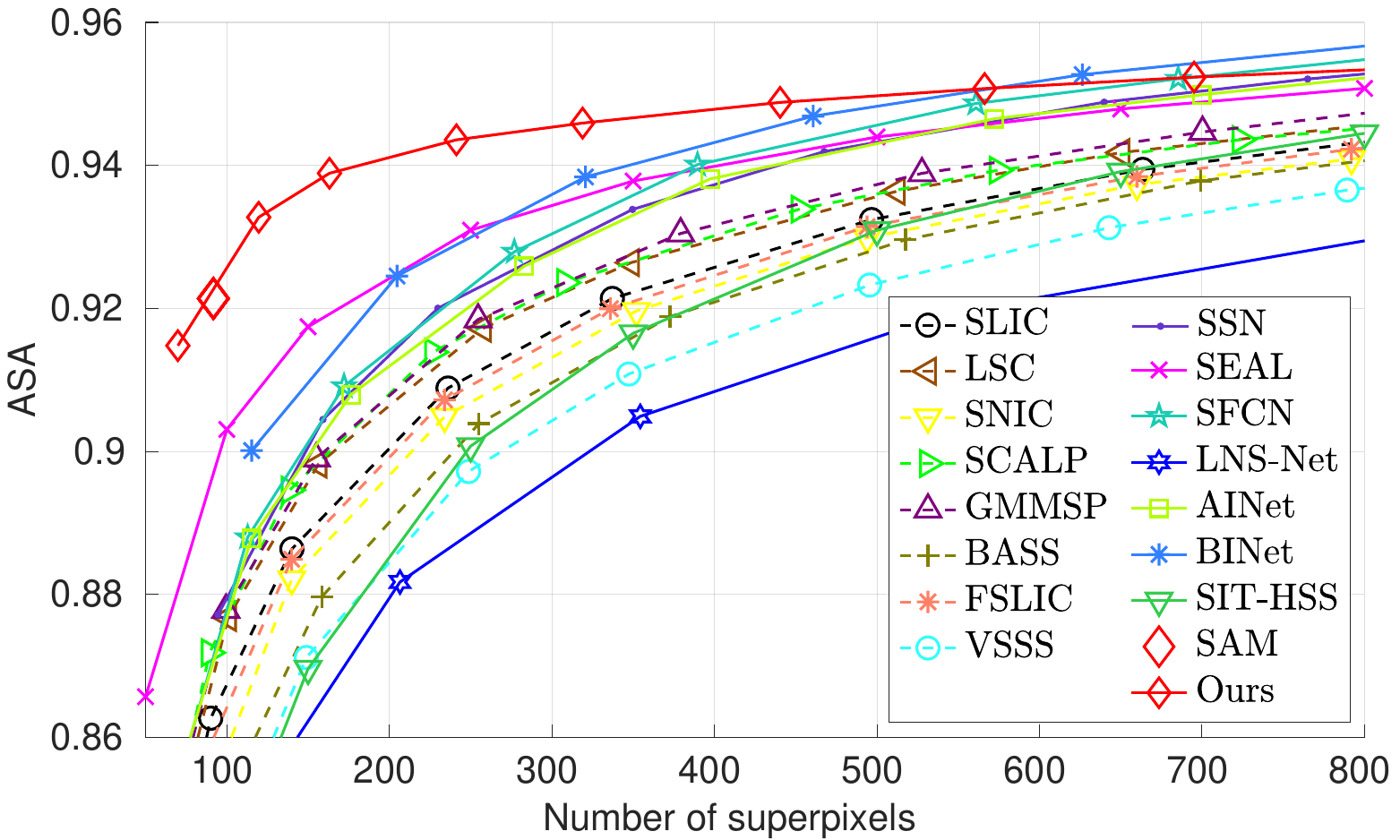}&
\includegraphics[width=0.23\textwidth,height=0.15\textwidth]{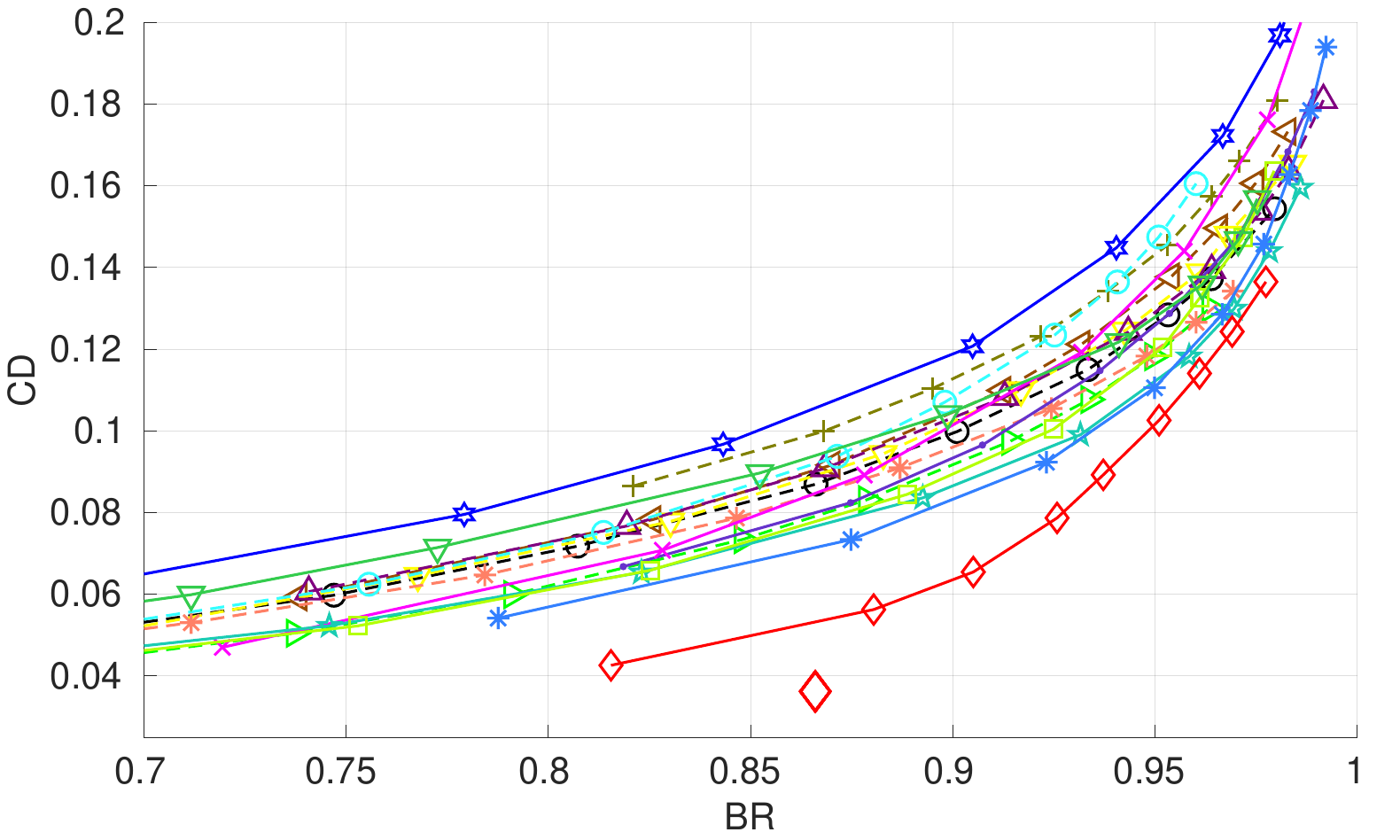}&
%(a) Object segmentation (ASA) & (b)  Contour detection (BR vs CD)   \\[2ex]
\includegraphics[width=0.23\textwidth,height=0.15\textwidth]{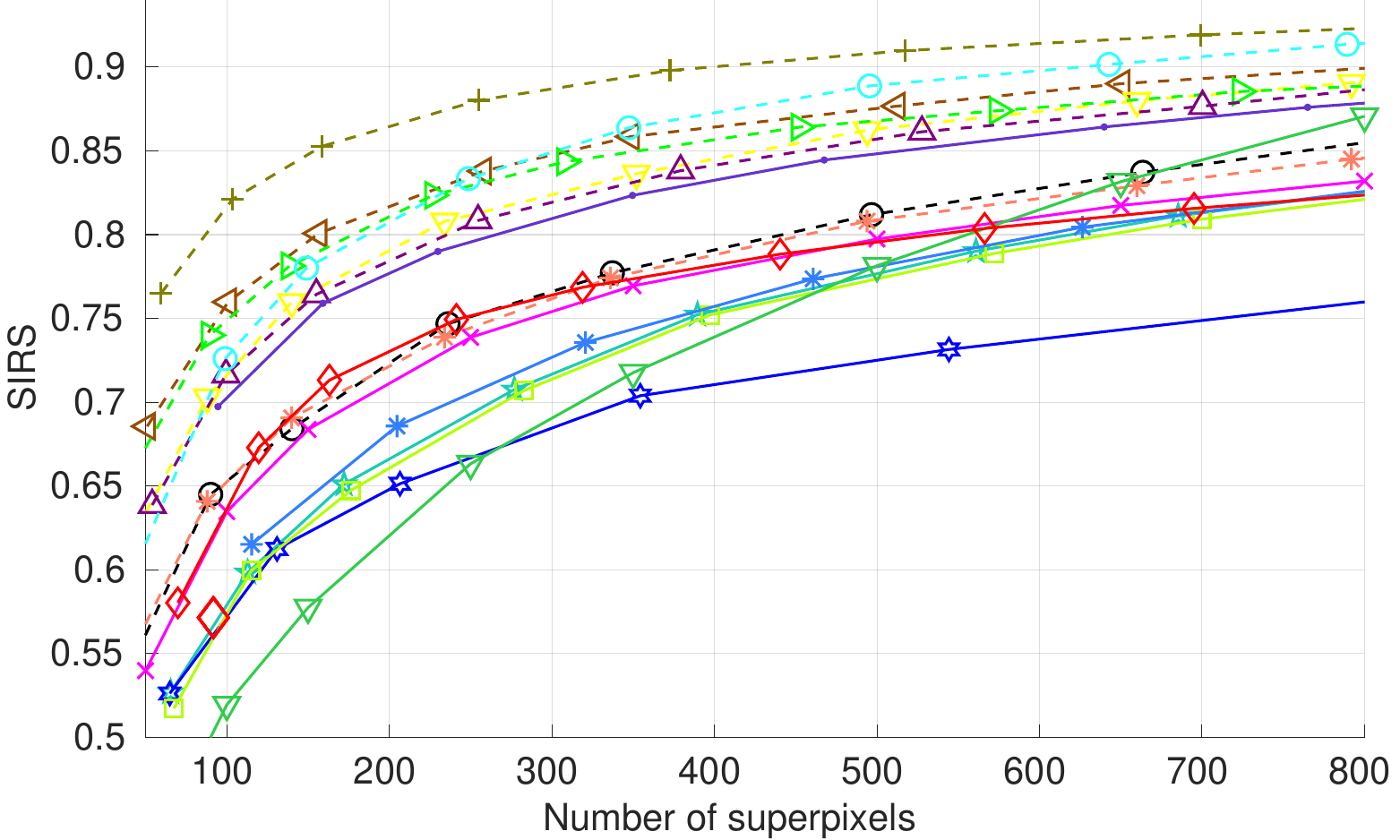}&
\includegraphics[width=0.23\textwidth,height=0.15\textwidth]{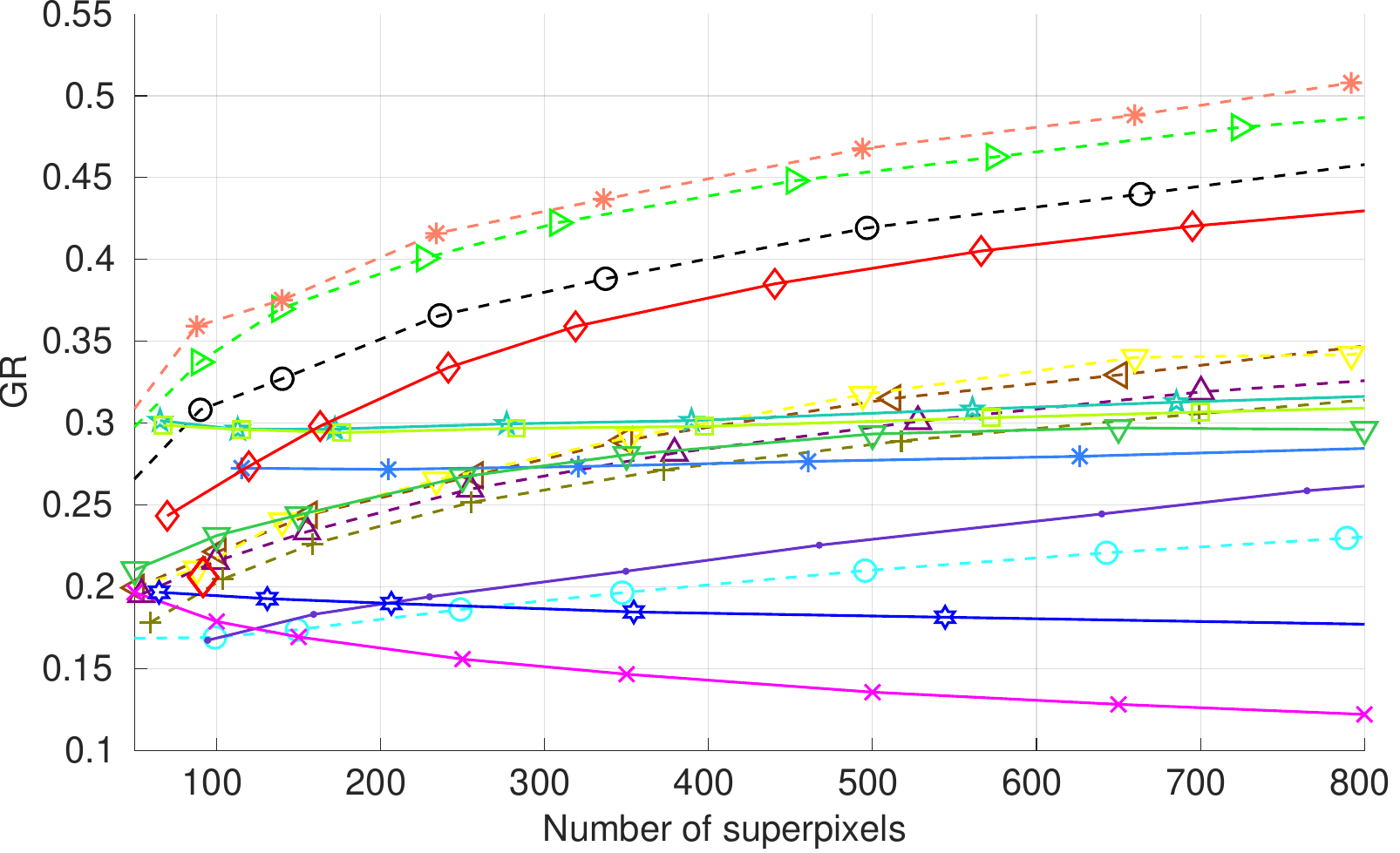}\\ %[-1ex]
 \rotatebox{90}{\hspace{1cm}{SBD}}&
\includegraphics[width=0.23\textwidth,height=0.15\textwidth]{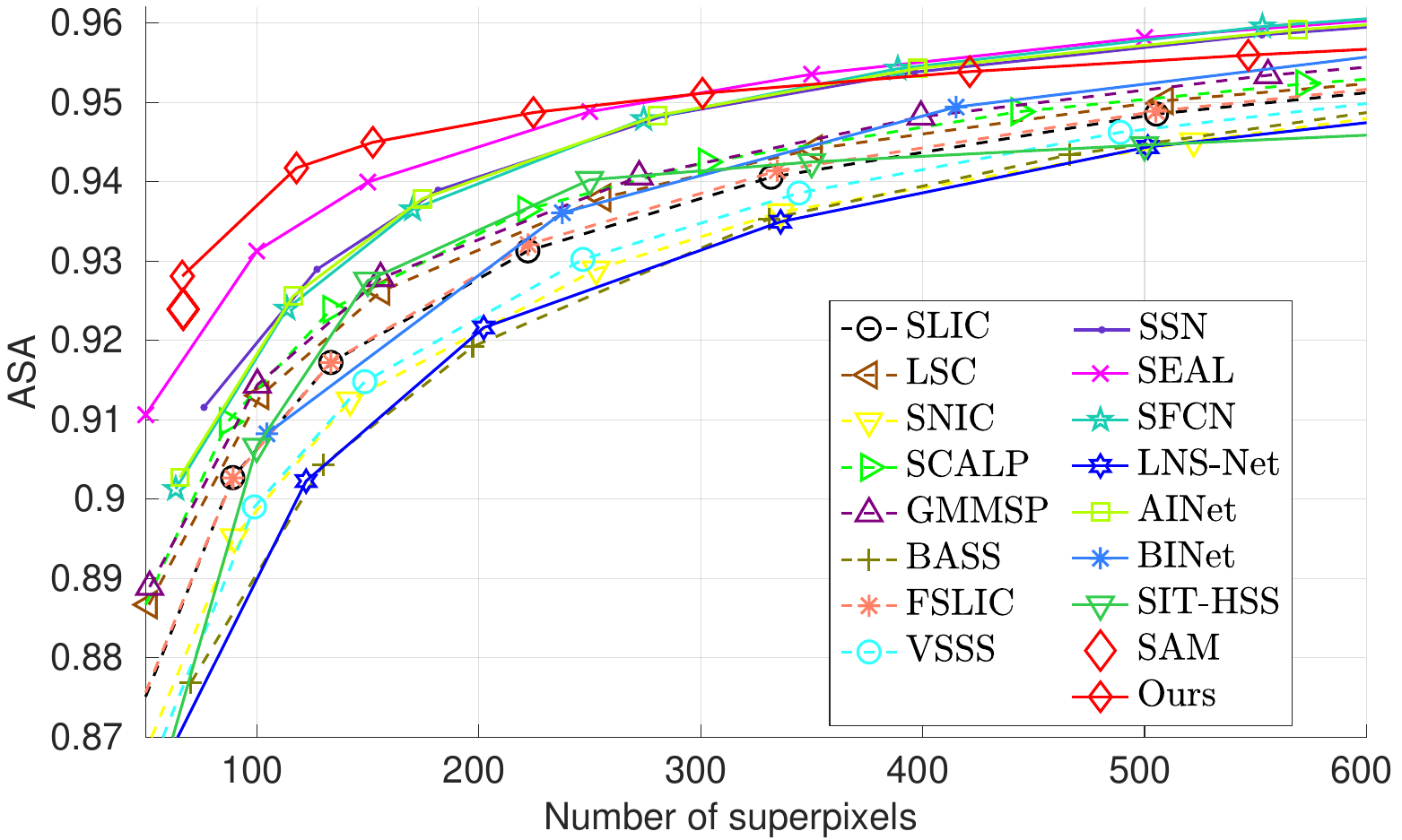}&
\includegraphics[width=0.23\textwidth,height=0.15\textwidth]{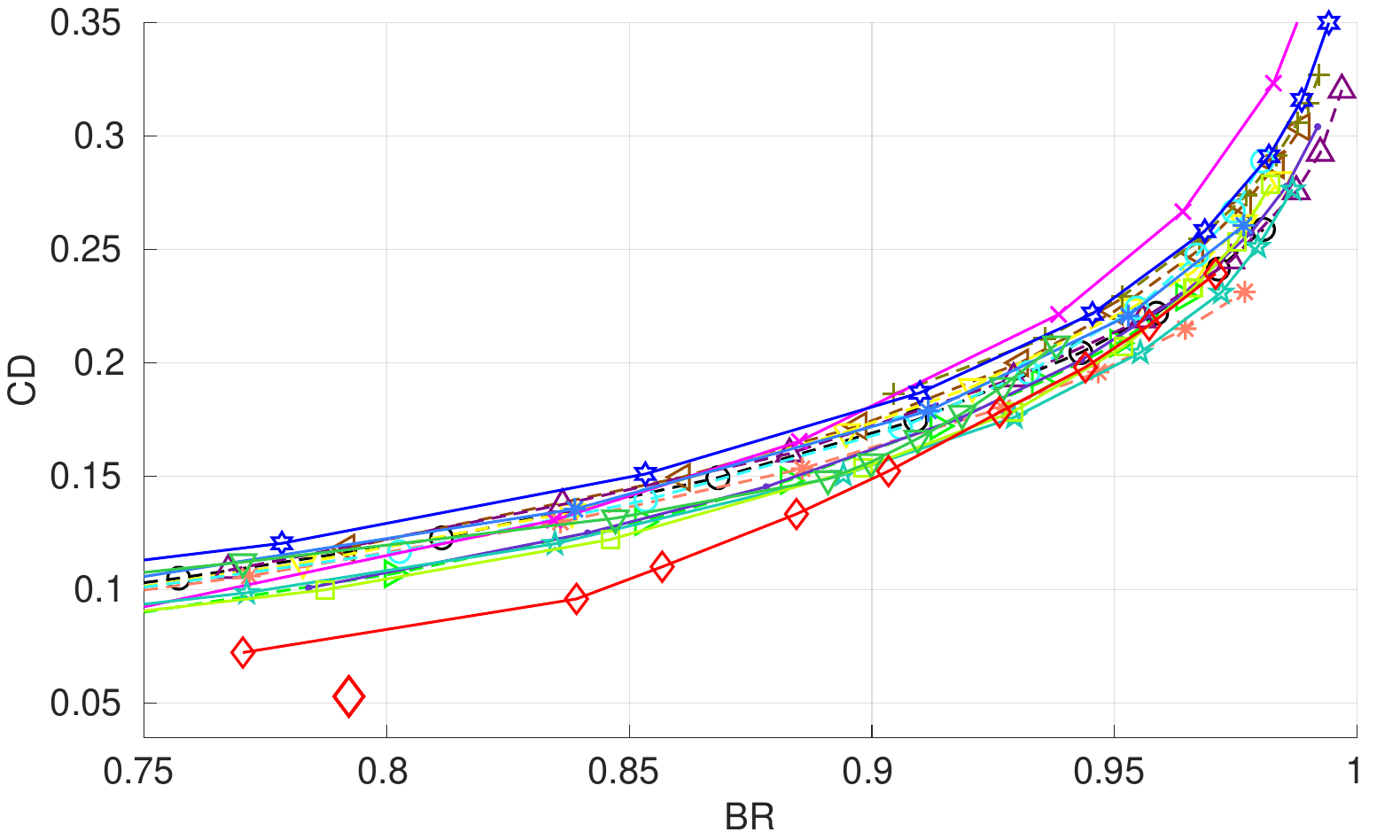}&
%(a) Object segmentation (ASA) & (b)  Contour detection (BR vs CD)   \\[2ex]
\includegraphics[width=0.23\textwidth,height=0.15\textwidth]{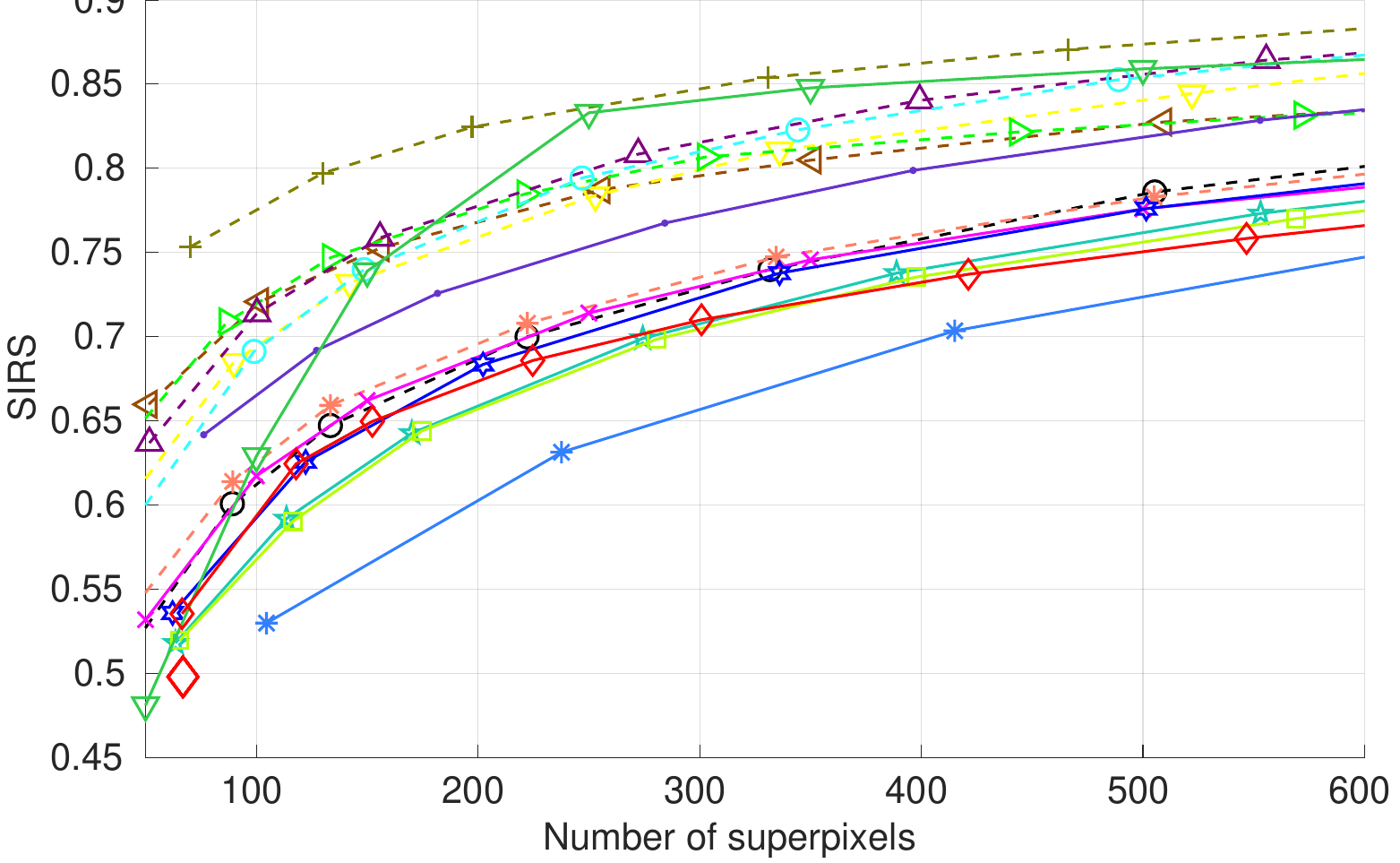}&
\includegraphics[width=0.23\textwidth,height=0.15\textwidth]{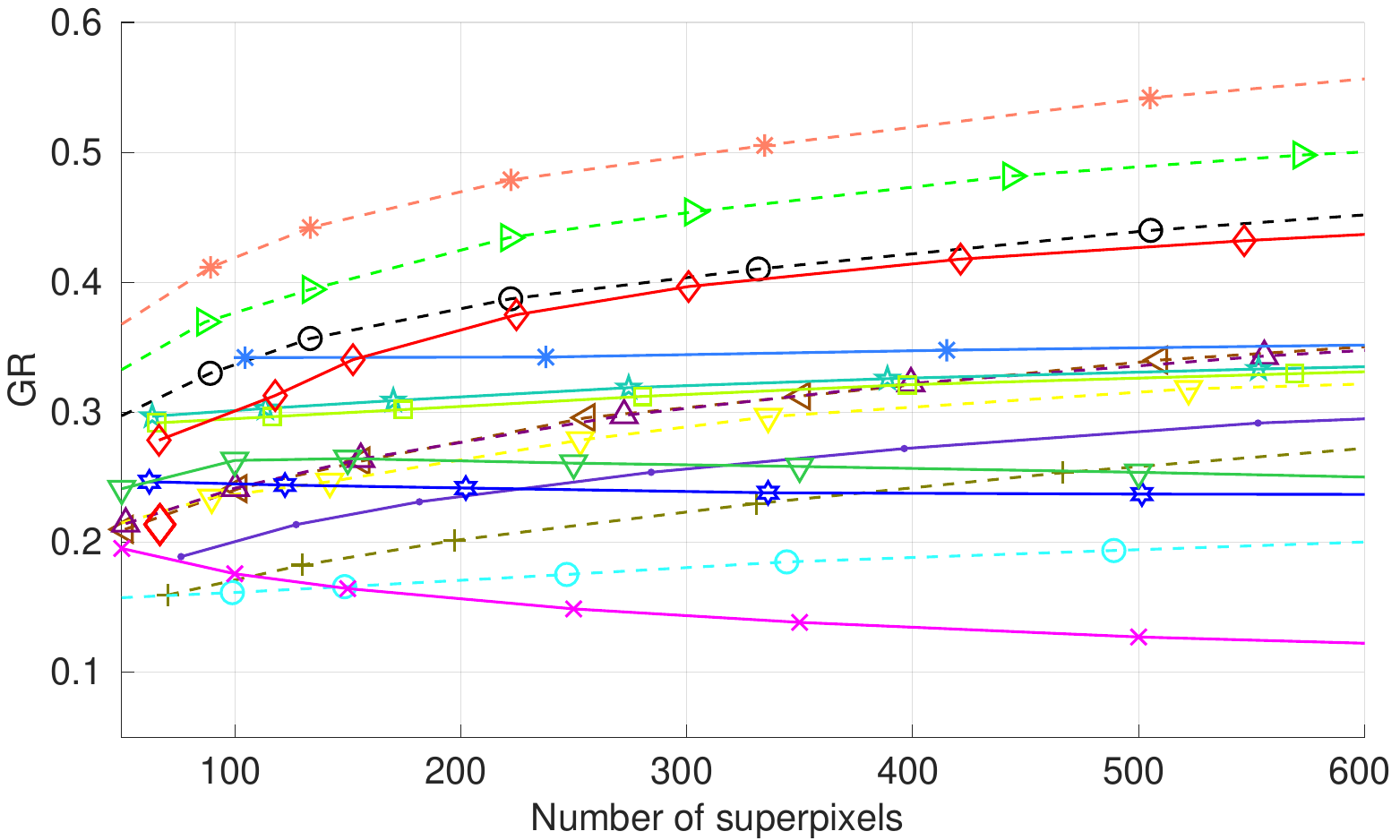}\\ %[-1ex],height=0.13\textwidth
%(c) Color homogeneity (EV)  & (d) Regularity (GR)  \\
\end{tabular}}
\caption{\textbf{Quantitative evaluation of superpixel methods on natural image datasets.} 
%Performance are reported as described in Section \ref{subsec:metrics}.
Traditional (-\hspace{0.15mm}-) and DL-based (--) methods
performance are reported on metrics recommended in Section \ref{subsec:metrics}.
%and DL-based methods are respectively illustrated with dotted and plain lines.
Our approach has the best object segmentation accuracy (ASA) with very accurate contour detection (BR vs CD). SIRS scores are consistent with other DL-based method while our regularity (GR) is among the highest and the best of DL-based methods. %\vspace{-0.2cm}
} 
  \label{fig:res_quanti}
  \end{figure*}

\begin{figure*}[t!]
\centering
{\scriptsize
 \begin{tabular}{@{\hspace{0mm}}c@{\hspace{2mm}}c@{\hspace{2mm}}c@{\hspace{2mm}}c@{\hspace{2mm}}c@{\hspace{0mm}}}
 \rotatebox{90}{\hspace{1cm}{MoNuSeg}}&
\includegraphics[width=0.23\textwidth,height=0.15\textwidth]{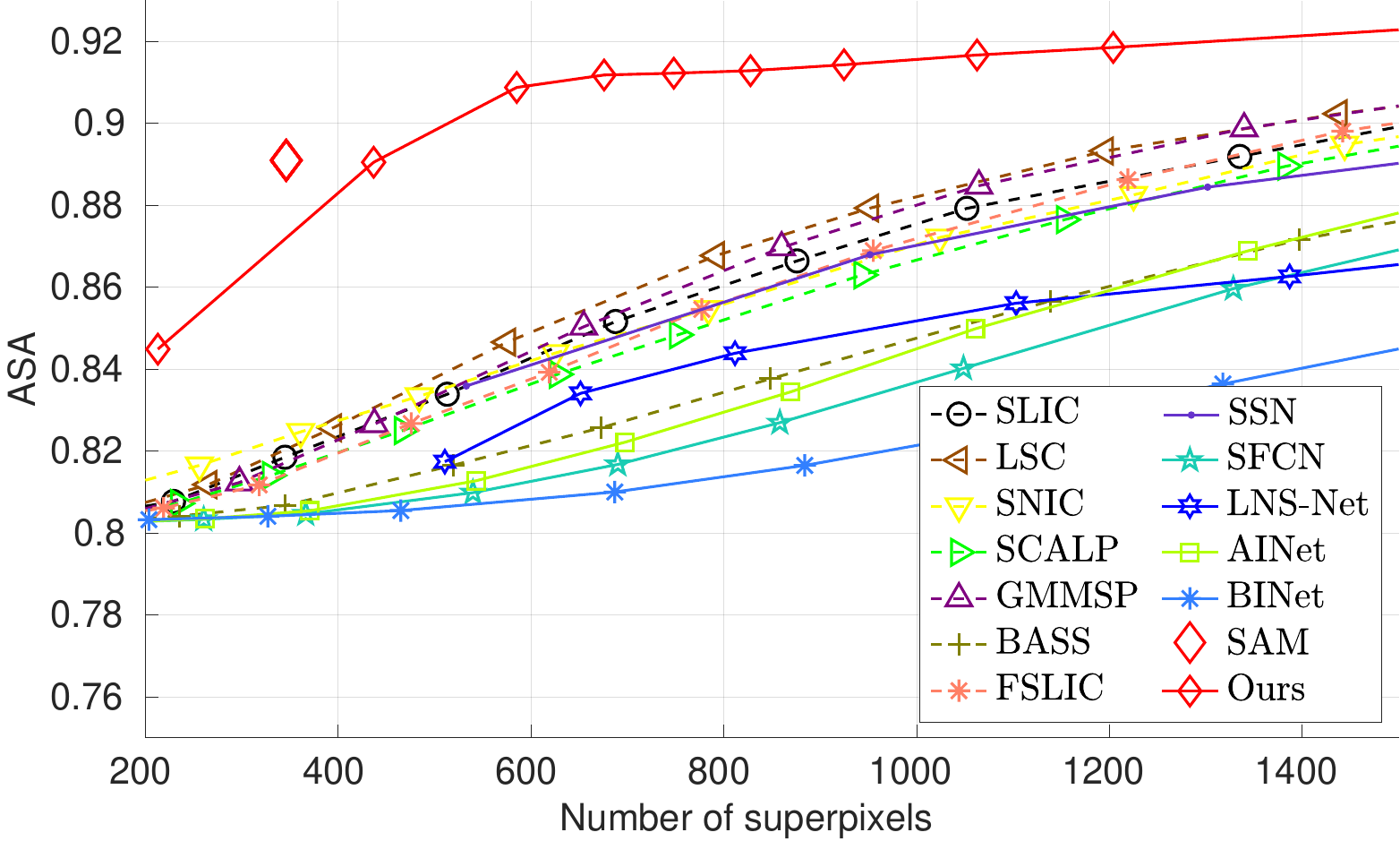}&
\includegraphics[width=0.23\textwidth,height=0.15\textwidth]{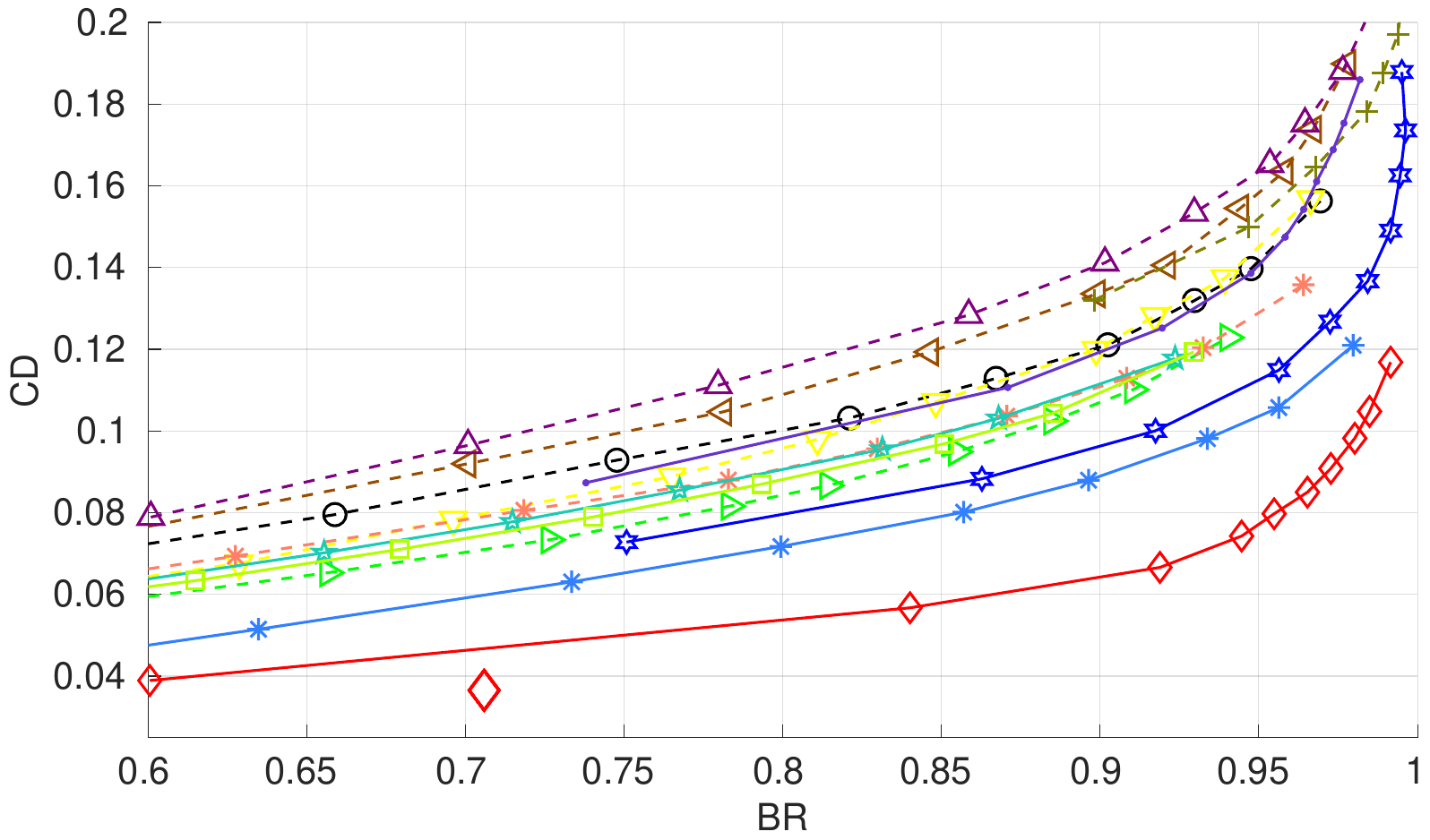}&
%(a) Object segmentation (ASA) & (b)  Contour detection (BR vs CD)   \\[2ex]
\includegraphics[width=0.23\textwidth,height=0.15\textwidth]{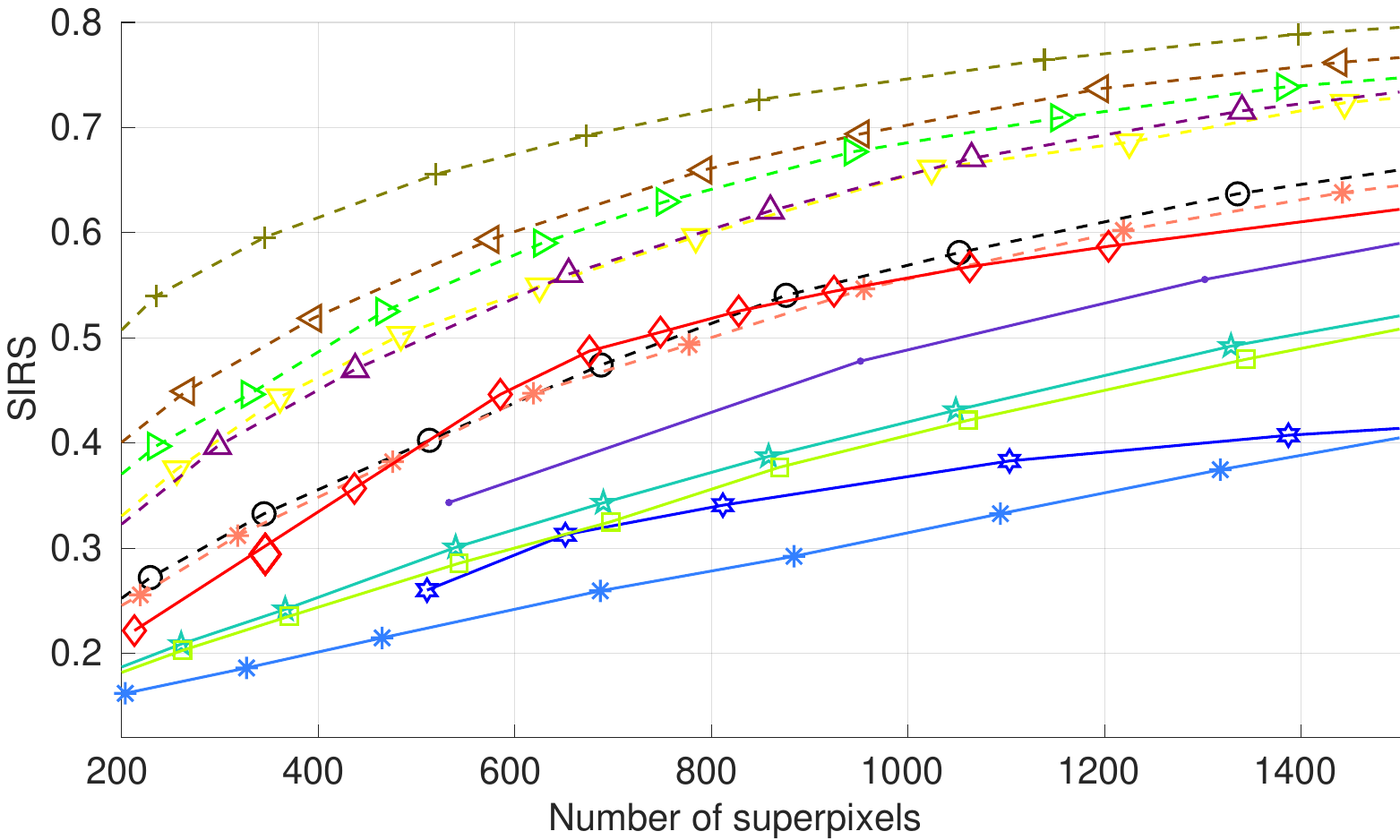}&
\includegraphics[width=0.23\textwidth,height=0.15\textwidth]{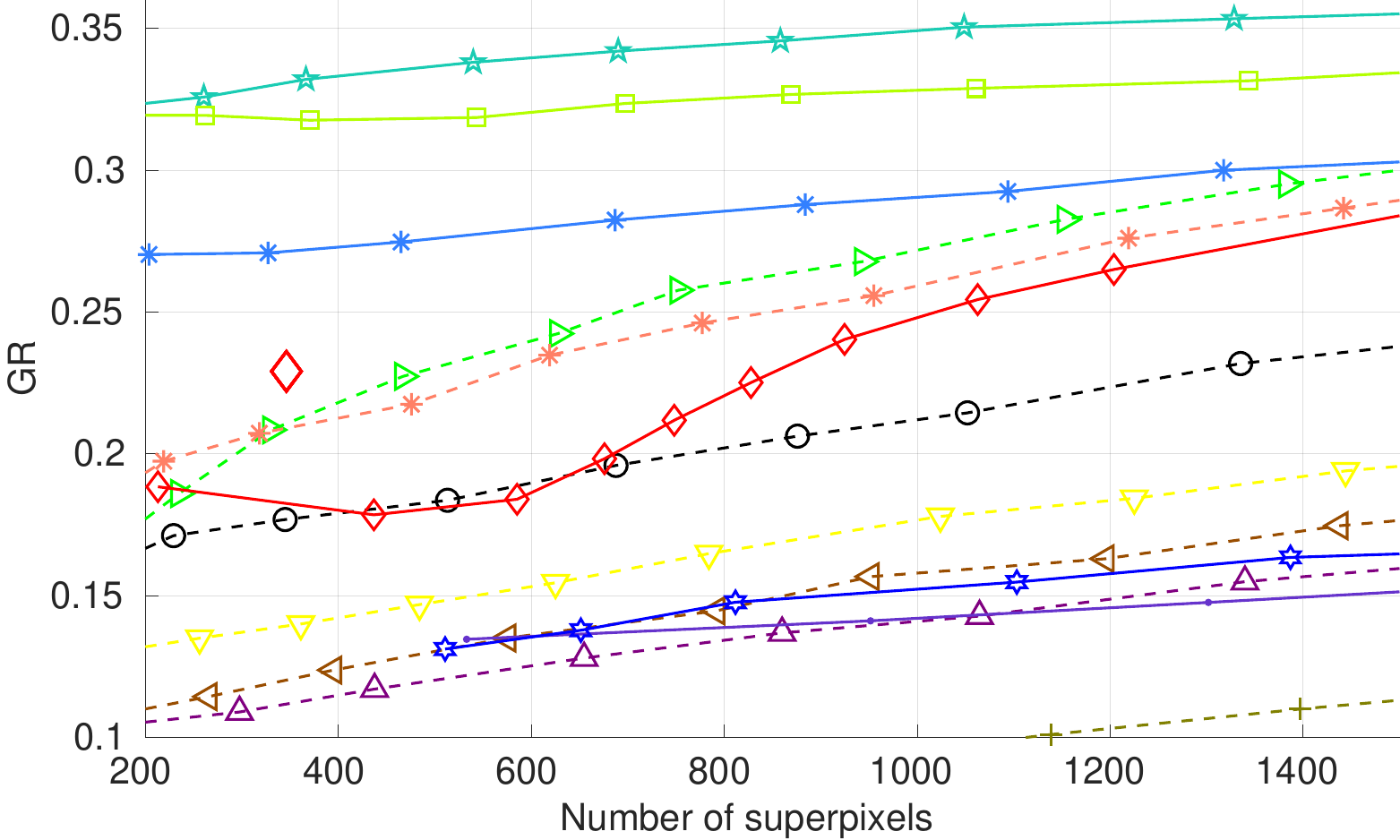}\\ %[-1ex]
\end{tabular}}
\caption{\revision{\textbf{Quantitative evaluation of superpixel methods on medical images (MoNuSeg dataset).} 
%Performance are reported as described in Section \ref{subsec:metrics}.
Traditional (-\hspace{0.15mm}-) and DL-based (--) methods
performance are reported on metrics recommended in Section \ref{subsec:metrics}.
%and DL-based methods are respectively illustrated with dotted and plain lines.
Note that some methods cannot process such images of size 1000x1000 pixels.
Our approach is significantly more accurate than the other methods, 
demonstrating its potential generalization capacity to other image domains.} %\vspace{-0.2cm}
} 
  \label{fig:monuseg}
  \end{figure*}

\begin{figure}[t!]
\centering
{\footnotesize
 \begin{tabular}{@{\hspace{0mm}}c@{\hspace{4mm}}c@{\hspace{0mm}}}
\includegraphics[width=0.39\textwidth, height=0.27\textwidth]{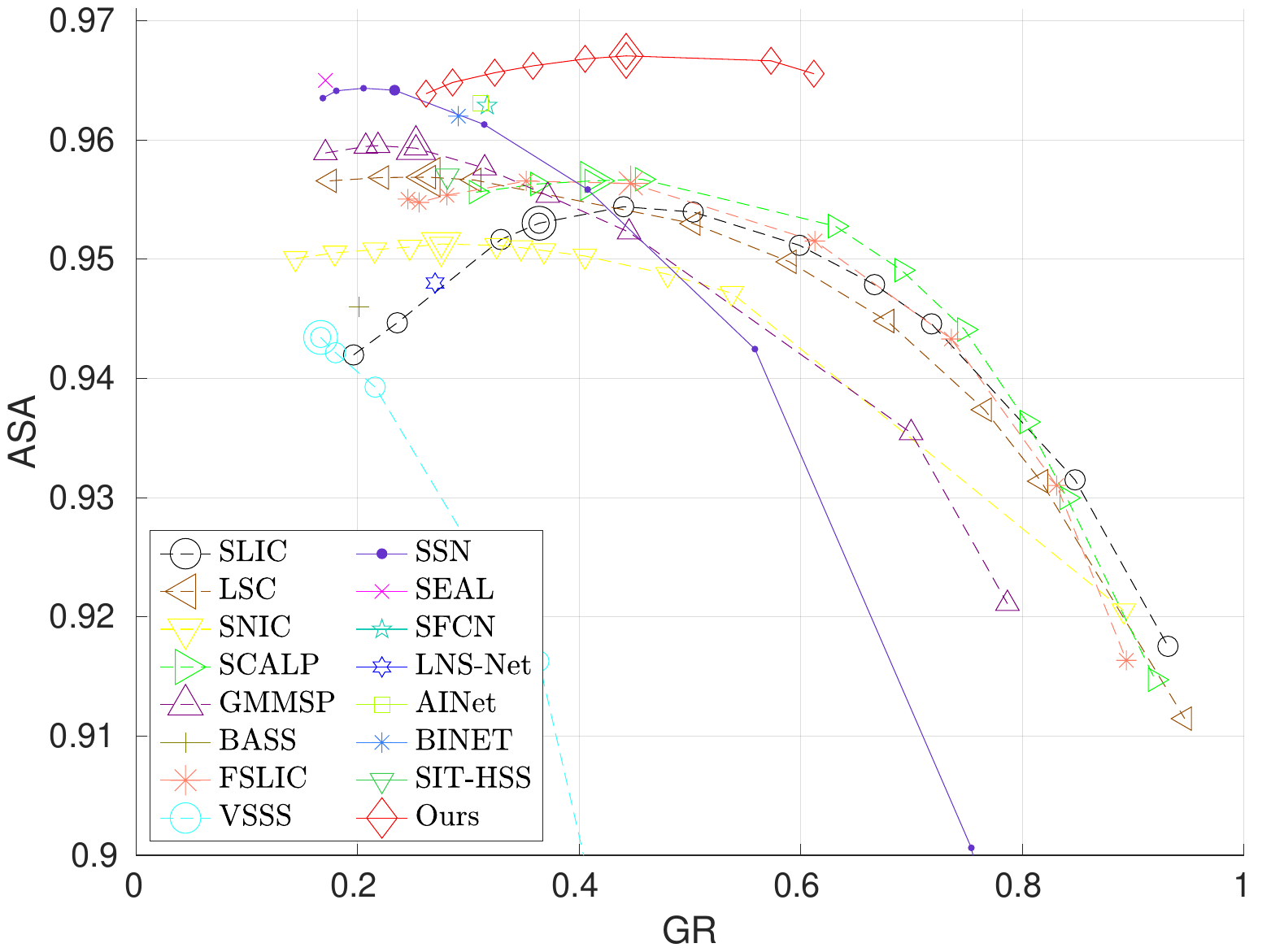}\\
\end{tabular}}
\caption{\textbf{Evaluation of ASA performance for several regularity levels.} The number of requested superpixels is adapted for each method, so the actual average number of generated superpixels is close to 350. Large markers indicate the performance obtained with default parameters.
  }  \vspace{-0.2cm}
  \label{fig:eval_gr}
  \end{figure}

\begin{figure*}[t!]
\centering
{\scriptsize
 \begin{tabular}{@{\hspace{0mm}}c@{\hspace{2mm}}c@{\hspace{2mm}}c@{\hspace{2mm}}c@{\hspace{2mm}}c@{\hspace{0mm}}}
 \rotatebox{90}{\hspace{0.9cm}{BSD noisy}}&
\includegraphics[width=0.23\textwidth,height=0.15\textwidth]{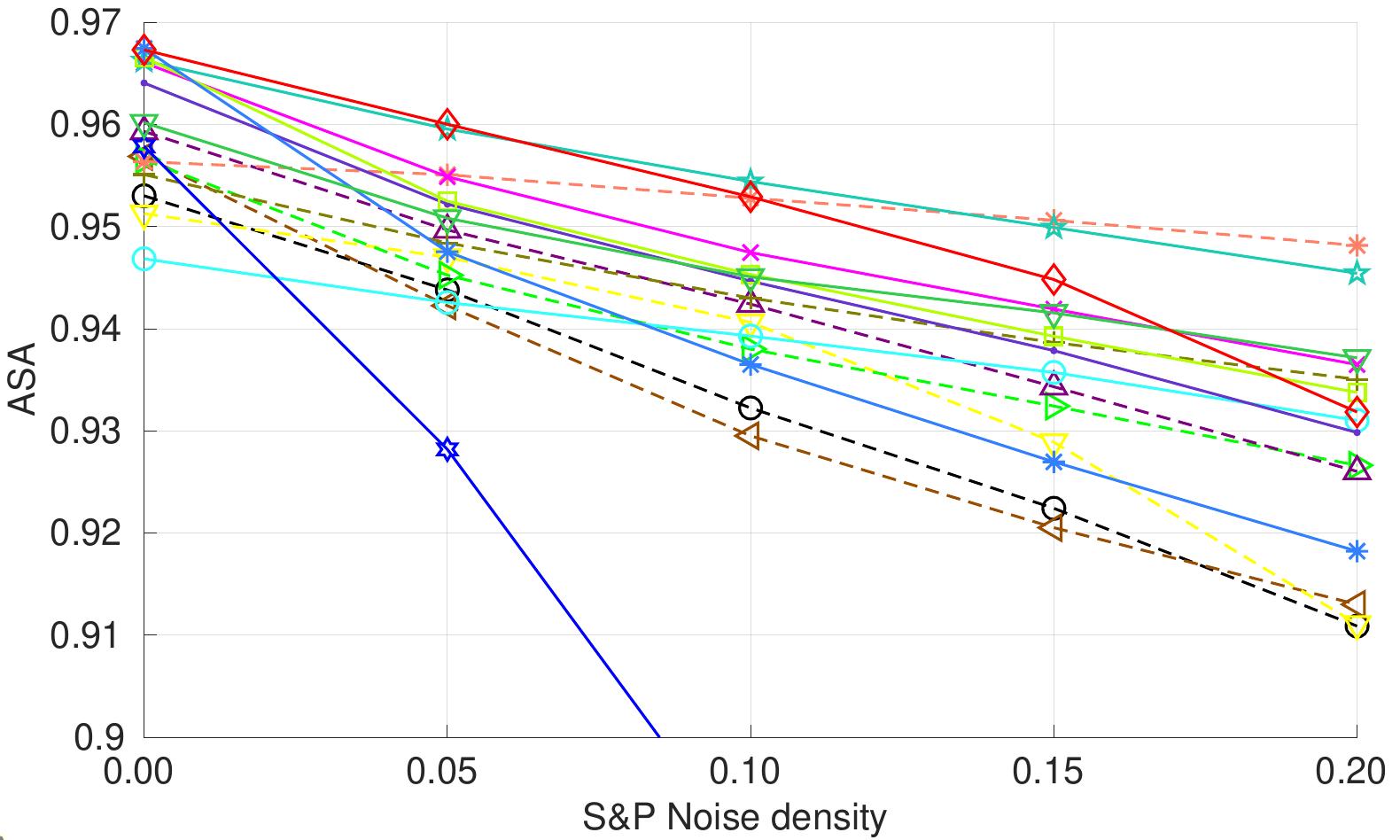}&
\includegraphics[width=0.23\textwidth,height=0.15\textwidth]{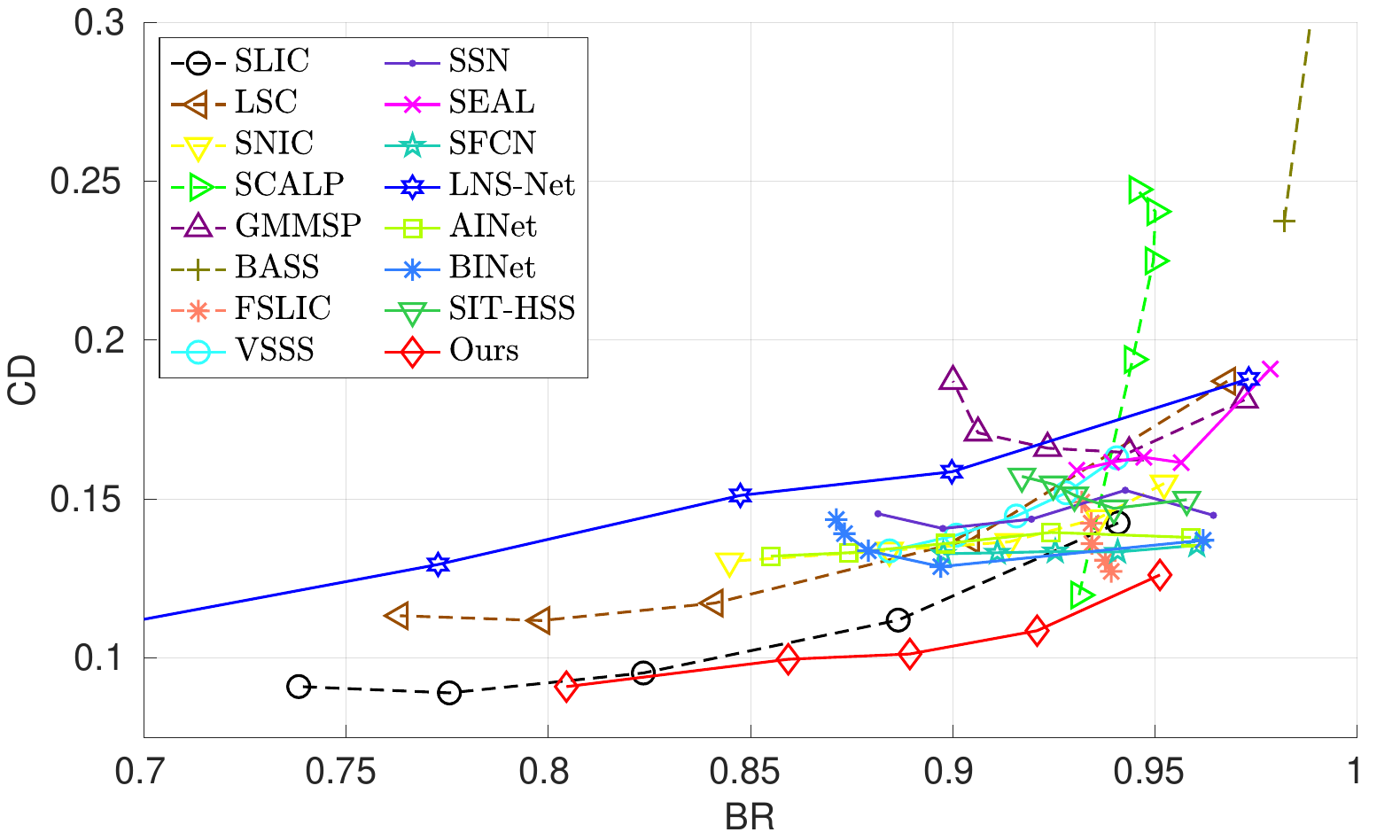}&
%(a) Object segmentation (ASA) & (b)  Contour detection (BR vs CD)   \\[2ex]
\includegraphics[width=0.23\textwidth,height=0.15\textwidth]{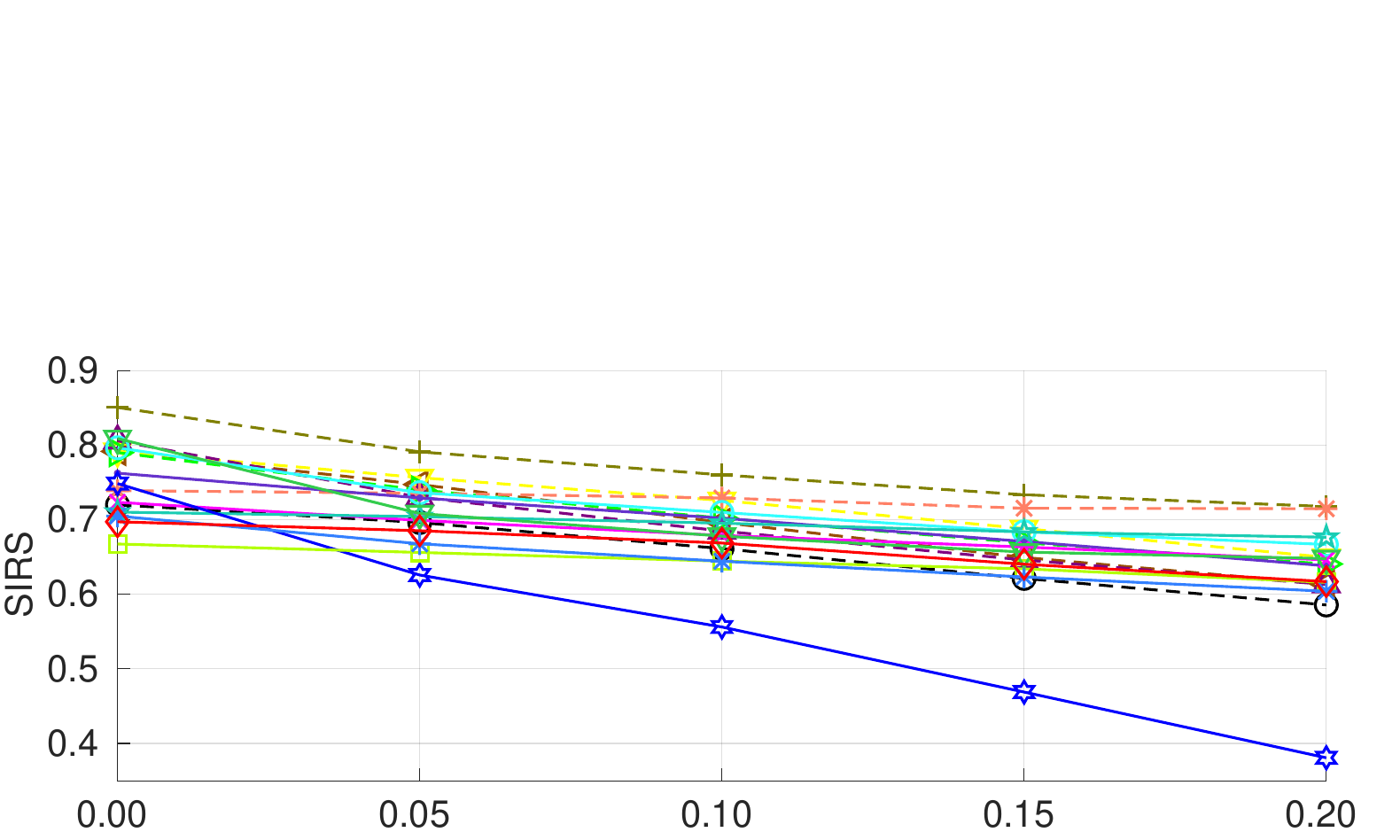}&
\includegraphics[width=0.23\textwidth,height=0.15\textwidth]{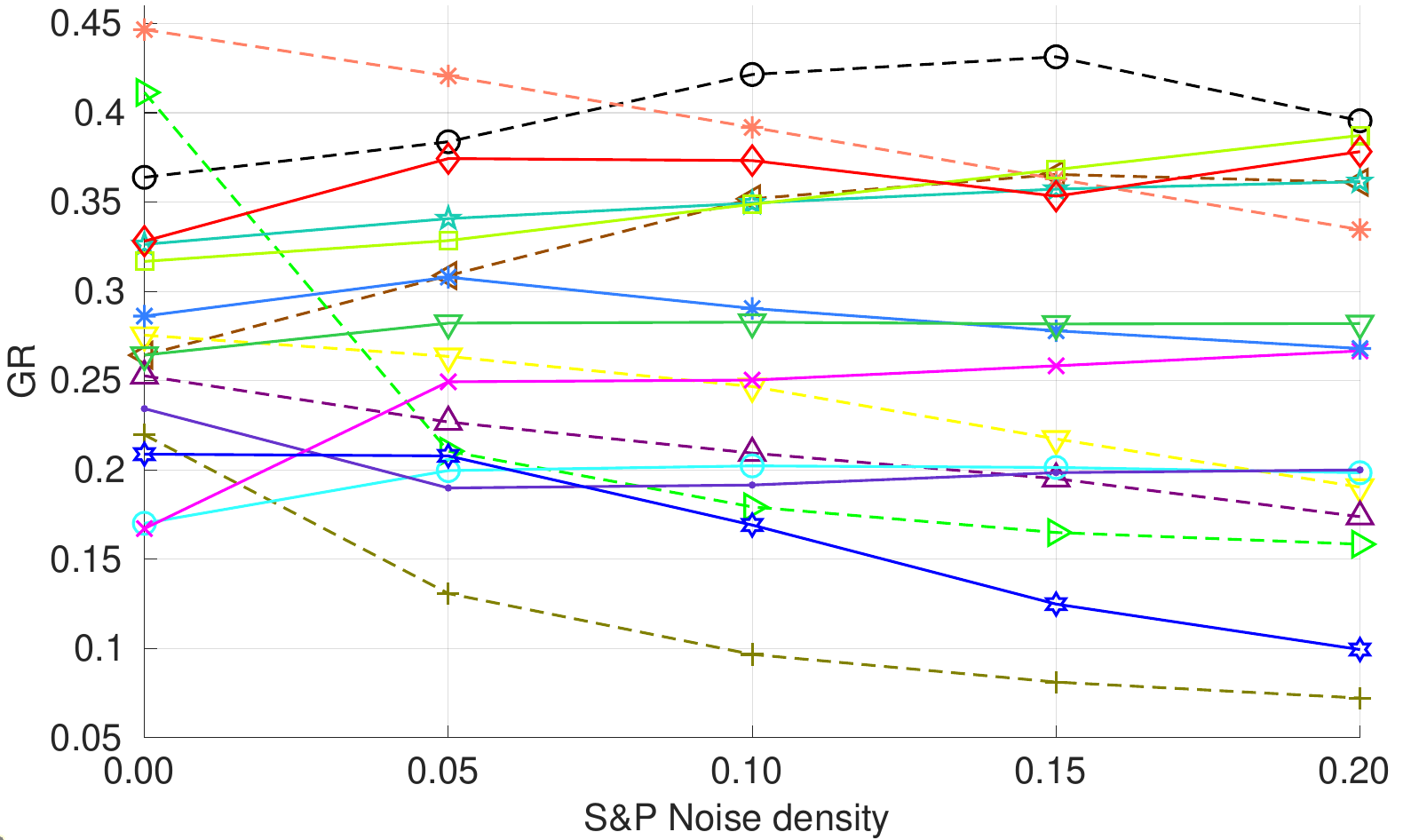}\\ %[-1ex]
\end{tabular}}
\caption{\revision{\textbf{Quantitative evaluation of superpixel methods under impulse noise (Salt-and-Pepper).} 
Results are reported for an approximate number of $K=400$ superpixels and for varying noise density levels.
%Performance are reported as described in Section \ref{subsec:metrics}.
Traditional (-\hspace{0.15mm}-) and DL-based (--) methods
performance are reported on metrics recommended in Section \ref{subsec:metrics}.
%and DL-based methods are respectively illustrated with dotted and plain lines.} 
}%\vspace{-0.2cm}
} 
  \label{fig:sp_noise}
  \end{figure*}

As demonstrated in Section \ref{subsec:metrics}, 
a relevant way to compare the performance of superpixel methods is to report
ASA, BR vs CD, \revision{SIRS} and GR scores according to the average number of generated superpixels and using recommended regularity parameters (see Figures \ref{fig:res_quanti} \revision{and \ref{fig:monuseg}}).
We also report ASA performance for several regularity levels 
as recommended in Section \ref{subsubsec:gr} in Figure \ref{fig:eval_gr}.
\revision{We also report in Figure~\ref{fig:sp_noise} the evaluation of performance 
on images affected by impulse noise}.
Finally qualitative evaluation with segmentation examples are given in Figure \ref{fig:res_quali}. 
% and supp. mat.).
%\ref{fig:nyuv2_ex} and \ref{fig:sbd_ex}).  

\subsubsection{Analysis of performance\label{subsubsec:perf}} 

\paragraph{Quantitative performance.} 
As shown in Sections~\ref{sec:ill_posed} and~\ref{sec:metric}, methods cannot maximize all aspects at once due to the potentially contradictory objectives of superpixel properties.
%ill-posed nature of superpixels.
%
In Figure \ref{fig:res_quanti}, we can see that on all datasets, our approach obtains the best object segmentation performance (ASA), 
which can be considered the main criteria to focus on.
%especially for a low number of superpixel due to the accuracy of the
%initial object segmentation.
Performance on contour detection (BR vs CD) is also among the best ones.
Color homogeneity (SIRS) is generally anticorrelated to accurate object segmentation (cf. Section \ref{subsec:metrics}).
Nonetheless, our method obtains SIRS scores similar to other DL-based methods.
Finally, our approach is
among the best in terms of regularity and particularly has the highest GR compared to the DL-based methods.

\revision{In Figure~\ref{fig:monuseg}, 
we also report results on MoNuSeg \citep{kumar2019multi}, a dataset of cell nuclei images,
to evaluate generalization capacity beyond natural images.
The proposed approach is significantly % an order of magnitude 
more accurate in terms of segmentation accuracy than the compared methods.
While the deep learning-based methods are trained on BSD, 
our approach leverages a large-scale pretrained model 
designed to better generalize and capture complex object structures.
}

The representation in Figure \ref{fig:eval_gr}
enables to assess the potential of each method and their behaviors for extreme regularity levels.
For instance, methods such as SLIC, can get close to a square grid (GR$\approx$1).
The proposed approach is also capable of generating superpixels with several regularity levels, contrary to most deep learning-based methods 
that do not allow to set a variable regularity level 
without a new training process and a dedicated model.

\revision{The clustering of the proposed method takes approximately 100ms on a Nvidia RTX A5000 GPU, 
for a BSD image with $K=500$ superpixels, which is similar to the computational time of SLIC.
Since it is based on maskSLIC, this computational time scales consistently with the number of superpixels as with SLIC.
When using SAM as a prior segmentation model, it takes an additional $\approx$2s for a fine sampling grid of $32{\times}32$.
Note that this prior computation can be reduced by using a coarser grid (cf. Fig~\ref{fig:ablation_sam}). 
For instance, the computational time drops to 0.75s for an 8${\times}$8 grid.
This overall computational time is similar to methods such as  
VSSS %\citep{zhou2023vine} 
(1.6s) or slower deep-learning-based methods such as 
%SEAL %\citep{tu2018learning} 
%(2.3s) or 
LNSNet %\citep{zhu2021learning} 
(1.2s).
Recent deep-learning methods, \emph{e.g.}, SFCN 
%\citep{yang2020superpixel} 
or AINet %\citep{wang2021ainet} 
can produce segmentations in less than 0.1s.
For a more comprehensive evaluation of the computational time of the state-of-the-art methods, we refer the reader to studies such as \citep{barcelos2024comprehensive,wu2024survey}.
Future works will investigate the use of lightweight models such as FastSAM \citep{zhao2023fast} in our method, which reports object segmentation in less than 50ms with very limited performance drop compared to \citep{kirillov23sam}.}

\paragraph{Robustness to noise.} 
\revision{As discussed in Section \ref{subsec:noise}, we advocate prioritizing the evaluation of robustness to impulse noise, \emph{e.g.} salt-and-pepper noise. 
We therefore investigate this aspect in Figure~\ref{fig:sp_noise} 
by considering a fixed intermediate superpixel scale of $K=400$ and several noise densities affecting the BSD images.
Since our method builds on SLIC superpixels, it is also highly subject to noise.
However, thanks to the robustness of the high-level prior from \citep{kirillov23sam}, 
%The results show that 
our approach remains comparatively robust as the noise level increases,
whereas some competing methods experience more significant performance degradation.
%and, in certain cases, collapse.
}

\paragraph{Qualitative examples.}

\begin{figure*}[t!]
\centering
\newcommand{\hhh}{0.1575\textwidth}
{\scriptsize
 \begin{tabular}{@{\hspace{0mm}}c@{\hspace{1mm}}c@{\hspace{1mm}}c@{\hspace{1mm}}c@{\hspace{1mm}}c@{\hspace{1mm}}c@{\hspace{1mm}}c@{\hspace{1mm}}@{\hspace{0mm}}}
\includegraphics[width=\hhh]{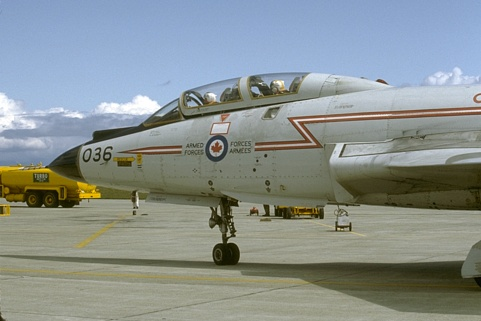}&
\includegraphics[width=\hhh]{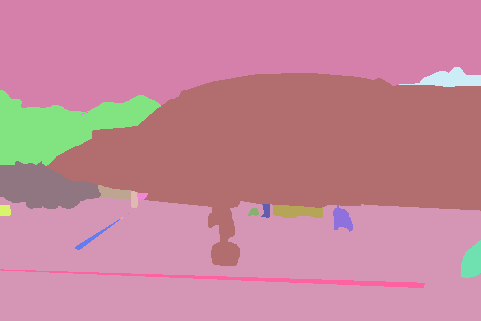}&
\includegraphics[width=\hhh]{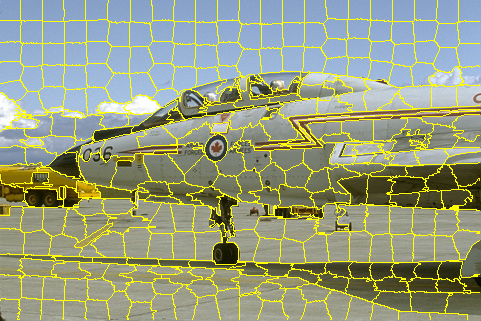}&
\includegraphics[width=\hhh]{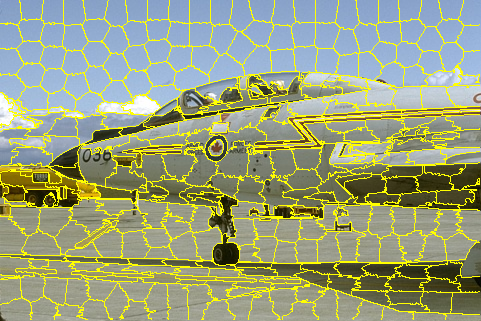}&
\includegraphics[width=\hhh]{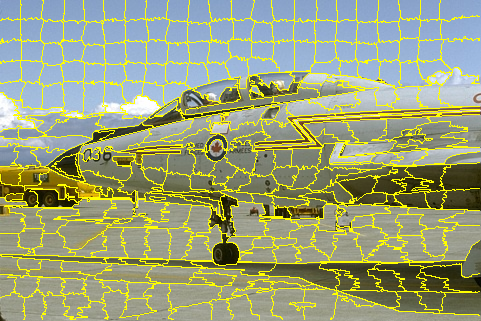}&
\includegraphics[width=\hhh]{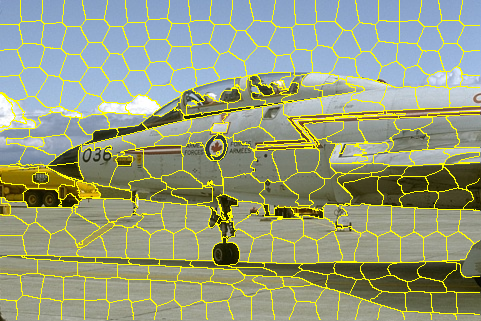}\\
Image & Groundtruth & 
SLIC %\citep{achanta2012} 
& LSC %\citep{li2015} 
& SNIC %\citep{achanta2017superpixels} 
& SCALP
\\[0.5ex]
\includegraphics[width=\hhh]{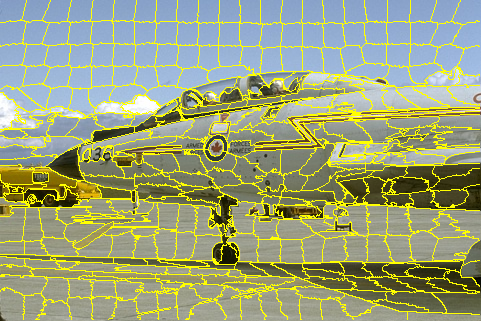}&
\includegraphics[width=\hhh]{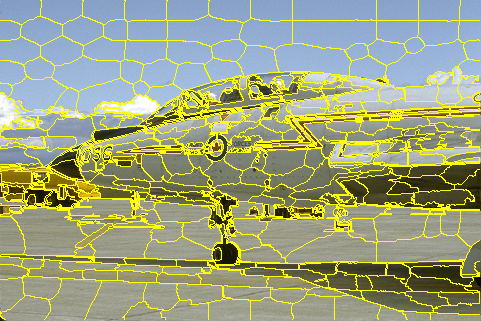}&
\includegraphics[width=\hhh]{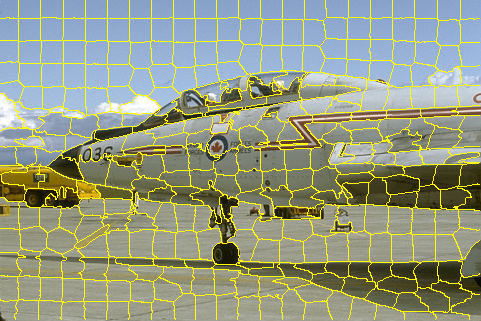}&
\includegraphics[width=\hhh]{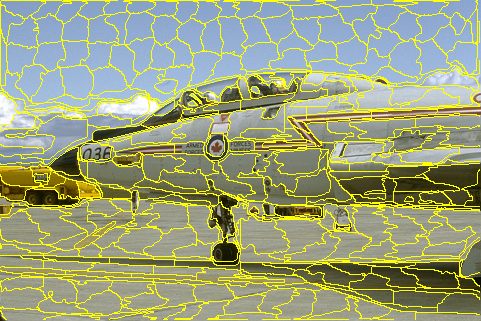}&
\includegraphics[width=\hhh]{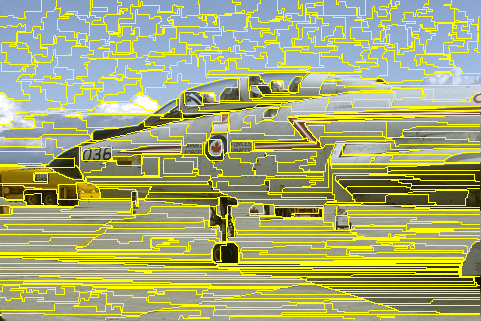}&
\includegraphics[width=\hhh]{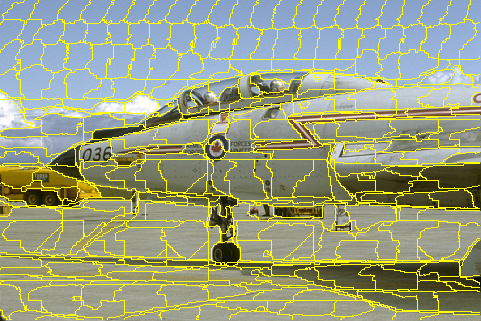}\\
GMMSP %\citep{Ban18} 
& BASS %\citep{Uziel:ICCV:2019:BASS} 
& FSLIC %\citep{wu2020fuzzy}
& SSN %\citep{jampani2018superpixel}
& SEAL
& SFCN %\citep{yang2020superpixel} 
 \\[0.5ex]
\includegraphics[width=\hhh]{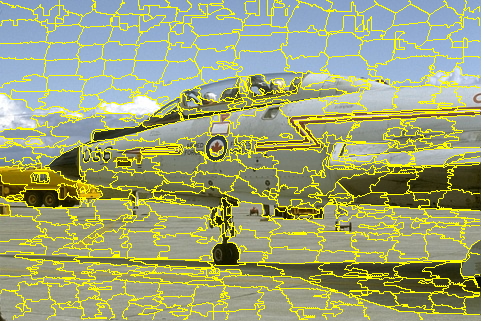}&
\includegraphics[width=\hhh]{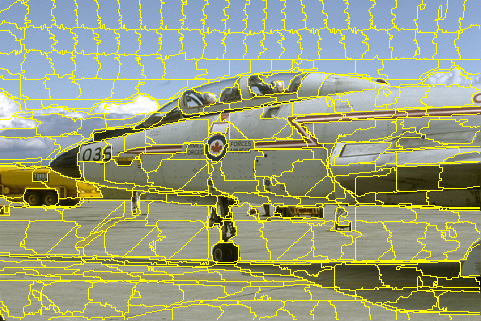}&
\includegraphics[width=\hhh]{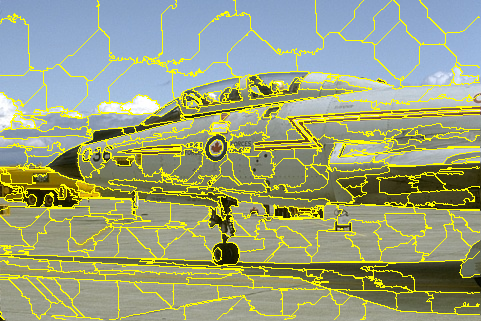}&
\includegraphics[width=\hhh]{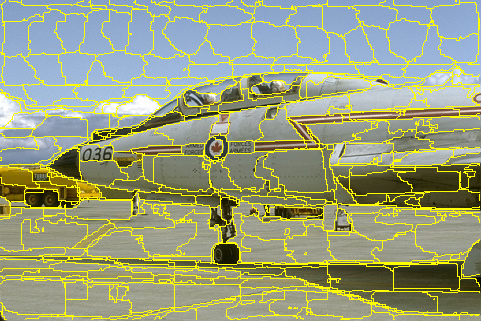}&
\includegraphics[width=\hhh]{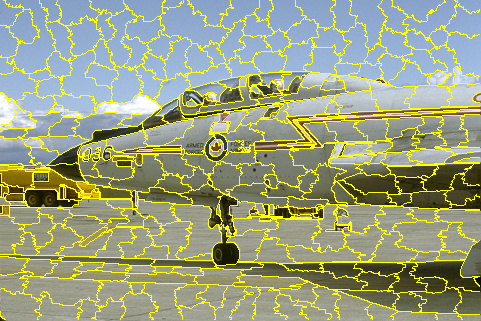}&
\includegraphics[width=\hhh]{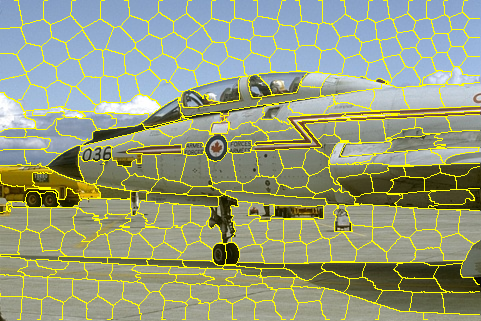}\\
 LNS-Net %\citep{zhu2021learning} 
& AINet %\citep{wang2021ainet} 
& \revision{VSSS}
& \revision{BINET}
& \revision{SIT-HSS}
& Ours %\citep{kirillov23sam} + \citep{irving2016mask}
\\[1ex]
\includegraphics[width=\hhh]{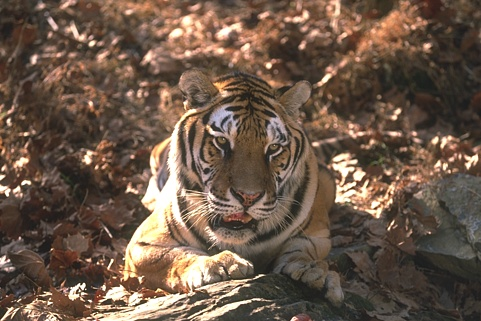}&
\includegraphics[width=\hhh]{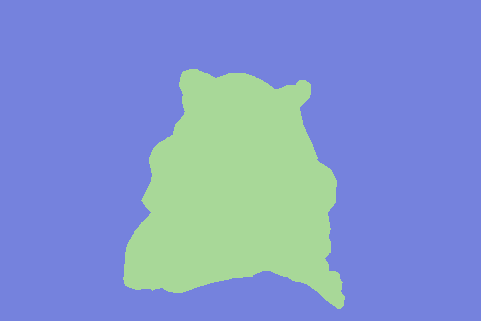}&
\includegraphics[width=\hhh]{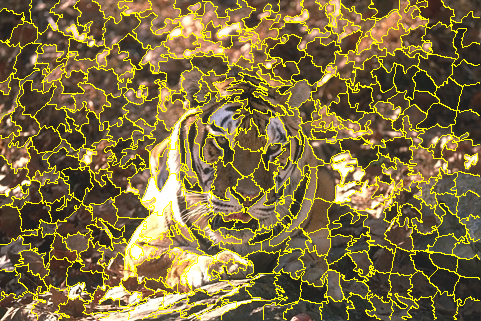}&
\includegraphics[width=\hhh]{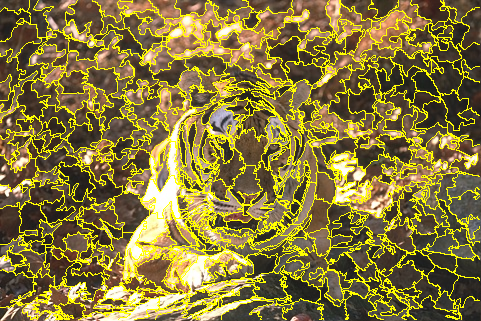}&
\includegraphics[width=\hhh]{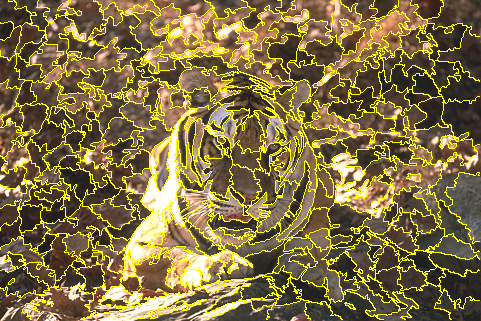}&
\includegraphics[width=\hhh]{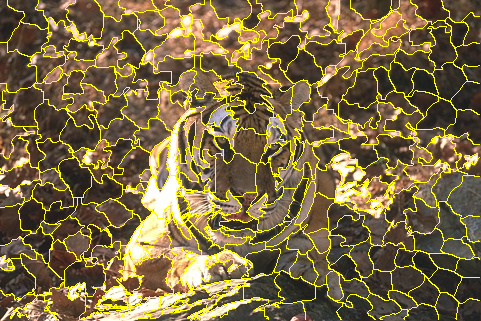}\\
Image & Groundtruth & 
SLIC %\citep{achanta2012} 
& LSC %\citep{li2015} 
& SNIC %\citep{achanta2017superpixels} 
& SCALP
\\[0.5ex]
\includegraphics[width=\hhh]{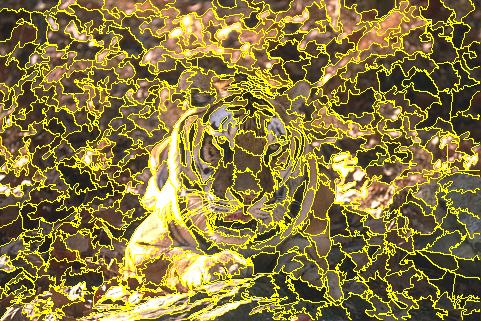}&
\includegraphics[width=\hhh]{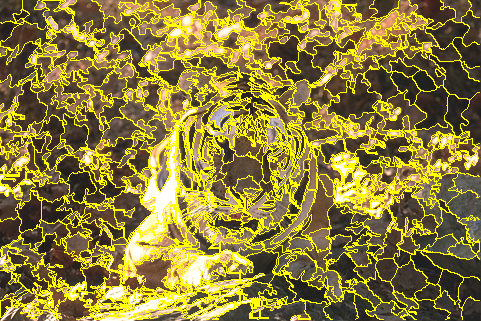}&
\includegraphics[width=\hhh]{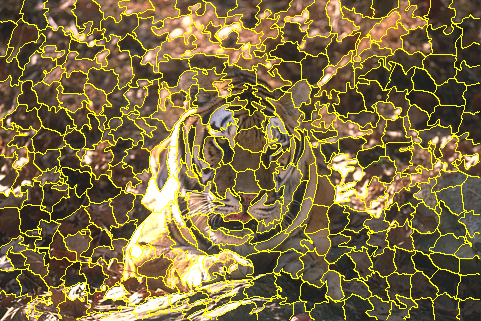}&
\includegraphics[width=\hhh]{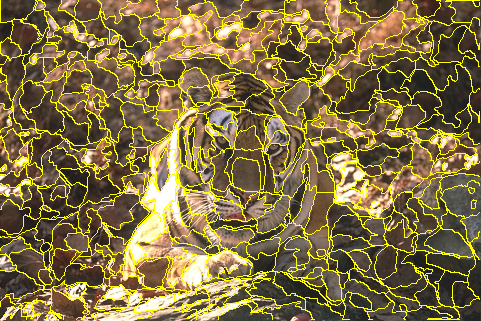}&
\includegraphics[width=\hhh]{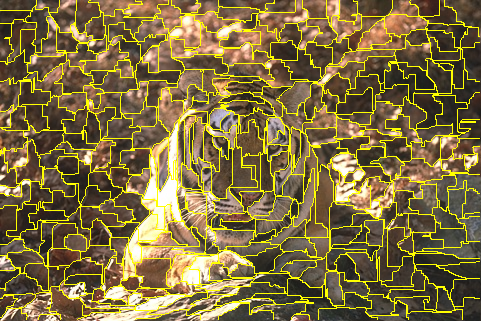}&
\includegraphics[width=\hhh]{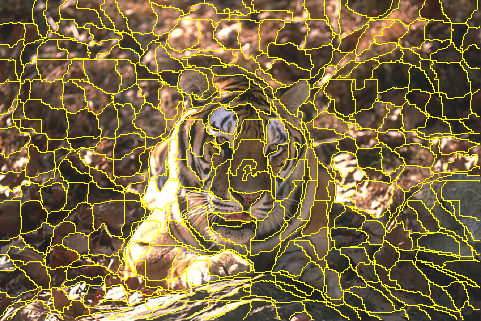}\\
GMMSP %\citep{Ban18} 
& BASS %\citep{Uziel:ICCV:2019:BASS} 
& FSLIC %\citep{wu2020fuzzy}
& SSN %\citep{jampani2018superpixel}
& SEAL
& SFCN %\citep{yang2020superpixel} 
 \\[0.5ex]
\includegraphics[width=\hhh]{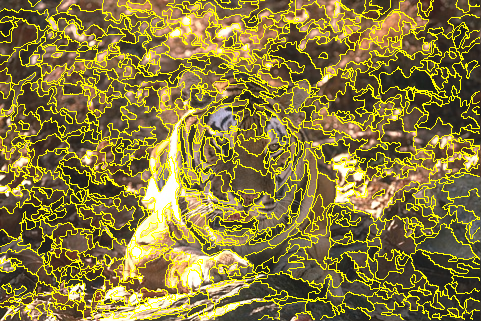}&
\includegraphics[width=\hhh]{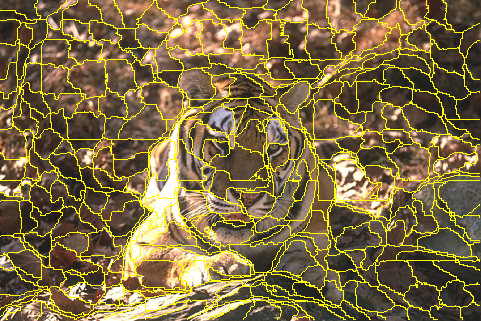}&
\includegraphics[width=\hhh]{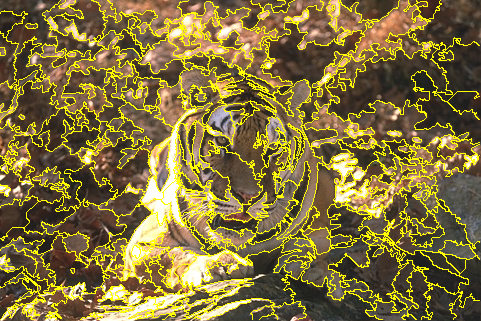}&
\includegraphics[width=\hhh]{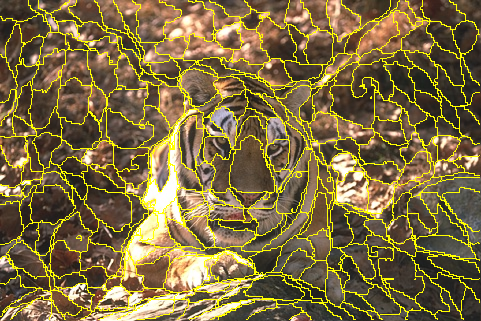}&
\includegraphics[width=\hhh]{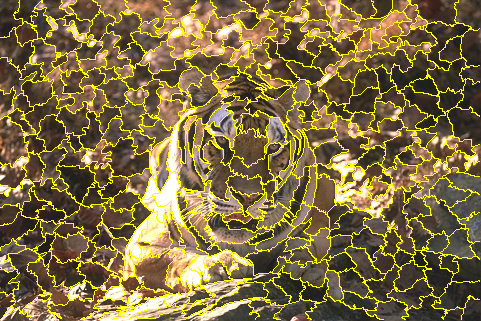}&
\includegraphics[width=\hhh]{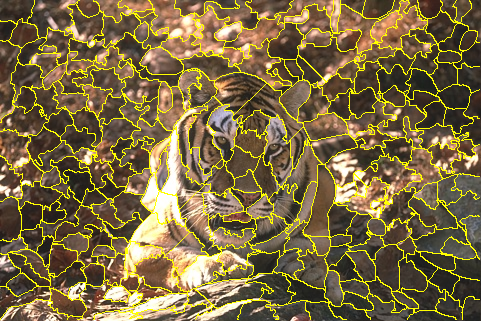}\\
 LNS-Net %\citep{zhu2021learning} 
& AINet %\citep{wang2021ainet} 
& \revision{VSSS}
& \revision{BINET}
& \revision{SIT-HSS}
& Ours %\citep{kirillov23sam} + \citep{irving2016mask}
\\
 %[2ex]
%\includegraphics[width=\hhh]{res_BSD_img_2.png}&
% \includegraphics[width=\hhh]{res_BSD_gt_2.png}&
% \includegraphics[width=\hhh]{res_BSD_slic_2.png}&
% \includegraphics[width=\hhh]{res_BSD_lsc_2.png}&
% \includegraphics[width=\hhh]{res_BSD_snic_2.png}\\
% Image & Groundtruth & SLIC \citep{achanta2012} & LSC \citep{li2015} & SNIC \citep{achanta2017superpixels}\\[0.5ex]
% \includegraphics[width=\hhh]{res_BSD_scalp_2.png}&
% \includegraphics[width=\hhh]{res_BSD_gmmsp_2.png}&
% \includegraphics[width=\hhh]{res_BSD_bass_2.png}&
% \includegraphics[width=\hhh]{res_BSD_fslic_2.png}&
% \includegraphics[width=\hhh]{res_BSD_ssn_2.png}\\
% SCALP \citep{giraud2018_scalp} & GMMSP \citep{Ban18} & BASS \citep{Uziel:ICCV:2019:BASS} & FSLIC \citep{wu2020fuzzy} & SSN \citep{jampani2018superpixel}\\[0.5ex]
% \includegraphics[width=\hhh]{res_BSD_seal_2.png}&
% \includegraphics[width=\hhh]{res_BSD_sfcn_2.png}&
% \includegraphics[width=\hhh]{res_BSD_lnsnet_2.png}&
% \includegraphics[width=\hhh]{res_BSD_ainet_2.png}&
% \includegraphics[width=\hhh]{res_BSD_samo_v5_2.png}\\
% SEAL \citep{tu2018learning} & SFCN \citep{yang2020superpixel} & LNS-Net \citep{zhu2021learning} & AINet \citep{wang2021ainet} & Ours \citep{kirillov23sam} + \citep{irving2016mask}\\ %[-2ex]
\end{tabular}}
\caption{\textbf{Qualitative evaluation of superpixel methods on BSD images.}
The number of requested superpixels is set to 350. 
By allowing irregular superpixels for high-level object segmentation, the proposed approach using SAM \citep{kirillov23sam} and maskSLIC \citep{irving2016mask} generates very accurate, 
and much more overall regular superpixels than other deep learning-based methods.
  } 
\label{fig:res_quali}
\end{figure*}

\begin{figure*}[t!]
\centering
\newcommand{\hhh}{0.1575\textwidth}
{\scriptsize
  \begin{tabular}{@{\hspace{0mm}}c@{\hspace{1mm}}c@{\hspace{1mm}}c@{\hspace{1mm}}c@{\hspace{1mm}}c@{\hspace{1mm}}c@{\hspace{1mm}}c@{\hspace{1mm}}c@{\hspace{1mm}}@{\hspace{0mm}}}
 \includegraphics[width=\hhh]{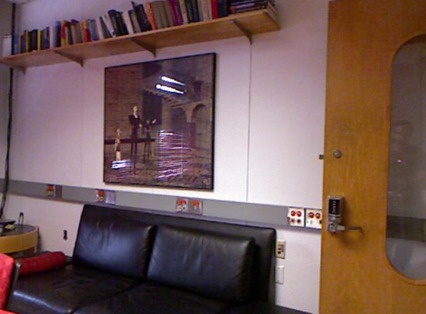}&
 \includegraphics[width=\hhh]{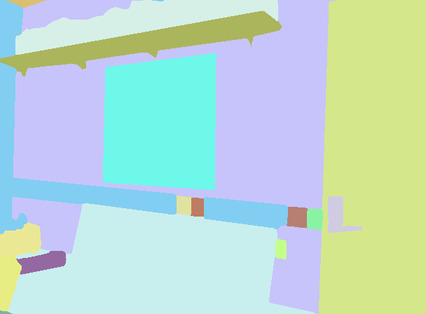}&
 \includegraphics[width=\hhh]{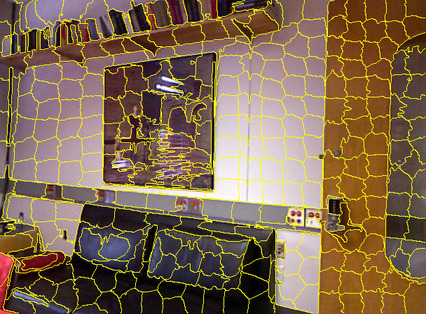}&
 \includegraphics[width=\hhh]{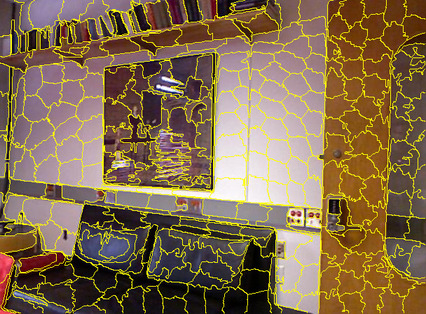}&
 \includegraphics[width=\hhh]{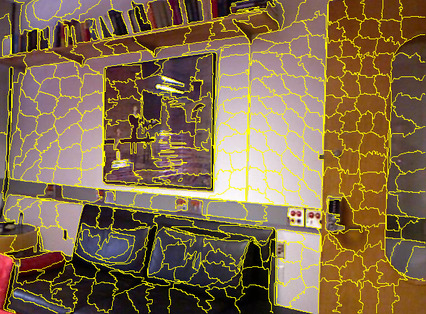}&
  \includegraphics[width=\hhh]{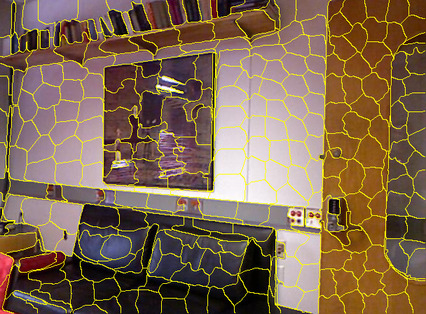} \\
  Image & Groundtruth & 
SLIC %\cite{achanta2012} 
& LSC %\cite{li2015} 
&  SNIC %\cite{achanta2017superpixels}
& SCALP 
\\[0.5ex]
 \includegraphics[width=\hhh]{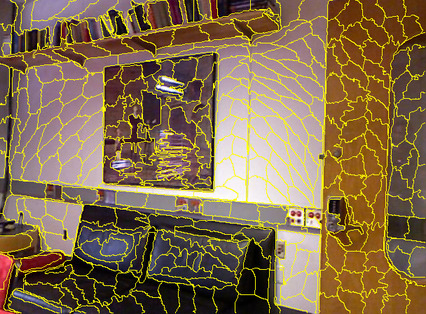}&
 \includegraphics[width=\hhh]{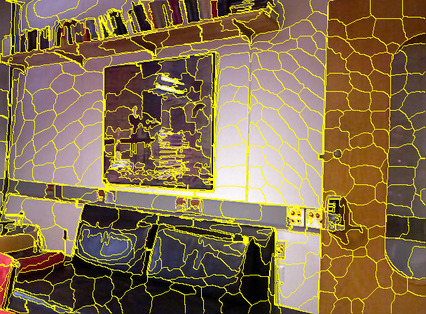}&
  \includegraphics[width=\hhh]{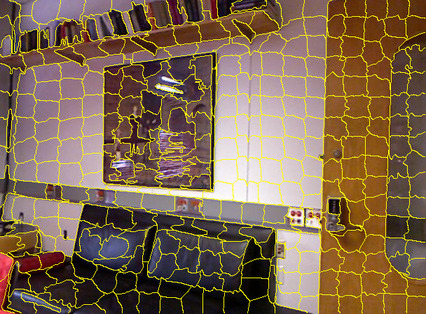}&
 \includegraphics[width=\hhh]{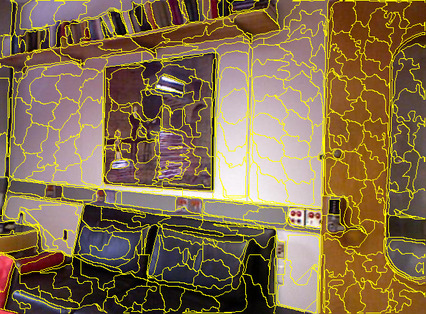}&
 \includegraphics[width=\hhh]{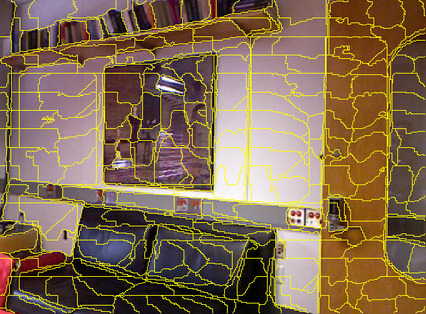}&
 \includegraphics[width=\hhh]{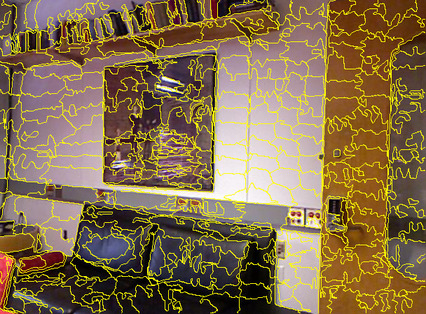}\\
GMMSP %\cite{Ban18} 
& BASS %\cite{Uziel:ICCV:2019:BASS}
& FSLIC %\cite{wu2020fuzzy} 
& SSN %\cite{jampani2018superpixel}
& SFCN
& LNS-Net
\\[0.5ex]
\end{tabular}
  \begin{tabular}{@{\hspace{0mm}}c@{\hspace{1mm}}c@{\hspace{1mm}}c@{\hspace{1mm}}c@{\hspace{1mm}}c@{\hspace{1mm}}@{\hspace{0mm}}}
 \includegraphics[width=\hhh]{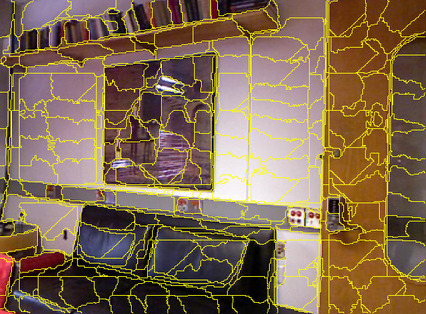}&
 \includegraphics[width=\hhh]{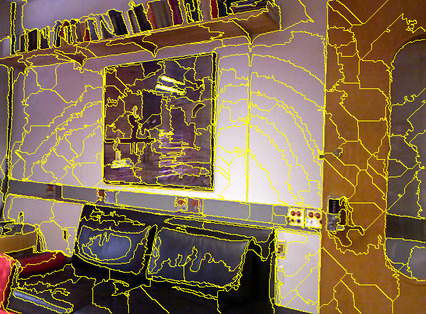}&
 \includegraphics[width=\hhh]{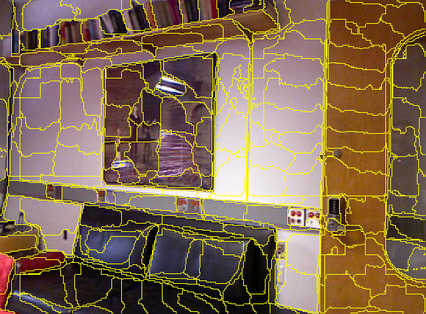}&
 \includegraphics[width=\hhh]{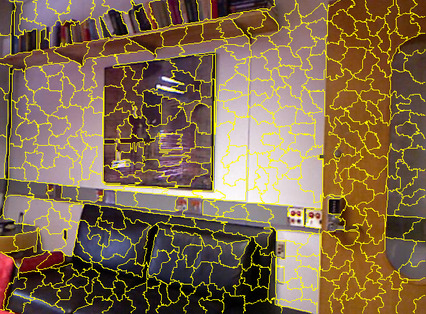}&
 \includegraphics[width=\hhh]{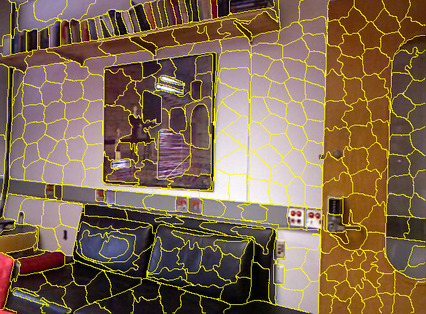}\\
AINet %\cite{wang2021ainet} 
& \revision{VSSS}
& \revision{BINET} 
& \revision{SIT-HSS}
& Ours %\cite{kirillov23sam} + \cite{irving2016mask}
\\[1ex]% \end{tabular}
\end{tabular}
  \begin{tabular}{@{\hspace{0mm}}c@{\hspace{1mm}}c@{\hspace{1mm}}c@{\hspace{1mm}}c@{\hspace{1mm}}c@{\hspace{1mm}}c@{\hspace{1mm}}@{\hspace{0mm}}}
  \includegraphics[width=\hhh]{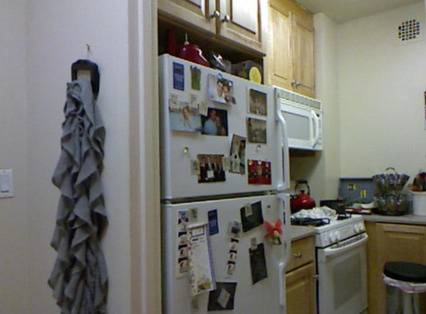}&
 \includegraphics[width=\hhh]{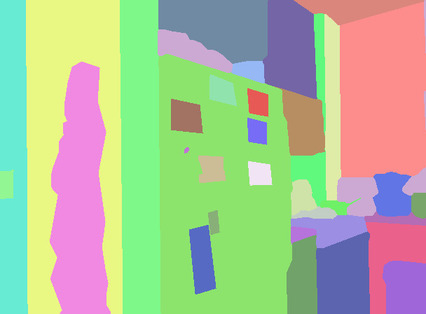}&
 \includegraphics[width=\hhh]{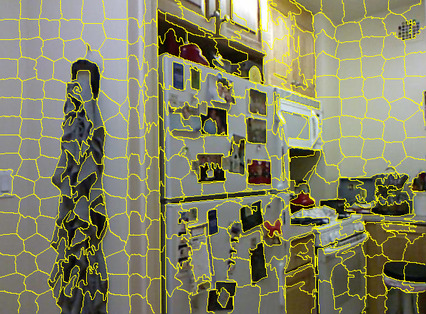}&
 \includegraphics[width=\hhh]{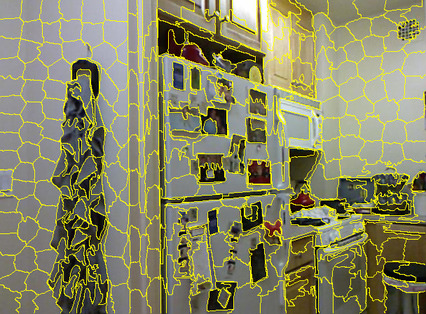}&
 \includegraphics[width=\hhh]{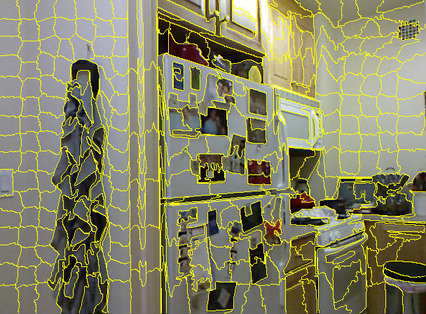}&
  \includegraphics[width=\hhh]{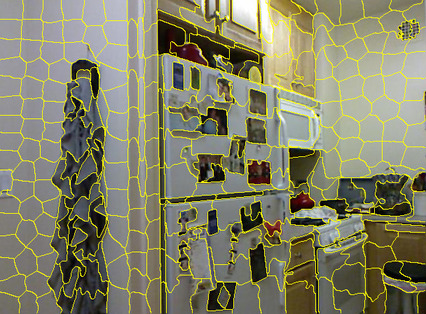} \\
  Image & Groundtruth & 
SLIC %\cite{achanta2012} 
& LSC %\cite{li2015} 
&  SNIC %\cite{achanta2017superpixels}
& SCALP 
\\[0.5ex]
 \includegraphics[width=\hhh]{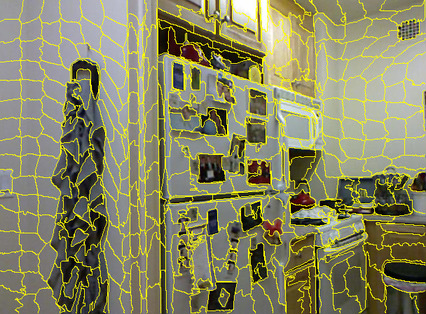}&
 \includegraphics[width=\hhh]{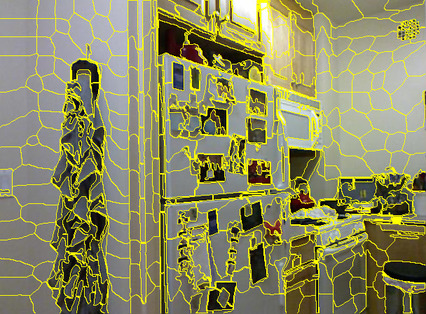}&
  \includegraphics[width=\hhh]{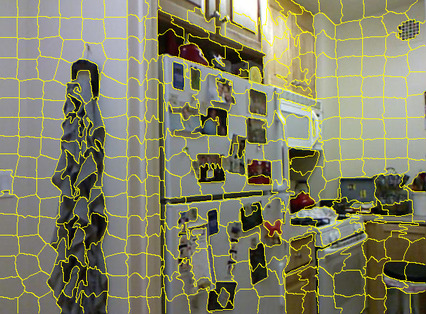}&
 \includegraphics[width=\hhh]{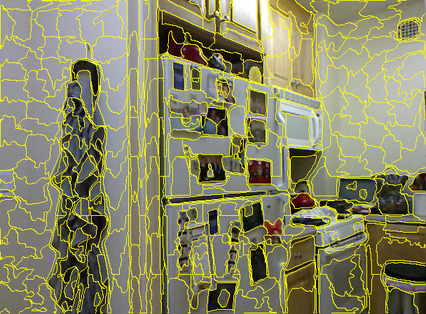}&
 \includegraphics[width=\hhh]{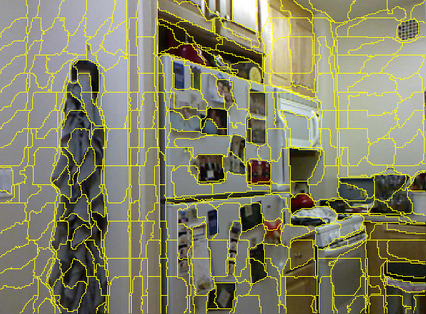}&
 \includegraphics[width=\hhh]{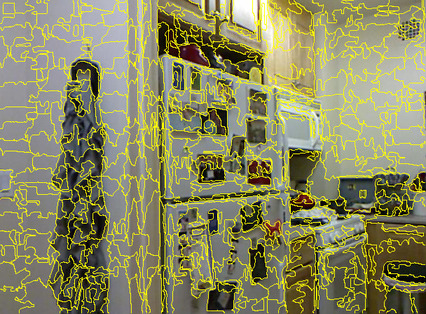}\\
GMMSP %\cite{Ban18} 
& BASS %\cite{Uziel:ICCV:2019:BASS}
& FSLIC %\cite{wu2020fuzzy} 
& SSN %\cite{jampani2018superpixel}
& SFCN
& LNS-Net
\\[0.5ex]
\end{tabular}
  \begin{tabular}{@{\hspace{0mm}}c@{\hspace{1mm}}c@{\hspace{1mm}}c@{\hspace{1mm}}c@{\hspace{1mm}}c@{\hspace{1mm}}@{\hspace{0mm}}}
 \includegraphics[width=\hhh]{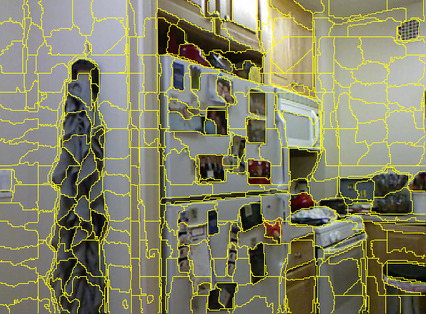}&
 \includegraphics[width=\hhh]{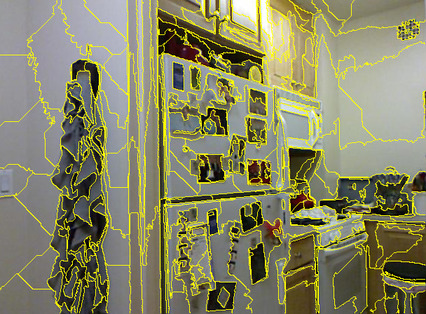}&
 \includegraphics[width=\hhh]{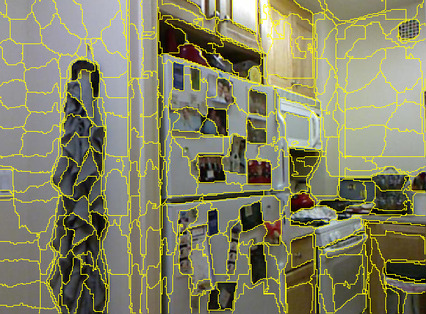}&
 \includegraphics[width=\hhh]{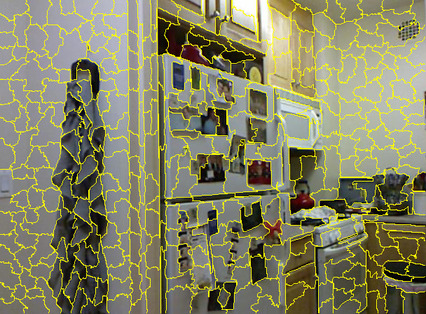}&
 \includegraphics[width=\hhh]{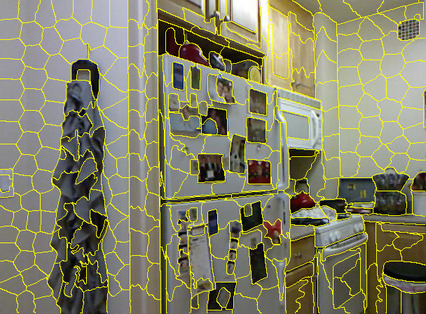}\\
AINet %\cite{wang2021ainet} 
& \revision{VSSS}
& \revision{BINET} 
& \revision{SIT-HSS}
& Ours %\cite{kirillov23sam} + \cite{irving2016mask}
\\[1ex]% \end{tabular}
\end{tabular}}
\caption{\textbf{Qualitative evaluation of superpixel methods on NYUV2 images.}
The number of requested superpixels is set to 350. 
Note that the SEAL method \citep{tu2018learning} cannot deal with images of such size (448x608).
By allowing irregular superpixels for high-level object segmentation, the proposed approach using SAM \citep{kirillov23sam} and maskSLIC \citep{irving2016mask} generates very accurate, 
and much more overall regular superpixels than other deep learning-based methods.
%Our method using SAM \cite{kirillov23sam} and maskSLIC \cite{irving2016mask} generates very accurate, and much more regular superpixels than other deep learning-based methods.
  } 
\label{fig:nyuv2_ex}
\end{figure*}

\begin{figure*}[t!]
\centering
\newcommand{\hhh}{0.1575\textwidth}
{\scriptsize
 \begin{tabular}{@{\hspace{0mm}}c@{\hspace{1mm}}c@{\hspace{1mm}}c@{\hspace{1mm}}c@{\hspace{1mm}}c@{\hspace{1mm}}c@{\hspace{1mm}}c@{\hspace{1mm}}@{\hspace{0mm}}}
 \includegraphics[width=\hhh]{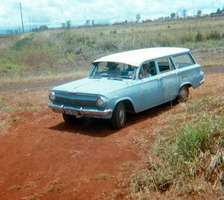}&
 \includegraphics[width=\hhh]{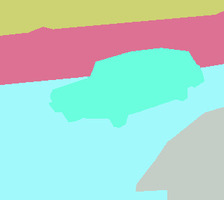}&
  \includegraphics[width=\hhh]{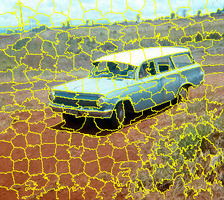}&
\includegraphics[width=\hhh]{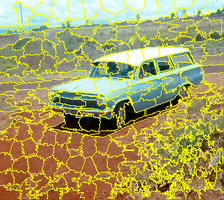}&
 \includegraphics[width=\hhh]{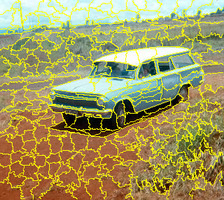}&
   \includegraphics[width=\hhh]{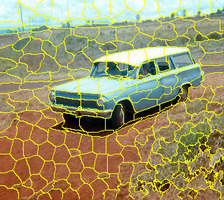}\\
  Image & Groundtruth & 
SLIC  %\cite{achanta2012} 
& LSC %\cite{li2015} 
& SNIC %\cite{achanta2017superpixels}
& SCALP %\cite{giraud2018_scalp} 
\\[0.5ex]
 \includegraphics[width=\hhh]{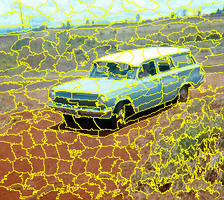}&
 \includegraphics[width=\hhh]{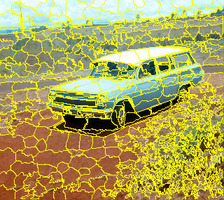}&
 \includegraphics[width=\hhh]{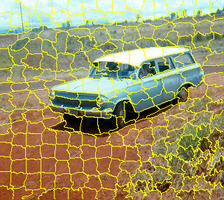}&
  \includegraphics[width=\hhh]{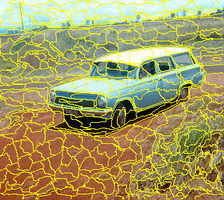}&
   \includegraphics[width=\hhh]{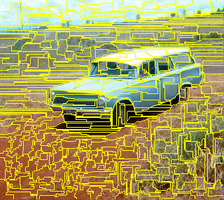}&
 \includegraphics[width=\hhh]{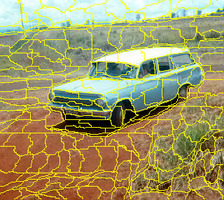}\\
 GMMSP %\cite{Ban18} 
 & BASS %\cite{Uziel:ICCV:2019:BASS} 
 & FSLIC %\cite{wu2020fuzzy} 
 & SSN %\cite{jampani2018superpixel}
 & SEAL %\cite{tu2018learning} 
 & SFCN
 \\[0.5ex]
 \includegraphics[width=\hhh]{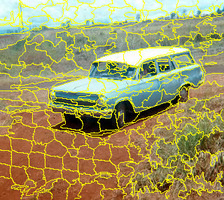}&
 \includegraphics[width=\hhh]{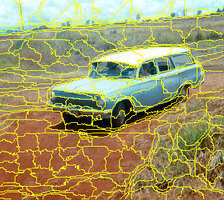}&
 \includegraphics[width=\hhh]{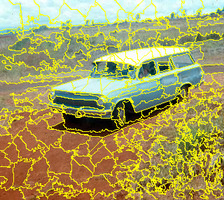}&
 \includegraphics[width=\hhh]{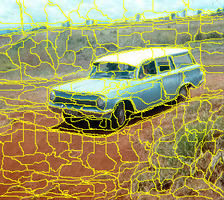}&
 \includegraphics[width=\hhh]{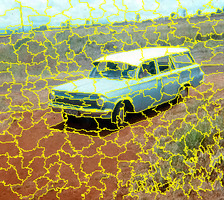}&
 \includegraphics[width=\hhh]{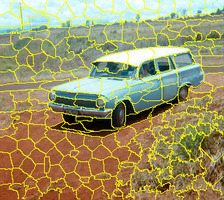}\\
   %\cite{yang2020superpixel} 
 LNS-Net %\cite{zhu2021learning} 
 & AINet %\cite{wang2021ainet} 
 & \revision{VSSS}
 & \revision{BINet} %\cite{wang2021ainet} 
 & \revision{SIT-HSS} %\cite{wang2021ainet} 
 & Ours %\cite{kirillov23sam} + \cite{irving2016mask}
 \\[1ex]
 \includegraphics[width=\hhh]{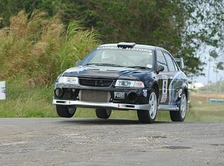}&
 \includegraphics[width=\hhh]{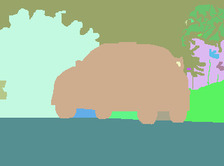}&
  \includegraphics[width=\hhh]{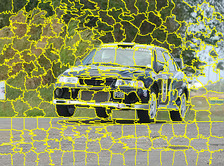}&
\includegraphics[width=\hhh]{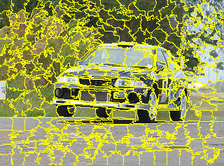}&
 \includegraphics[width=\hhh]{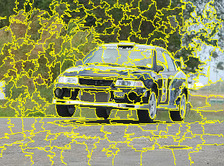}&
   \includegraphics[width=\hhh]{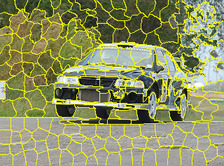}\\
  Image & Groundtruth & 
SLIC  %\cite{achanta2012} 
& LSC %\cite{li2015} 
& SNIC %\cite{achanta2017superpixels}
& SCALP %\cite{giraud2018_scalp} 
\\[0.5ex]
 \includegraphics[width=\hhh]{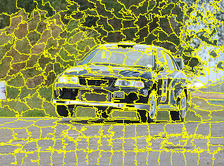}&
 \includegraphics[width=\hhh]{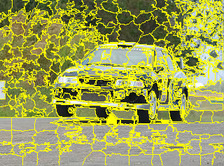}&
 \includegraphics[width=\hhh]{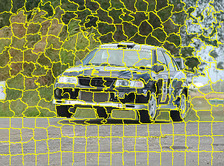}&
  \includegraphics[width=\hhh]{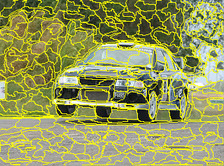}&
   \includegraphics[width=\hhh]{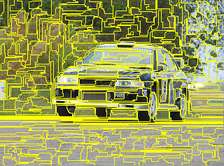}&
 \includegraphics[width=\hhh]{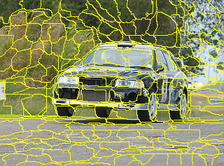}\\
  GMMSP %\cite{Ban18} 
 & BASS %\cite{Uziel:ICCV:2019:BASS} 
 & FSLIC %\cite{wu2020fuzzy} 
 & SSN %\cite{jampani2018superpixel}
 & SEAL %\cite{tu2018learning} 
 & SFCN
 \\[0.5ex]
 \includegraphics[width=\hhh]{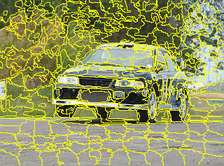}&
 \includegraphics[width=\hhh]{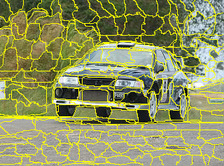}&
 \includegraphics[width=\hhh]{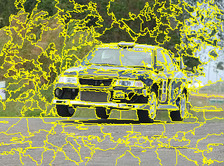}&
 \includegraphics[width=\hhh]{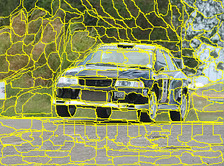}&
 \includegraphics[width=\hhh]{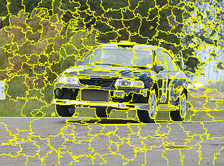}&
 \includegraphics[width=\hhh]{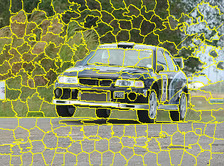}\\
   %\cite{yang2020superpixel} 
  LNS-Net %\cite{zhu2021learning} 
 & AINet %\cite{wang2021ainet} 
 & \revision{VSSS}
 & \revision{BINet} %\cite{wang2021ainet} 
 & \revision{SIT-HSS} %\cite{wang2021ainet} 
 & Ours %\cite{kirillov23sam} + \cite{irving2016mask}
\end{tabular}}
\caption{\textbf{Qualitative evaluation of superpixel methods on SBD images.}
The number of requested superpixels is set to 250. 
By allowing irregular superpixels for high-level object segmentation, the proposed approach using SAM \citep{kirillov23sam} and maskSLIC \citep{irving2016mask} generates very accurate, 
and much more overall regular superpixels than other deep learning-based methods.
%Our method using SAM \cite{kirillov23sam} and maskSLIC \cite{irving2016mask} generates very accurate, and much more regular superpixels than other deep learning-based methods.
} 
\label{fig:sbd_ex}
\end{figure*}

\begin{figure*}[t!]
\centering
\newcommand{\hhh}{0.19\textwidth}
{\scriptsize
 \begin{tabular}{@{\hspace{0mm}}c@{\hspace{1mm}}c@{\hspace{1mm}}c@{\hspace{1mm}}c@{\hspace{1mm}}c@{\hspace{1mm}}c@{\hspace{1mm}}c@{\hspace{1mm}}@{\hspace{0mm}}}
 \includegraphics[width=\hhh]{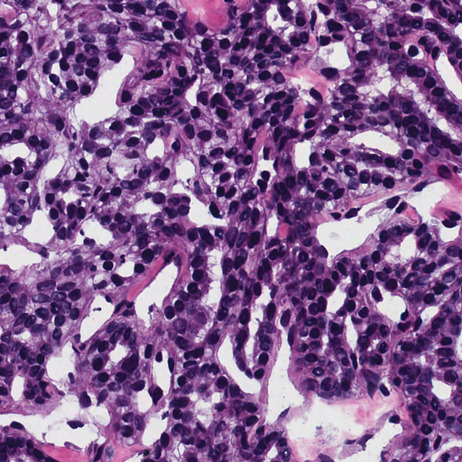}&
 \includegraphics[width=\hhh]{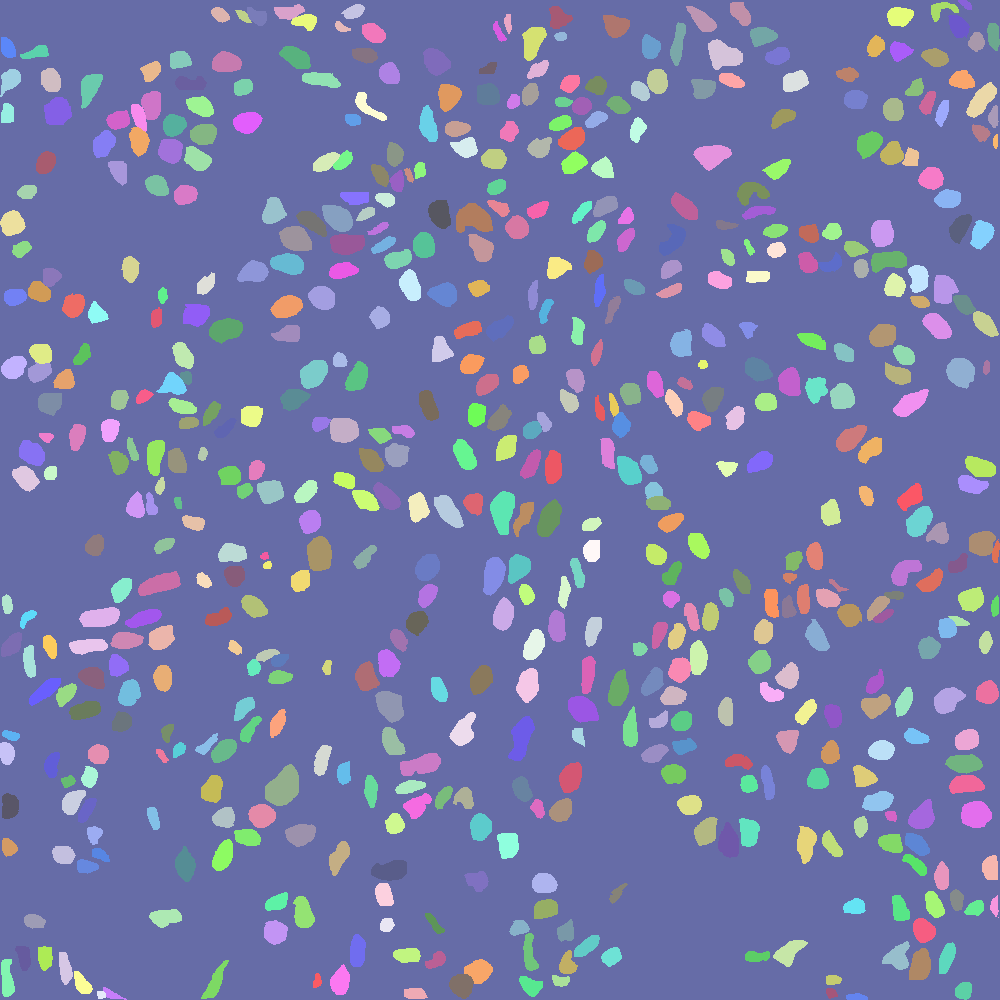}&
  \includegraphics[width=\hhh]{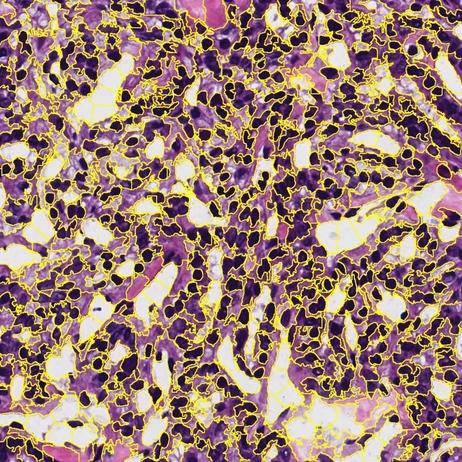}&
\includegraphics[width=\hhh]{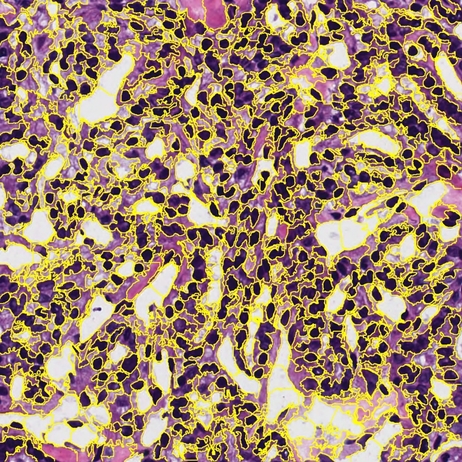}&
 \includegraphics[width=\hhh]{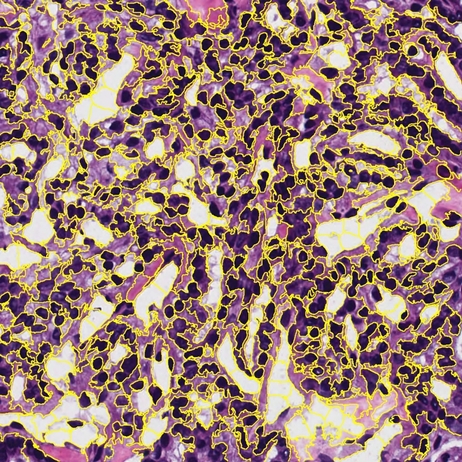}\\
  Image & Groundtruth & 
SLIC  %\cite{achanta2012} 
& LSC %\cite{li2015} 
& SNIC %\cite{achanta2017superpixels}
\\[0.5ex]
   \includegraphics[width=\hhh]{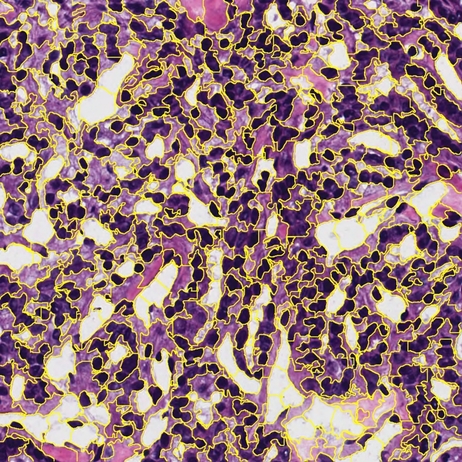}&
 \includegraphics[width=\hhh]{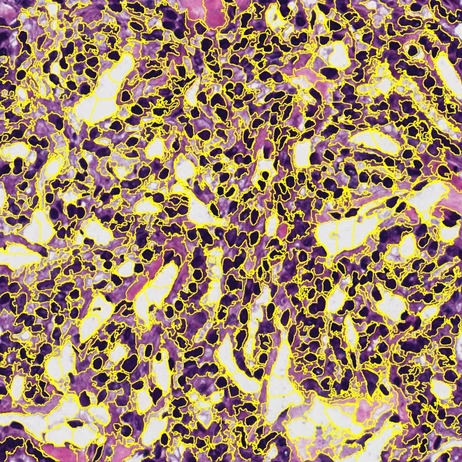}&
 \includegraphics[width=\hhh]{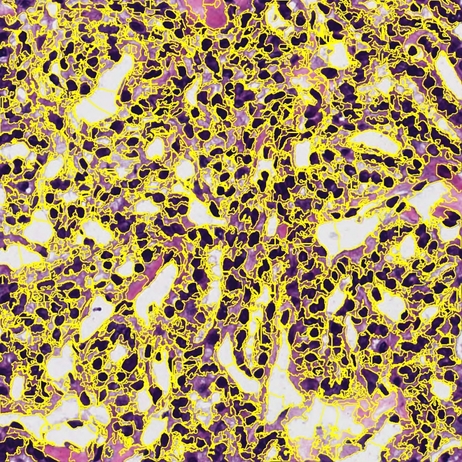}&
 \includegraphics[width=\hhh]{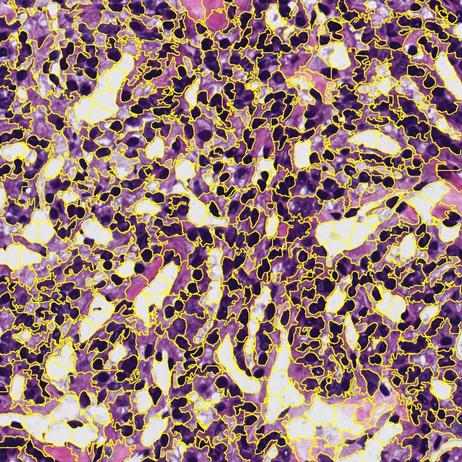}&
  \includegraphics[width=\hhh]{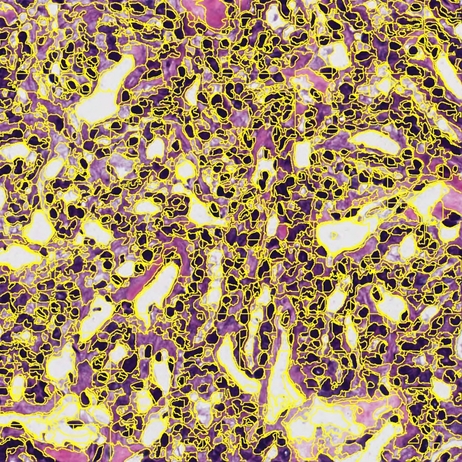}\\
 SCALP %\cite{giraud2018_scalp} 
 & GMMSP %\cite{Ban18} 
 & BASS %\cite{Uziel:ICCV:2019:BASS} 
 & FSLIC %\cite{wu2020fuzzy} 
 & SSN %\cite{jampani2018superpixel} 
 \\[0.5ex]
 \includegraphics[width=\hhh]{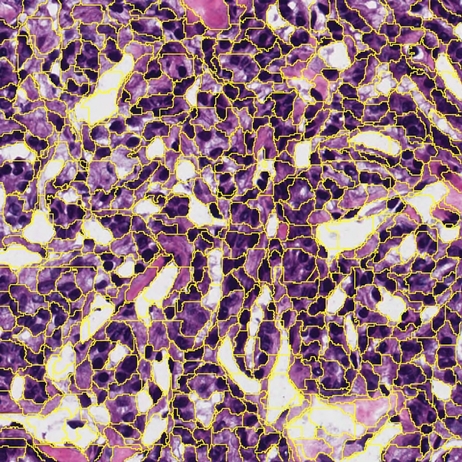}&
 \includegraphics[width=\hhh]{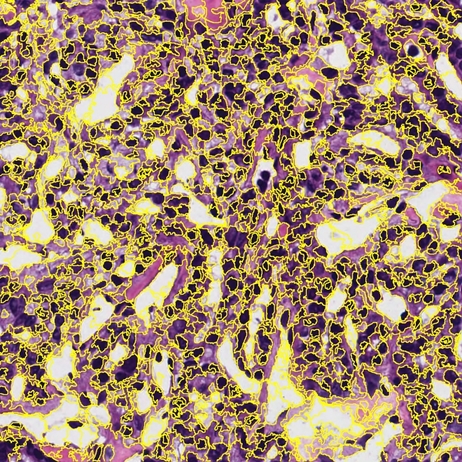}&
 \includegraphics[width=\hhh]{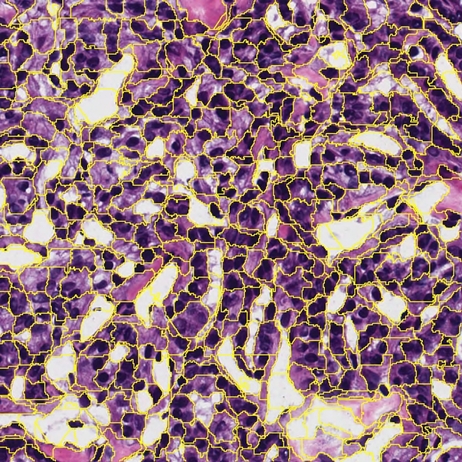}&
 \includegraphics[width=\hhh]{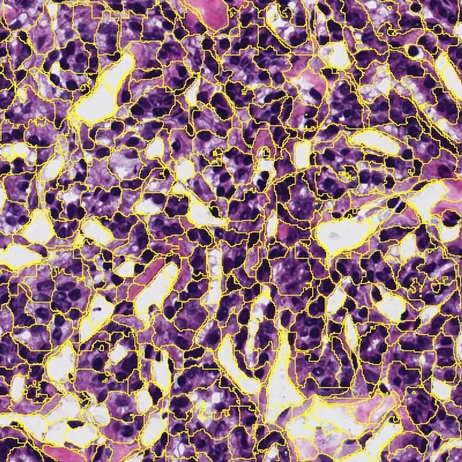}&
 \includegraphics[width=\hhh]{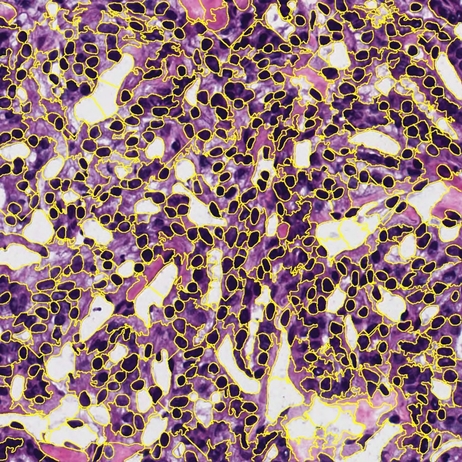}\\
   %\cite{yang2020superpixel} 
   SFCN 
 &LNS-Net %\cite{zhu2021learning} 
 & AINet %\cite{wang2021ainet} 
 & \revision{BINet} %\cite{wang2021ainet} 
 & Ours %\cite{kirillov23sam} + \cite{irving2016mask}
\end{tabular}}
\caption{\revision{\textbf{Qualitative evaluation of superpixel methods on a MoNuSeg image.}
The number of requested superpixels is set to 500. 
Note that some methods cannot processed such images of size 1000x1000 pixels.
With the high-level objet prior from \citep{kirillov23sam},
our method is able to generalize to such images and more accurately capture the nuclei.
%and maskSLIC \cite{irving2016mask} generates very accurate, and much more regular superpixels than other deep learning-based methods.
}} 
\label{fig:monuseg_ex}
\end{figure*}

\begin{figure*}[t!]
\centering
\newcommand{\hhh}{0.1575\textwidth}
{\scriptsize
 \begin{tabular}{@{\hspace{0mm}}c@{\hspace{1mm}}c@{\hspace{1mm}}c@{\hspace{1mm}}c@{\hspace{1mm}}c@{\hspace{1mm}}c@{\hspace{1mm}}c@{\hspace{1mm}}@{\hspace{0mm}}}
 \includegraphics[width=\hhh]{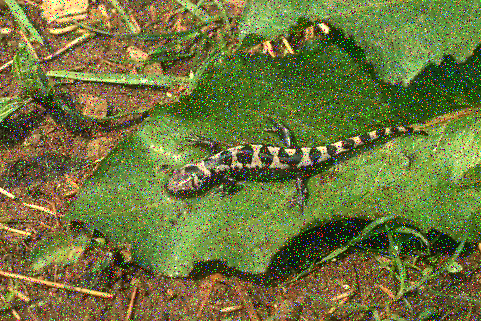}&
 \includegraphics[width=\hhh]{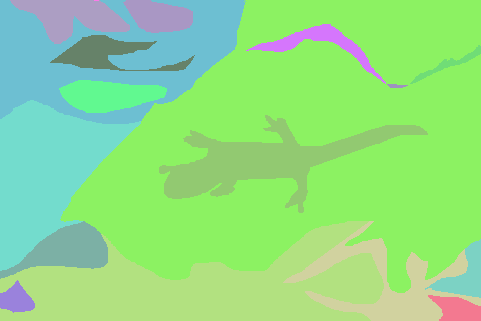}&
  \includegraphics[width=\hhh]{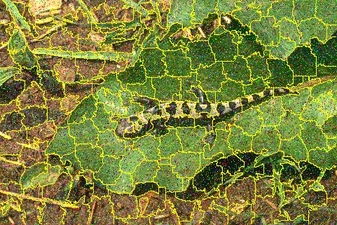}&
\includegraphics[width=\hhh]{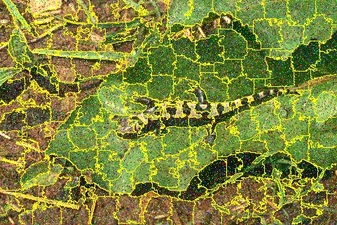}&
 \includegraphics[width=\hhh]{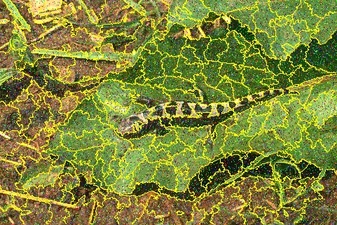}&
   \includegraphics[width=\hhh]{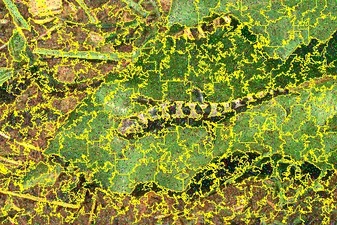}\\
  Image & Groundtruth & 
SLIC  %\cite{achanta2012} 
& LSC %\cite{li2015} 
& SNIC %\cite{achanta2017superpixels}
& SCALP %\cite{giraud2018_scalp} 
\\[0.5ex]
 \includegraphics[width=\hhh]{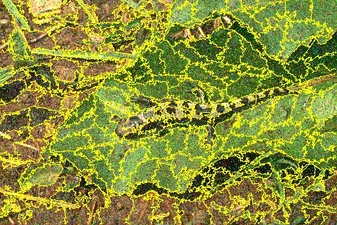}&
 \includegraphics[width=\hhh]{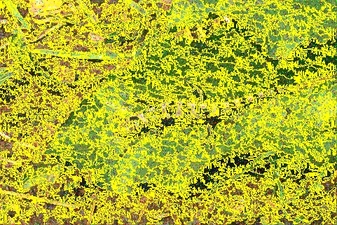}&
 \includegraphics[width=\hhh]{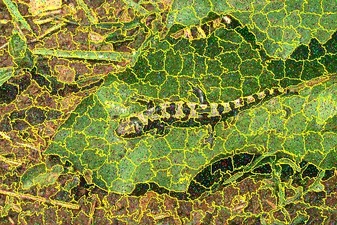}&
  \includegraphics[width=\hhh]{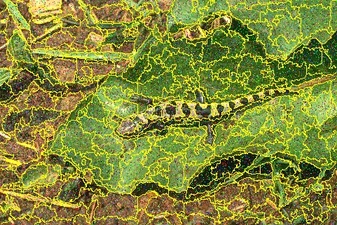}&
   \includegraphics[width=\hhh]{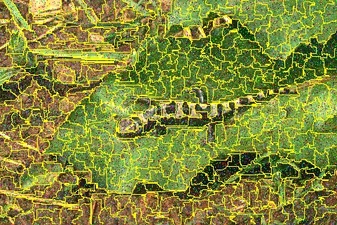}&
 \includegraphics[width=\hhh]{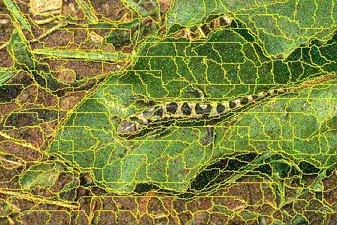}\\
 GMMSP %\cite{Ban18} 
 & BASS %\cite{Uziel:ICCV:2019:BASS} 
 & FSLIC %\cite{wu2020fuzzy} 
 & SSN %\cite{jampani2018superpixel}
 & SEAL %\cite{tu2018learning} 
 & SFCN
 \\[0.5ex]
 \includegraphics[width=\hhh]{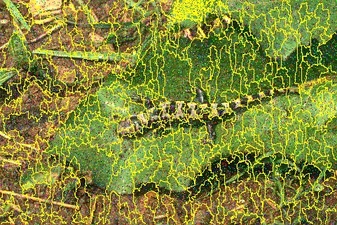}&
 \includegraphics[width=\hhh]{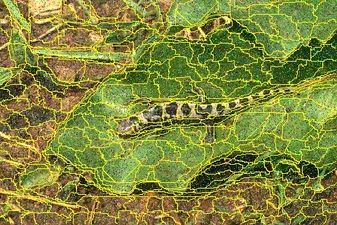}&
 \includegraphics[width=\hhh]{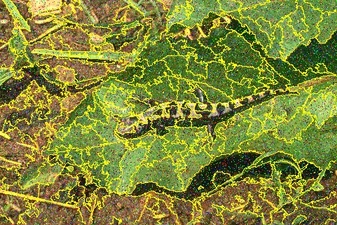}&
 \includegraphics[width=\hhh]{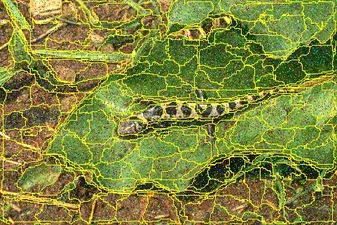}&
 \includegraphics[width=\hhh]{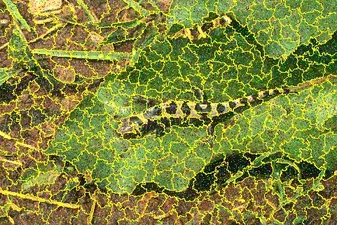}&
 \includegraphics[width=\hhh]{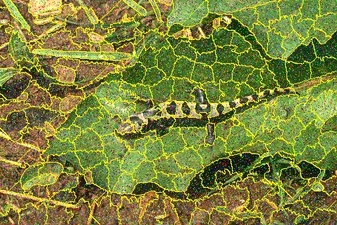}\\
   %\cite{yang2020superpixel} 
 LNS-Net %\cite{zhu2021learning} 
 & AINet %\cite{wang2021ainet} 
 & \revision{VSSS}
 & \revision{BINet} %\cite{wang2021ainet} 
 & \revision{SIT-HSS} %\cite{wang2021ainet} 
 & Ours %\cite{kirillov23sam} + \cite{irving2016mask}
 \\
\end{tabular}}
\caption{\revision{\textbf{Qualitative evaluation of superpixel methods on a noisy image of the BSD (impulse Salt-and-Pepper noise with a density of 0.1).}
The number of requested superpixels is set to 350. 
Thanks to the robustness to noise of the high-level prior, our approach remains comparatively robust as the noise level increases, whereas some competing methods experience more significant performance degradation or very noisy superpixel borders.
}}
\label{fig:sp_noise_ex}
\end{figure*}

The overall high regularity and accuracy can be visually assessed on BSD, NYUV2, SBD and MoNuSeg images in Figure \ref{fig:res_quali}, \ref{fig:nyuv2_ex}, \ref{fig:sbd_ex} and \ref{fig:monuseg_ex} respectively. 
%and in the supp. mat. for NYUV2 and SBD images.
%
We also show in Figure \ref{fig:sp_noise_ex} segmentations on
a noisy BSD image.
Initial object proposal allows to catch large structures, 
while later maskSLIC generates more regular superpixels within.
We observe noisy superpixels, \emph{e.g.}, 
for traditional LSC or GMMSP methods, 
and the irregular nature of DL-based methods.
%where our methods generates very regular superpixels that respect the image objects.
%

\smallskip

\paragraph{Discussion.} 
The presented approach obtains state-of-the-art performance, outperforming recent DL architectures, without even being trained specifically to generate superpixels.
These results demonstrate that 
superpixel segmentation can be efficiently tackled as a high-level object segmentation problem, followed by a standard local pixel-level clustering to get the best of both worlds, \emph{i.e.}, to generate accurate yet regular superpixels.

\revision{To strengthen this observation, we also reported the performance obtained using SAM alone, without the mask-based superpixel refinement (large diamond marker).
As expected, SAM achieves relatively low superpixel-oriented performance, mainly due to the limited number of regions it segments and the presence of uncovered areas. This highlights the complementarity between high-level region segmentation models such as SAM and the finer, low-level clustering scale of superpixels that may help to capture smaller regions or object subparts. 
%We have clarified this point in the introduction of the proposed method to better explain that our approach does not reduce superpixel segmentation to object segmentation, but rather leverages high-level priors to guide a subsequent regular and dense partitioning.
}

\revision{Our method is, to the best of our knowledge, the first to jointly maximize segmentation accuracy and global regularity, thereby challenging the traditional assumption that these two properties are necessarily opposed.
In our case, regular superpixels remain particularly beneficial in structured pipelines where spatial consistency and region comparability are required, such as graph-based matching \citep{gould2008}, region adjacency graph construction, or patch-based correspondence methods like SuperPatchMatch \citep{giraud2017_spm}. They are also valuable in interactive segmentation frameworks such as Ilastik \citep{berg2019ilastik}, where clearly identifiable and uniformly sized regions improve user control and stability.
%In such contexts—graph matching, region-level feature aggregation, or superpixel-based tracking—the regular layout provides robustness and interpretability that highly irregular decompositions may fail to guarantee.
}
For instance, a user could generate superpixels in one specific predefined mask to go deeper in the annotation, \emph{e.g.}, by identifying subparts of an
object, where each would be captured by a superpixel.

%
%Our method provides a generic segmentation tool producing accurate and regular superpixels, that can be easily adapted in terms of scale.

Finally, the use of a large-scale pretrained model like SAM also
enables a broader generalization capacity \citep{kirillov23sam}, 
and such models 
have been shown to be finetuned very efficiently to more specific domains.

\section{Conclusion}

In this work, we exposed %{\color{red} showed, demonstrated?} 
the ill-posed nature of the superpixel segmentation problem,
due in particular to an implicit regularity constraint.
%
%This context require a relevant evaluation of methods,
%
%While numerous metrics exist, 
This paradigm leads to a potentially biased evaluation of methods 
on several aspects.
Nevertheless, we demonstrated how to relevantly use one key metric per property,
\revision{and we encourage the research community to adopt these recommendations, \emph{i.e.},
using at least this set of metrics to consistently evaluate the four main aspects of superpixel segmentation.}
%to ensure a fair minimal evaluation of methods.

%This work also highlights the  different behaviors of methods that may impact
%to carefully choose a superpixel method, 
%the targeted downstream task. 
This context has allowed the %significant 
shift of deep learning-based methods 
towards irregular superpixels focused solely on object segmentation performance.
%
%This context has become so object-centered that
We proved that we can %answer %this ill-posed nature
%respect the initial 
%of superpixel segmentation,
generate accurate superpixels
by directly using an efficient pre-trained semantic-agnostic object segmentation model,
to capture potentially thin and large objects 
and then generating
regular and clearly identifiable superpixels overall.
Future works may focus on improving of the intra-object
superpixel clustering method.
%without the need to trained to generate superpixels.
%
%To such an extent, that we obtained comparable state-of-the-art performance, 
%
%straightforwardly using an object segmentation method that is not even specifically 
%trained to generate superpixels.
%

%accuracy, adjustable regularity, control over the number of superpixels, 
%or hierarchical decomposition.
%
%By addressing these aspects, we can enhance the utility of superpixel segmentation methods, 
%accordint to the targeted downstream task.
%ensuring they meet diverse application needs while maintaining high performance and regularity.

%% Loading bibliography style file
%\bibliographystyle{model1-num-names}
\bibliographystyle{cas-model2-names}
%\setcitestyle{authoryear,open={(},close={)},maxnames=2}

% Loading bibliography database
\bibliography{biblio.bib}

% Biography
%\bio{}
% Here goes the biography details.
%\endbio

%\bio{pic1}
% Here goes the biography details.
%\endbio

\end{document}